\documentclass[sn-nature]{sn-jnl}

\usepackage{graphicx}%
\usepackage{multirow}%
\usepackage{amsmath,amssymb,amsfonts}%
\usepackage{amsthm}%
\usepackage{mathrsfs}%
\usepackage[title]{appendix}%
\usepackage{xcolor}%
\usepackage{textcomp}%
\usepackage{manyfoot}%
\usepackage{booktabs}%
\usepackage{algorithm}%
\usepackage{algorithmicx}%
\usepackage{algpseudocode}%
\usepackage{listings}%

\usepackage{float}
\usepackage{booktabs}
\usepackage{longtable}
\usepackage{tabularx}
\usepackage{ltablex} 
\usepackage{caption}
\usepackage{tcolorbox}
\tcbuselibrary{listingsutf8}
\usepackage[colorinlistoftodos]{todonotes}

\keepXColumns  

\theoremstyle{thmstyleone}%
\theoremstyle{thmstyletwo}%

\theoremstyle{thmstylethree}%

\begin{document}

\title{Human-like Content Analysis for Generative AI with Language-Grounded Sparse Encoders}

\author*[$1$]{\fnm{Yiming} \sur{Tang}}\email{yiming@nus.edu.sg}

\author[$1$]{\fnm{Arash} \sur{Lagzian}}

\author[$1$]{\fnm{Srinivas} \sur{Anumasa}}

\author[$1$]{\fnm{Qiran} \sur{Zou}}

\author[$1,2$]{\fnm{Yingtao} \sur{Zhu}}

\author[$1$]{\fnm{Ye} \sur{Zhang}}

\author[$3$]{\fnm{Trang} \sur{Nguyen}}

\author[$1$]{\fnm{Yih-Chung} \sur{Tham}}

\author[$3$]{\fnm{Ehsan} \sur{Adeli}}

\author[$1$]{\fnm{Ching-Yu} \sur{Cheng}}

\author[$4$]{\fnm{Yilun} \sur{Du}}

\author*[$1$]{\fnm{Dianbo} \sur{Liu}$^\dagger$}\email{dianbo@nus.edu.sg}

\affil[]{$^1$National University of Singapore, Singapore, Singapore} 
\affil[]{$^2$Tsinghua University, Beijing, China}
\affil[]{$^3$Stanford University, Palo Alto, U.S. , $^4$Harvard University, Boston, U.S.}

\abstract{The rapid development of generative AI has transformed content creation, communication, and human development. However, this technology raises profound concerns in high-stakes domains, demanding rigorous methods to analyze and evaluate AI-generated content. While existing analytic methods often treat images as indivisible wholes, real-world AI failures generally manifest as specific visual patterns that can evade holistic detection and suit more granular and decomposed analysis. Here we introduce a content analysis tool, Language-Grounded Sparse Encoders (LanSE), which decompose images into interpretable visual patterns with natural language descriptions. Utilizing interpretability modules and large multimodal models, LanSE can automatically identify visual patterns within data modalities. Our method discovers more than 5,000 visual patterns with 93\% human agreement, provides decomposed evaluation outperforming existing methods, establishes the first systematic evaluation of physical plausibility, and extends to medical imaging settings. Our method's capability to extract language-grounded patterns can be naturally adapted to numerous fields, including biology and geography, as well as other data modalities such as protein structures and time series, thereby advancing content analysis for generative AI.}

\maketitle

\vspace{-0.3cm}
\section*{Introduction}
\label{introduction}

For millennia, human-created content served as the exclusive record of civilization, with generally every painting, text, and diagram traceable to human thought and effort \citep{morriss2010evolution,raaflaub2013thinking, gripshover2020writing}. Generative AI has brought an end to this era \citep{LivingBraveAI2023}, producing seemingly indistinguishable alternatives at unprecedented scale \citep{review2024generative,porter2024ai, nightingale2022ai}, permeating every domain from scientific publications \citep{gil2022will, kitano2021nobel,geng2024impactlargelanguagemodels,zou2026fmlbenchbenchmarkingmachinelearning} to medical diagnostics \citep{mahmood2019deep, teixeira2018generatingsyntheticxrayimages} to artistic designs \citep{rombach2022highresolutionimagesynthesislatent, kottapallitransforming}. Yet this transformative technology raises profound concerns about its deployment across critical domains: from diagnostic reliability in medical applications \citep{alber2025medical, hager2024evaluation, kundu2023measuring} and accelerated misinformation propagation \citep{lindsay2023llms, feng2025unravelingmisinformationpropagationllm} to cultural homogenization \citep{doshi2024generative, suzuki2023we, choudhury2023generative} and data contamination \citep{shumailov2024ai, smith2024ai}. These consequential risks necessitate immediate development of rigorous methods to analyze, evaluate, and control AI-generated content, protecting the integrity and trustworthiness of AI systems in high-stakes applications \citep{10.1093/polsoc/puaf001,KRAKOWSKI2025100560,tang2023integratedforwardforwardalgorithmintegrating,dai2025sanhypothesizinglongtermsynaptic,anumasa2026navigatingheterogeneousproteinlandscapes}.

Content analysis for generative AI encounters a fundamental scope-specificity dilemma. While effective content analysis requires comprehensive assessment across multiple quality dimensions, from prompt adherence to anatomical correctness, real-world failures typically manifest as specific visual patterns that require precisely the granular detection capabilities that many broad-scope methods fail to provide \citep{chen2023understandingvulnerabilityclipimage}. Existing analytic methods typically prioritize either comprehensive scope or detection specificity. Some analytic approaches treat generated content as indivisible units, providing holistic assessments through distributional alignment metrics \citep{jayasumana2024rethinkingfidbetterevaluation, barratt2018noteinceptionscore} or semantic similarity scores \citep{chen2025contrastivelocalizedlanguageimagepretraining, hessel2022clipscorereferencefreeevaluationmetric}, while other developments invent specialized detection tools for specific failure modes—such as text generation errors \citep{yao2024hifi, sampaio2024typescoretextfidelitymetric}, model hallucination \citep{li2023haluevallargescalehallucinationevaluation}, or anatomical anomalies \citep{fang2024humanrefinerbenchmarkingabnormalhuman}. These specialized approaches achieve high precision but lack adaptability across different content domains.

\begin{figure}[h]
    \centering
    \includegraphics[width=\linewidth]{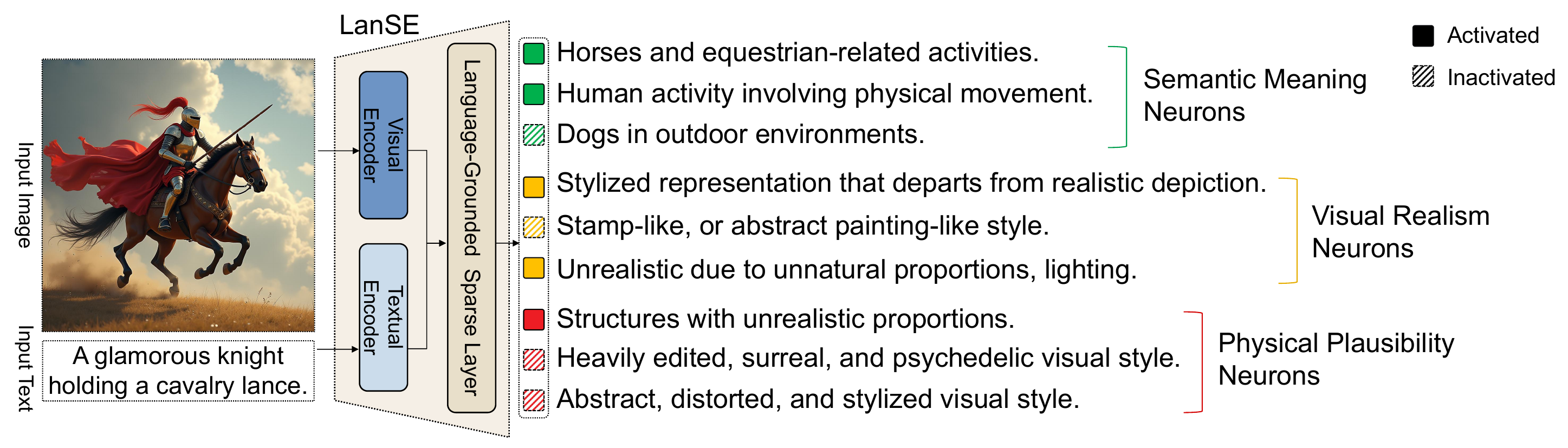}
    \caption{\textbf{Content Analysis with Language-Grounded Sparse Encoders.} LanSE systematically identifies visual patterns in AI-generated content and describes them in natural language (e.g., "dogs in outdoor environments" in natural images and "evidence of bibasilar atelectasis" in medical images). These patterns enable systematic content analysis across multiple applications: quality evaluation, multi-label annotation, and inter-model analysis of generative models across various domains.}
    \label{fig:lanse}
\end{figure}

Comparatively, human evaluation of AI-generated content operates through fundamentally different mechanisms than automated analytic approaches. Consider how art critics assess AI-generated paintings: rather than providing holistic scores, they articulate which aspects work and which fail: the lighting may be striking, the composition balanced, the anatomy flawed, which are all inherently visual patterns interpreted with natural language. This pattern-based evaluation process reflects broader cognitive mechanisms documented in human visual assessment \citep{biederman1987recognition, zeki1999art, clark1996using, nightingale2017can, farid2022creating, rhodes1998facial, Santos03042018, landy2013texture}, suggesting a promising approach to content analysis: developing methods that identify language-grounded visual patterns in AI-generated content.

Here we introduce Language-Grounded Sparse Encoders (LanSE), a content analysis method that systematically identifies interpretable visual patterns in AI-generated content and grounds them with natural language descriptions (See Figure~\ref{fig:lanse}). LanSE leverages recent advances in mechanistic interpretability \citep{templeton2024scaling, lieberum2024gemmascopeopensparse}, particularly sparse autoencoders \citep{cunningham2023sparseautoencodershighlyinterpretable, bussmann2025learningmultilevelfeaturesmatryoshka, gao2024scalingevaluatingsparseautoencoders} and transcoders \citep{dunefsky2024transcodersinterpretablellmfeature, paulo2025transcodersbeatsparseautoencoders,lindsey2024sparse}, combined with large multimodal models (LMM) analysis to comprehensively discover, interpret and organize visual patterns across multiple content aspects. We validate LanSE's content analysis capability through extensive human evaluation involving over 11,000 annotations from six independent annotators. This evaluation demonstrates strong alignment with human judgment, achieving 93\% agreement when detecting visual patterns in natural images. This robust validation establishes LanSE as a reliable foundation for systematic content analysis across diverse applications.

LanSE's core capability to comprehensively identify visual patterns enables three complementary applications in content analysis for generative AI. For generative model evaluation, LanSE enables systematic assessment through four evaluation metrics: prompt match (whether concepts in prompts exist in the output \citep{huang2025t2icompbenchenhancedcomprehensivebenchmark}), visual realism (whether the output is photorealistic \citep{kynkäänniemi2019improvedprecisionrecallmetric}), physical plausibility (whether the output adheres to physics constraints \citep{yang2018physicsinformeddeepgenerativemodels}), and content diversity (whether outputs show sufficient variation \citep{gerstgrasser2024modelcollapseinevitablebreaking}). LanSE enables systematic automated evaluation of physical plausibility in AI-generated content and reveals failure modes that existing methods overlook. For content understanding, LanSE provides interpretable multi-label annotations that enable natural language-based image retrieval, systematic dataset curation, and granular failure mode detection across thousands of visual patterns. Beyond single-model analysis, LanSE enables inter-model correlation analysis that reveals how different generative systems encode visual knowledge, showing that models converge on semantic representations while diverging in their specific failure modes. For cross-domain analysis, the core methodology extends naturally beyond natural images: we demonstrate clinical pattern detection in medical images achieving 74\% agreement with expert radiologists, with potential applications spanning satellite imagery analysis, biological microscopy analysis, and scientific data interpretation. We believe LanSE represents a critical step toward transparent and accountable AI systems, addresses the content analysis need for this generative AI era, and paves the way to interpretable AI architectures where human understanding guides system behavior.

\section*{Results}
\label{results}
LanSE identifies 5,309 reliable visual patterns across natural images, validated through 11,160 human assessments achieving 93\% agreement on pattern-image correspondence. We extend LanSE to medical imaging, building CXR-LanSE that identifies 899 diagnosis-relevant patterns with 74\% alignment with expert radiologists. For individual content analysis, LanSE decomposes images into interpretable visual patterns, enabling fine-grained understanding of content properties and systematic dataset characterization. For generative model analysis, we design four diagnostic metrics that enable decomposed analysis of critical aspects: prompt match, visual realism, physical plausibility, and content diversity. Using these metrics, we evaluate eight state-of-the-art generative models and reveal performance characteristics that existing metrics fail to capture. Furthermore, we conduct inter-model correlation analysis revealing that generative models converge in semantics but diverge in failure modes.

\subsection*{LanSE Identifies 5,309 Visual Patterns with 93\% Accuracy}
\label{results:LanSE}
LanSE systematically discovers visual patterns by collecting millions of candidate neurons from interpretability modules, using LMMs to describe the visual patterns each neuron represents, and retaining only those with accuracies greater than 80\% (detailed in Method). This process yields 5,309 high-quality neurons, each corresponding to a distinct visual pattern found in AI-generated images. These patterns are grouped into three main categories with nine subgroups: semantic content—(1) human, (2) animal, (3) object, (4) activity, (5) environment; unrealistic patterns—(6) style, (7) artifact; and physics violations—(8) distortion, (9) structure. Each category plays a specific role in LanSE’s evaluation method (examples in Appendix~\ref{appendix:example_neurons}).

\begin{figure}[h]
    \centering
    \includegraphics[width=\linewidth]{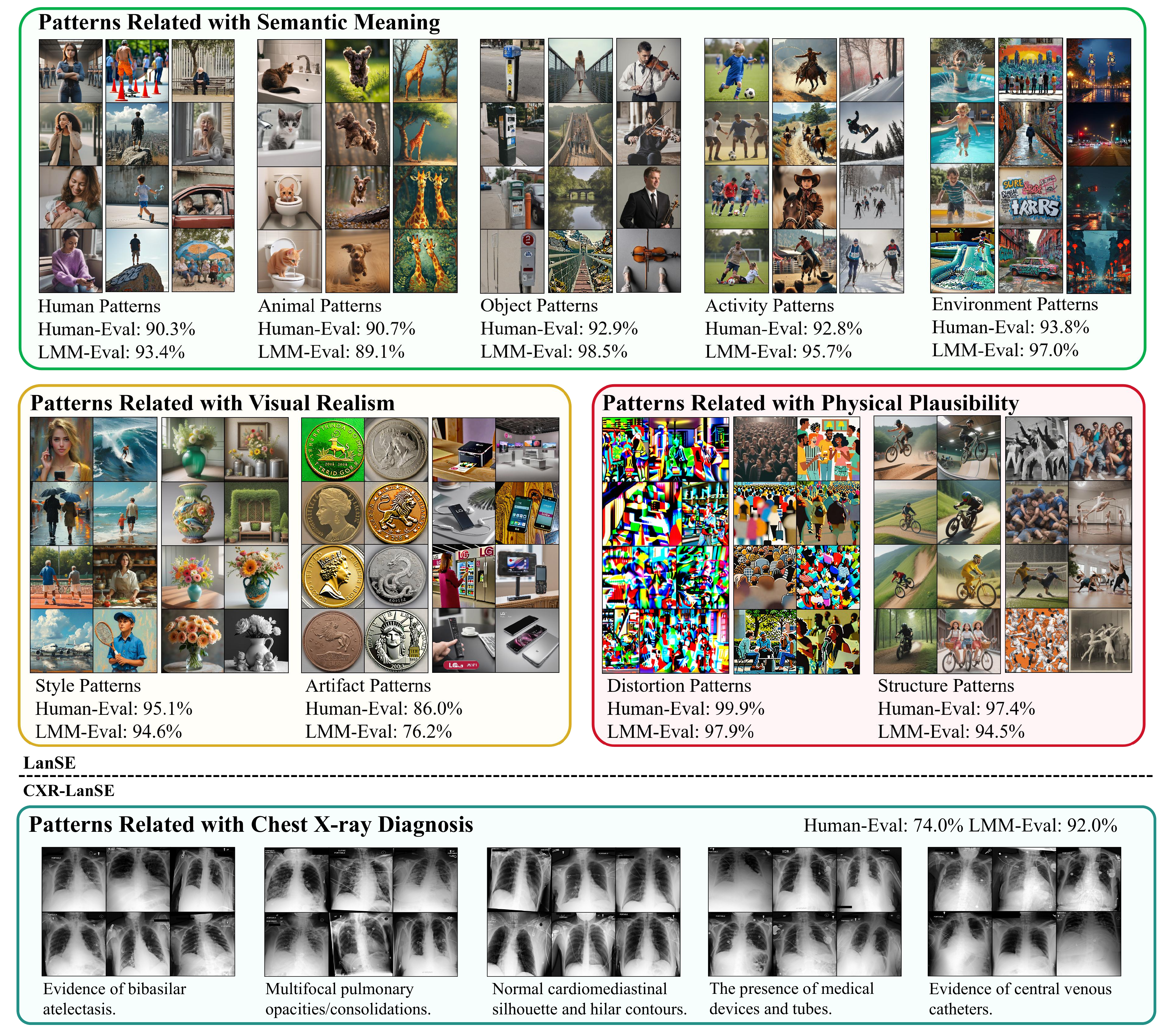}
    \caption{\textbf{Visual patterns identified by LanSE for natural and medical images.} A total of 5,309 different visual patterns in natural images and 899 visual patterns in medical images are automatically identified by LanSE. More than 11,000 human annotations from in total six independent human annotators, along with 18,000 annotations from two LMMs are collected to determine the accuracy of these visual patterns.}
    \label{fig:lanse_features}
\end{figure}

To understand if the visual patterns detected by LanSE are both accurate and reliable for evaluating image quality, we validated them against large-scale human and automated assessments. We conducted evaluations involving over 11,160 human annotations from four independent annotators (See Appendix~\ref{appendix:human_evaluation}). Each human annotator reviewed 2,790 neuron–image pairs, where each pair comprised a neuron’s interpreted description and an image that activated it, and judged whether the activation matched the described pattern. As shown in Figure~\ref{fig:lanse_features}, visual patterns discovered by LanSE achieved 93.2\% accuracy under human evaluation, with category-specific performance ranging from 86.0\% to 99.9\%. The four annotators achieved an inter-annotator agreement of 85.93\%. In addition to human annotation, we also included automated assessments from two state-of-the-art LMMs. On the same dataset, automated evaluation via LMM shows strong agreement with human annotation achieving 90.4\% accuracy with category-specific performance ranging from 76.2\% to 98.5\%. 

\subsection*{CXR-LanSE Detects 899 Clinical Relevant Visual Patterns}
\label{results:medical_extension}
CXR-LanSE extends our content analysis approach to medical imaging by systematically discovering clinical visual patterns in chest X-ray images. Using the same interpretability framework adapted for medical data, CXR-LanSE identifies 899 distinct radiological patterns from MIMIC-CXR datasets and AI-generated medical images. These patterns capture clinically meaningful features ranging from anatomical landmarks to pathological findings and medical devices. To validate the clinical relevance of discovered patterns, we conducted evaluations with two expert radiologists who assessed pattern-image correspondence on randomly selected neuron-image pairs. Each pair consisted of a CXR-LanSE pattern description and a chest X-ray image that activated the corresponding neuron. The radiologists judged whether the automated interpretation accurately captured clinically meaningful features present in the image.

CXR-LanSE achieved 74\% agreement with expert radiologists, with representative patterns including "evidence for bibasilar atelectasis," "evidence for pulmonary opacities," and "evidence for central venous catheters." This validation demonstrates CXR-LanSE's capability to automatically identify clinically relevant visual patterns without requiring manual feature engineering or domain-specific training beyond the underlying multimodal representations. The systematic discovery of interpretable medical patterns enables applications in clinical decision support, automated radiology quality assessment, and AI-generated medical image validation, addressing critical needs for explainable AI in healthcare settings.

\subsection*{LanSE Enables Pattern-based Datapoint Analysis}
\label{results:datapoint_analysis}

\begin{figure}[h]
    \centering
    \includegraphics[width=\linewidth]{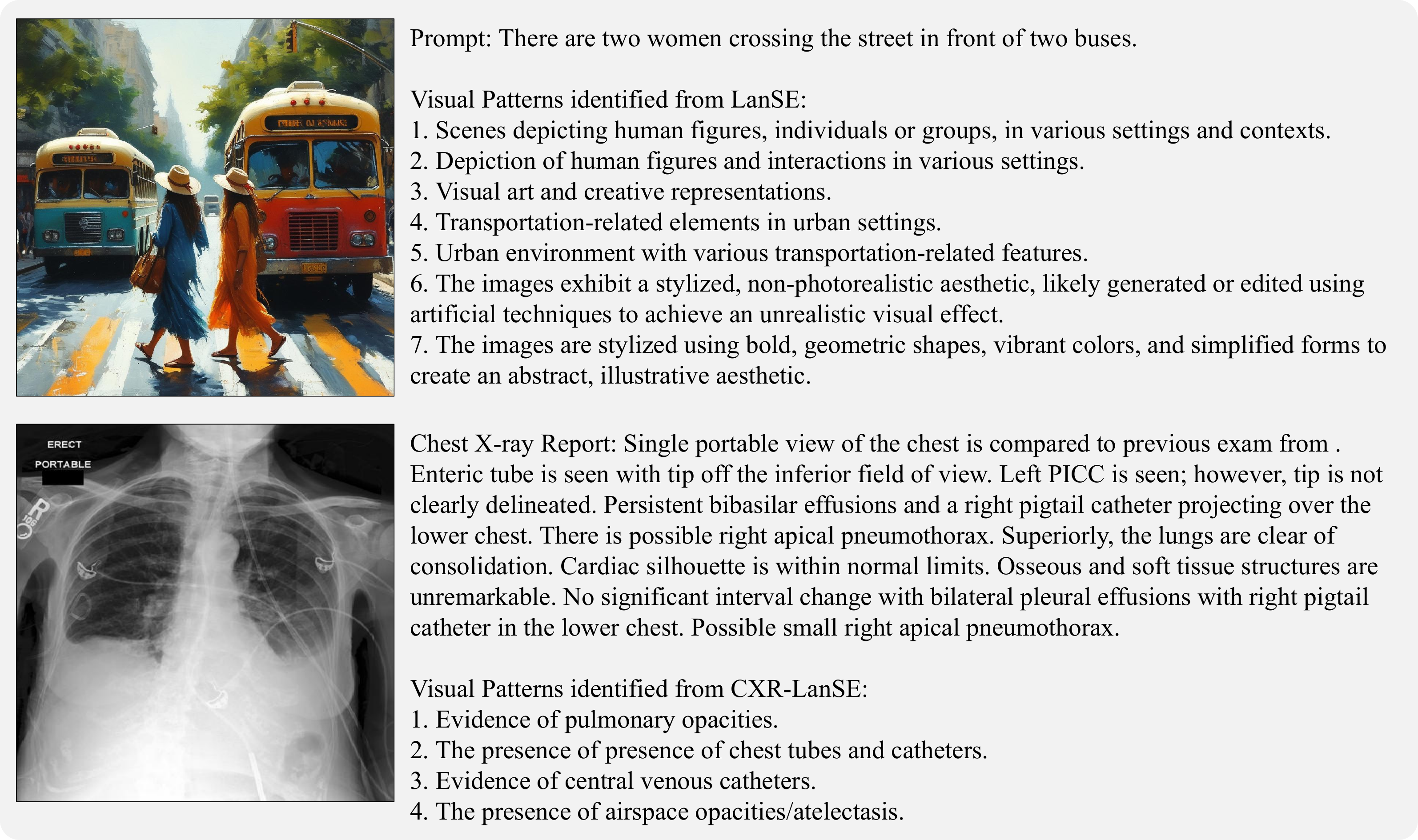}
    \caption{\textbf{Multi-label pattern detection enables detailed image analysis across domains.} LanSE automatically identifies all activated visual patterns in generated images (top) and medical images (bottom), providing comprehensive semantic fingerprints that characterize both content and potential errors. This granular annotation supports applications from dataset curation to targeted model debugging.}
    \label{fig:image_example}
\end{figure}

LanSE provides detailed multi-label annotations for individual images by identifying all activated visual patterns, enabling comprehensive content characterization at the datapoint level as demonstrated in Figure~\ref{fig:image_example}. Across our test set of 48,000 generated images, we observe that images typically activate 15-30 of the 5,309 LanSE neurons (see Appendix~\ref{appendix:ablation}). Individual images reveal complex pattern combinations—for instance, simultaneously activating neurons for "depiction of human figures" (human category), "transportation-related elements in urban settings" (environment category), and "anatomically impossible structures" (structure category), providing comprehensive characterization of image samples (more examples in Appendix~\ref{appendix:example_images}).

This multi-labeling capability enables transformative applications across domains. First, LanSE enables natural language-based image retrieval, allowing users to search datasets using semantic queries that combine multiple visual patterns. Second, integrating LanSE into training loops could enable pattern-aware optimization, where models receive granular feedback about specific visual patterns they fail to generate correctly. Third, CXR-LanSE enables clinical interpretability by automatically detecting and describing radiological features in natural language.

\subsection*{LanSE Provides Diagnostic Metrics for Model Analysis}
\label{results:metrics}
\begin{figure}[h]
    \centering
    \includegraphics[width=\linewidth]{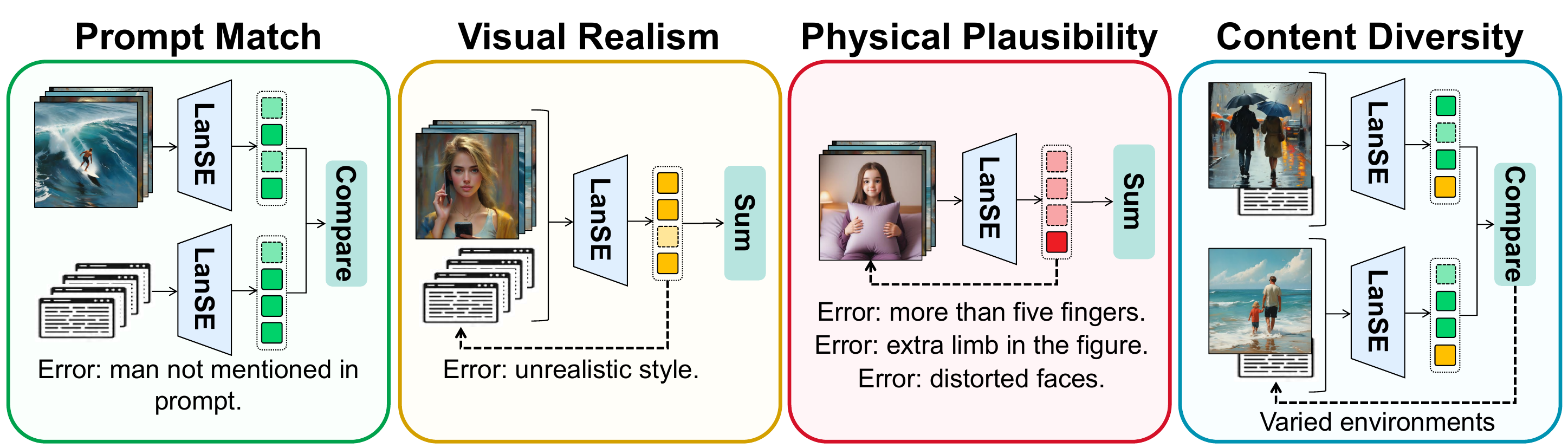}
    \caption{\textbf{Image diagnostic metrics derived from LanSE visual patterns.} Utilizing visual patterns obtained using LanSE, we define four diagnostic metrics for AI-generated natural images, each measuring a distinct aspect of the generative model.}
    \label{fig:metrics}
\end{figure}

LanSE's comprehensive visual pattern vocabulary enables the construction of four diagnostic metrics that systematically characterize different aspects of generated content: \textbf{prompt match}, \textbf{visual realism}, \textbf{physical plausibility}, and \textbf{content diversity} (detailed definition in Appendix~\ref{appendix:metrics}). These metrics target complementary aspects that various existing metrics fail to distinguish as shown in Figure~\ref{fig:metrics}.

Formally, let \(\mathrm{LanSE}(x,y)=\boldsymbol{n}=[n_1,\ldots,n_d]\in\{0,1\}^d\) denote the sparse activation vector for an image–text pair \((x,y)\). Each neuron \(n_i\) is associated with a natural-language explanation \(\epsilon_i\) (e.g., “horses and equestrian-related activities”). The \(d\) neurons are partitioned into 3 disjoint groups (\(\mathcal{C}=\{\mathbf{prompt}, \mathbf{real}, \mathbf{phy}\}\)). For \(g\in\mathcal{C}\), we write \(\mathrm{N}_g(x,y)\in\{0,1\}^{d_g}\) for the subvector of activations in group \(g\).

Prompt mismatches occur when models fail to faithfully generate outputs based on the prompts. This can stem from hallucinated or omitted elements \citep{lim2025evaluatingimagehallucinationtexttoimage}. Images with severe prompt mismatches are often undesirable and unsuitable for model fine-tuning \citep{kondapaneni2024textimagealignmentdiffusionbasedperception}. To measure this, we introduce the prompt match metric, which quantifies the divergence between neuron activations prompted by visual and textual modalities: $\mathcal{M}_{\mathbf{prompt}}(\mathcal{X}, \mathcal{Y}) = \mathbb{E}_{\mathbf{D}}||\mathrm{N}_{\mathbf{prompt}}^{\mathbf{v}}(\mathbf{x})\oplus \mathrm{N}_{\mathbf{prompt}}^{\mathbf{t}}(\mathbf{y})||_1$, where $\mathrm{N}_{\mathbf{prompt}}^{v}(x)$ and $\mathrm{N}_{\mathbf{prompt}}^{t}(y)$ are the activations for single-modality LanSEs onto the same latent space.

Visually unrealistic outputs occur when generative models produce patterns that make images photographically implausible. These can include intentional stylistic effects, such as a watercolor style \citep{inoue2018crossdomainweaklysupervisedobjectdetection}, as well as unintended artifacts, such as repetitive prompt-induced patterns that deviate from real-world conventions \citep{ramesh2022hierarchicaltextconditionalimagegeneration}. To quantify such deviations, we propose the visual realism metric, which measures the activation of neurons associated with both styles and artifacts: $\mathcal{M}_{\mathbf{real}}(\mathcal{X},\mathcal{Y}) = \mathbb{E}_{\mathbf{D}}||\mathrm{N}_{\mathbf{real}}(\mathbf{x},\mathbf{y})||_1$.

Physical implausibility refers to error patterns in generated images that violate physical principles. These include distorted surfaces and anatomically impossible structures, such as hands with more than five fingers \citep{Narasimhaswamy_2024, kwon2024graspdiffusionsynthesizingrealisticwholebody}. Although often subtle, such errors represent critical failure modes that directly undermine the practical usability of generated images. To capture these violations, we define the physical plausibility metric, which quantifies the presence of such errors: $\mathcal{M}_{\mathbf{phy}}(\mathcal{X},\mathcal{Y}) = \mathbb{E}_{\mathbf{D}}||\mathrm{N}_{\mathbf{phy}}^{\mathbf{v}}(\mathbf{x})||_1$.

Content diversity captures the degree of variation in outputs produced by a generative model \citep{lagzian2025multinoveltyimprovediversitynovelty}. Recent research has emphasized the importance of models being creative and generating diverse content \citep{baer2014creativity, chakrabarty2024artartificelargelanguage}. We consider semantic meaning neurons (e.g., human, animal, object, activity, environment) and style neurons as content-relevant neurons and compute the content diversity metric using the normalized divergence between two samples:
$\mathcal{M}_{\mathbf{con}}(\mathcal{X},\mathcal{Y})  = \mathbb{E}_{\mathbf{D}}\frac{||\mathrm{N}_{\mathbf{con}}(\mathbf{x_i},\mathbf{y_i}) \oplus \mathrm{N}_{\mathbf{con}}(\mathbf{x_j},\mathbf{y_j})||_1}{||\mathrm{N}_{\mathbf{con}}(\mathbf{x_i},\mathbf{y_i})||_1\cdot ||\mathrm{N}_{\mathbf{con}}(\mathbf{x_j},\mathbf{y_j})||_1}$.

\subsection*{Diagnostic Metrics Align with Human Quality Judgments}

To assess whether the diagnostic metrics automatically derived from LanSE validly capture quality distinctions in AI-generated images, we performed meta-evaluations comparing each metric against human judgments. For the first three metrics—prompt match, visual realism, and physical plausibility—we construct dedicated validation sets and label them using human annotators. Human evaluation of content diversity requires nuanced expertise to assess differences across output distributions and was therefore excluded from this validation.

\begin{figure}[h]
    \centering
    \includegraphics[width=1\linewidth]{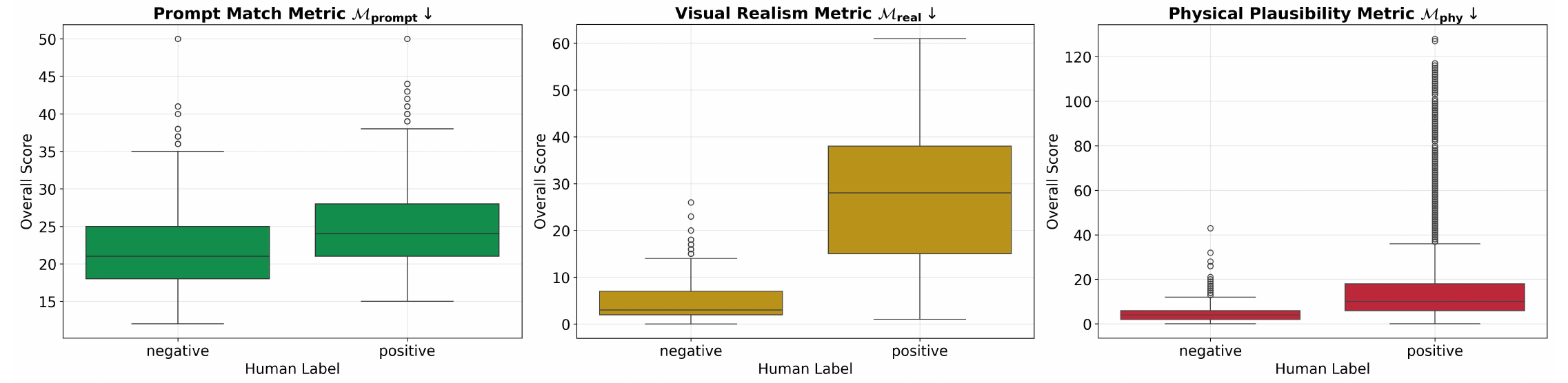}
    \caption{\textbf{Validation of LanSE metrics against human judgments.} Comparison of average metric values between positive (error-present) and negative (error-free) image sets as classified by human annotators and LMMs. Shown here for visual realism metric across natural images, generated images, and subsets classified by error type. Significant separation between positive and negative sets validates metric effectiveness.}
    \label{fig:meta-evaluation}
\end{figure}

The validation protocol was as follows: from a pool of 3,000 generated images, two independent annotators classified each image for each metric category as positive (containing the corresponding error type) or negative (error-free). An image was assigned to the positive set if the annotator indicated the presence of relevant error. We then compute the average metric scores for the positive and negative sets to assess discriminative power. Our results shown in Figure~\ref{fig:meta-evaluation} demonstrate a strong alignment between LanSE metrics and human judgments. Across the three metrics, the images in the negative sets show significantly lower metric values than those in the positive sets.

\subsection*{Fine-grained Evaluation of Eight Prevalent Generative Models}
\label{results:benchmark}

Utilizing our LanSE model and derived diagnostic metrics, we evaluate the quality of generated images produced by eight generative models, including four variants from the Stable Diffusion family, SDXL-turbo, SDXL-base, SDXL-medium, and SDXL-large \citep{meng2022sdeditguidedimagesynthesis}, as well as DALL$\cdot$E 3 \citep{dalle3}, FLUX \citep{FLUX.1-dev}, Kolors \citep{Kolors-diffusers}, and Stable-Cascade \citep{stablecascade2025}.

\begin{figure}[h]
    \centering
    \includegraphics[width=\linewidth]{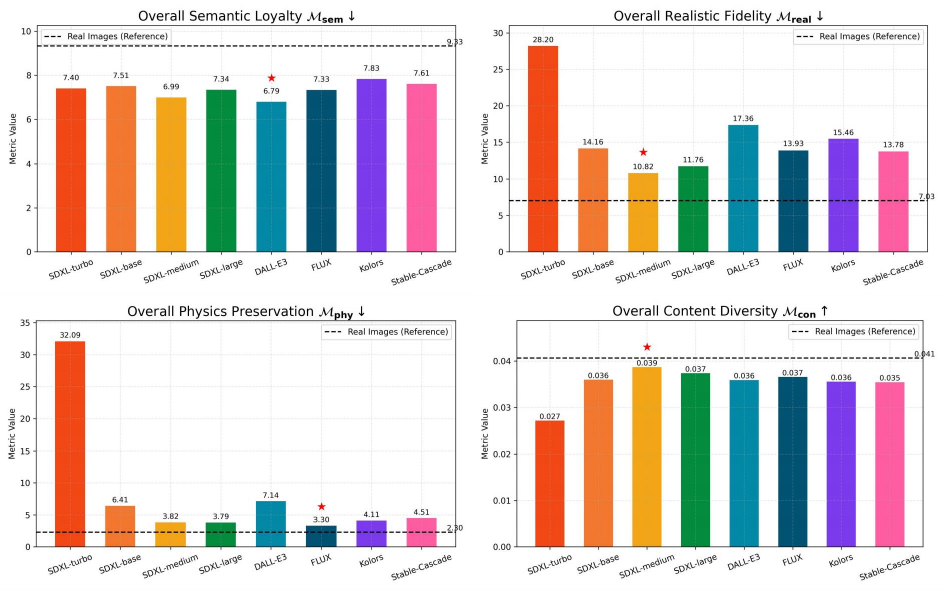}
    \caption{\textbf{Evaluating generative machine learning models with LanSE-derived diagnostic metrics.} Performance comparison of eight generative machine learning models across four LanSE-derived diagnostic metrics: prompt match, visual realism, physical plausibility, and content diversity. Lower scores indicate better performance for the first three metrics, while higher scores indicate better content diversity.}
    \label{fig:benchmark}
\end{figure}

\begin{figure}[H]
    \centering
    \includegraphics[width=\linewidth]{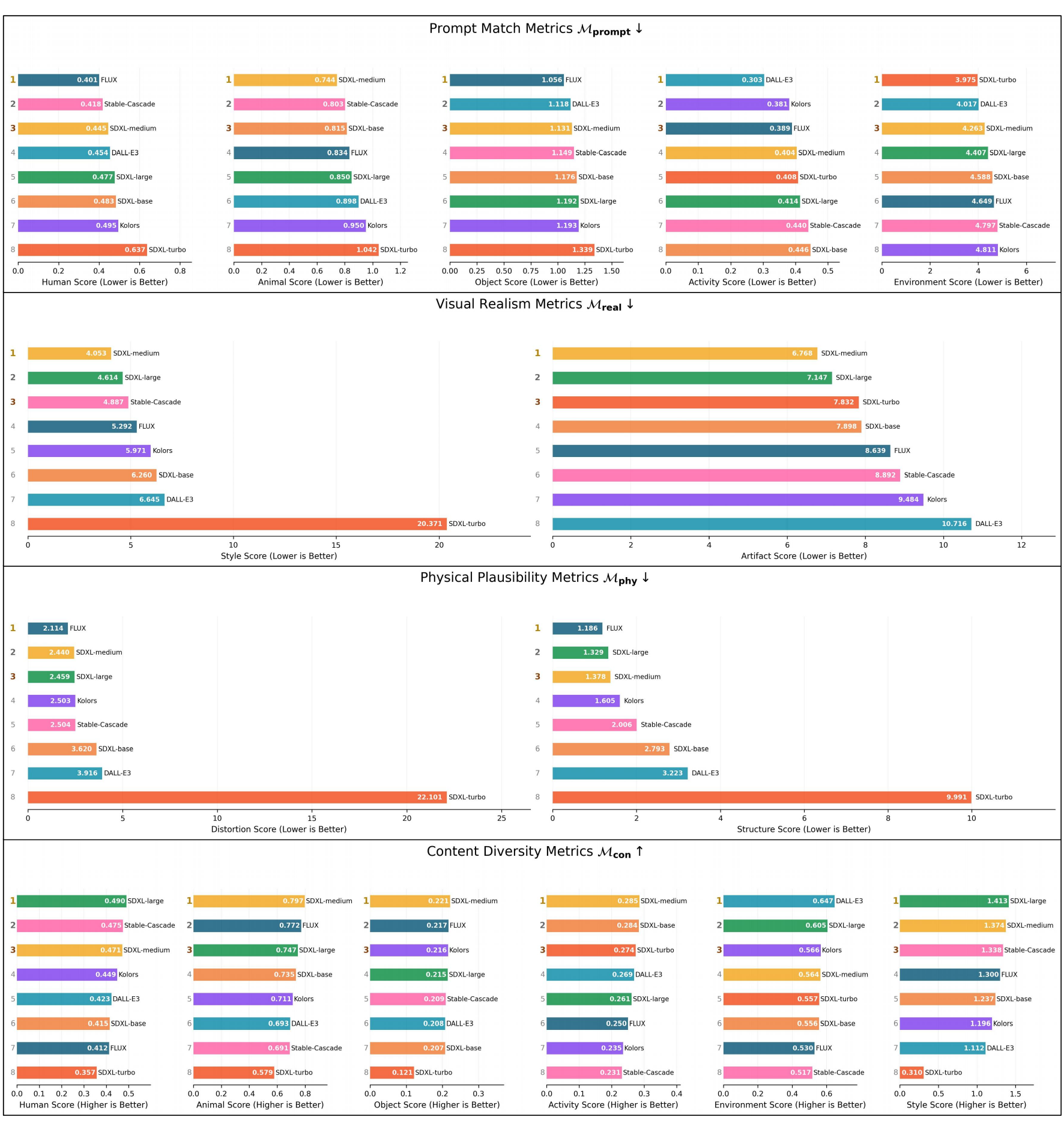}
    \caption{\textbf{Model leaderboards for subgroup scores of LanSE-derived diagnostic metrics.} Different generative models exhibit distinct performance profiles across evaluation dimensions. For prompt match, environmental elements represent the most common source of semantic misalignment across all models. FLUX achieves superior performance in human and object semantics, while SDXL-medium excels at animal semantics, DALL·E3 at activity semantics, and SDXL-turbo at environmental semantics. In visual realism metrics, SDXL-medium demonstrates the strongest performance in avoiding both unmentioned art styles and invented visual artifacts, with SDXL-large as the second-best performer. Physical plausibility metrics reveal FLUX as the leading model in preventing both surface distortions and anatomically implausible structures, followed by SDXL-medium and SDXL-large. Content diversity analysis shows complementary strengths: SDXL-large produces the most varied human representations, SDXL-medium generates the greatest diversity in animals, objects, and activities, while DALL·E3 the highest in environment and SDXL-large the highest in style.}
    \label{fig:benchmark_detail}
\end{figure}

Our results presented in Figure~\ref{fig:benchmark} reveal distinct performance profiles for each model. FLUX achieves the highest physical plausibility scores, making it particularly suitable for applications requiring strict physical plausibility. SDXL-medium excels in both photorealism and content diversity, producing the most realistic and varied outputs. Figure~\ref{fig:benchmark_detail} presents detailed group-wise performance metrics that reveal more nuanced characteristics. These fine-grained insights provide practitioners with actionable guidance for choosing models based on their application requirements, whether prioritizing prompt match, visual realism, physical plausibility or content diversity.

In the evaluation of recent generative models, our results reveal a striking pattern: while they achieve strong prompt match, often exceeding that of real-world datasets, they consistently struggle with visual realism, physical plausibility, and content diversity. Recent models frequently exhibit distinctive stylistic signatures that deviate from photorealistic representation, generate distorted surfaces, and produce physically implausible structures. Their content diversity remains significantly below that of natural image distributions. These findings suggest the machine learning community should recalibrate its priorities, shifting focus from prompt match, a largely solved problem, toward these more nuanced aspects of visual generation.

\subsection*{LanSE Enables Inter-model Correlation Analysis}
\label{results:correlation}

LanSE's visual patterns enable correlation analysis that reveals fundamental similarities and differences between generative models. We generate 6,000 image-caption pairs for each model and compute pairwise correlation matrices across all neuron groups.

\begin{figure}[h]
    \centering
    \includegraphics[width=0.94\linewidth]{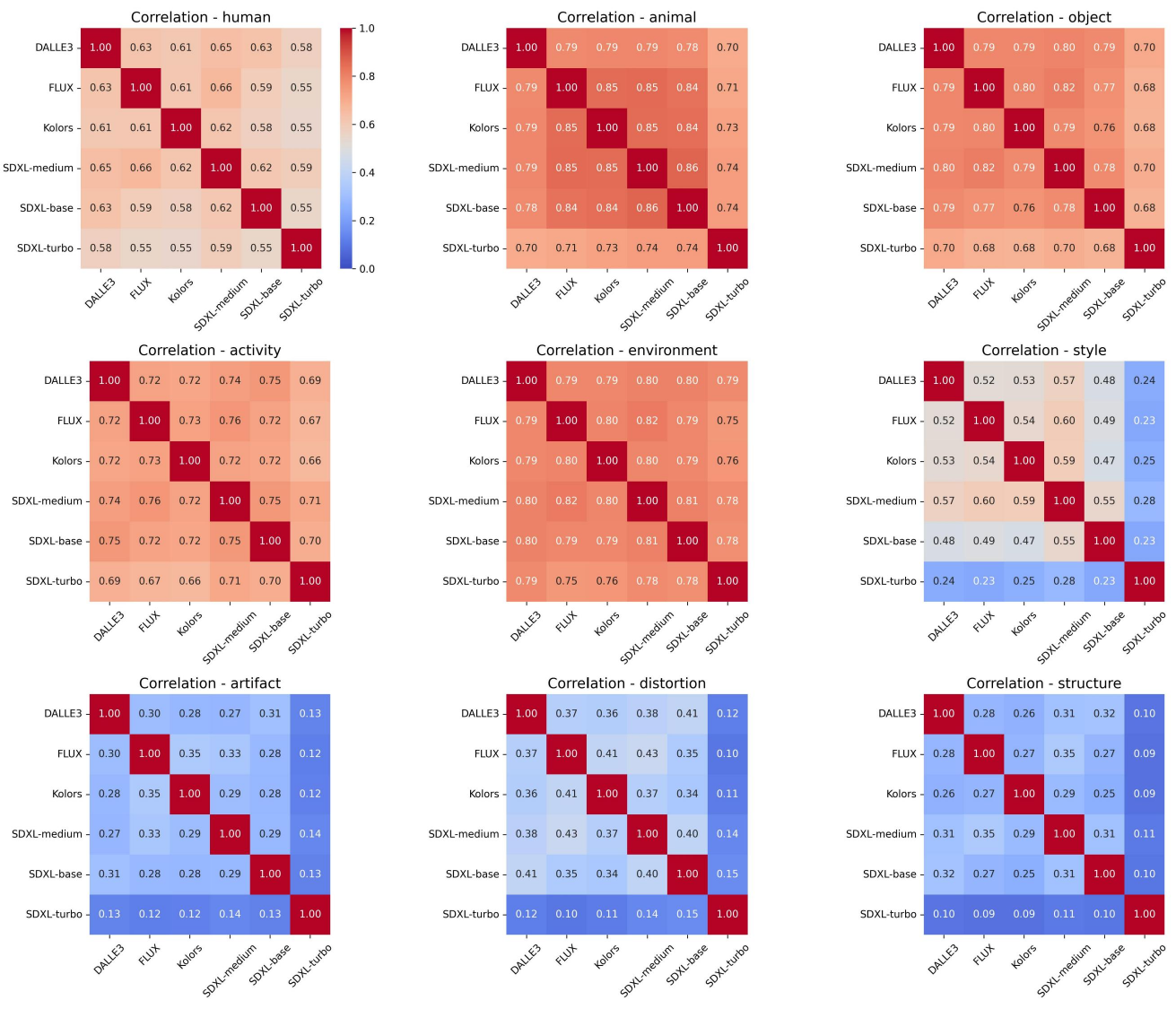}
    \caption{\textbf{Generative models share semantic understanding but produce unique errors.} Pairwise correlation matrices of LanSE visual patterns across six models reveal high inter-model agreement for content-related neurons but substantial divergence in error-detecting neurons. Warmer colors indicate higher correlations.}
    \label{fig:correlation}
\end{figure}

Figure~\ref{fig:correlation} reveals two striking patterns in model behavior. First, content neurons, covering human, animal, object, activity, environment, and style, exhibit high inter-model correlations (mean $r=0.74$), whereas error-signaling neurons, artifact, distortion, and structure, show markedly lower correlations (mean $r=0.31$). This demonstrates that while models converge on semantic representations, they develop distinct failure modes. Second, specific model pairs show notable similarities: FLUX and SDXL-medium display particularly high correlations in style ($r=0.60$), environment ($r=0.82$), and activity ($r=0.76$) neurons, suggesting shared architectural or training strategies. Conversely, SDXL-turbo exhibits the most distinctive activation patterns, with the lowest average style correlation ($r=0.51$).

\subsection*{LanSE Reveals Model Differences Missed by Prevalent Methods}
\label{results:comparison}
To demonstrate LanSE's advantages for content analysis, we compare our diagnostic capabilities with widely-adopted baseline methods: Fréchet Inception Distance (FID) \citep{jayasumana2024rethinkingfidbetterevaluation}, Inception Score (IS) \citep{barratt2018noteinceptionscore}, and CLIP score \citep{chen2025contrastivelocalizedlanguageimagepretraining}. 

\begin{figure}[h]
\centering
\includegraphics[width=0.99\linewidth]{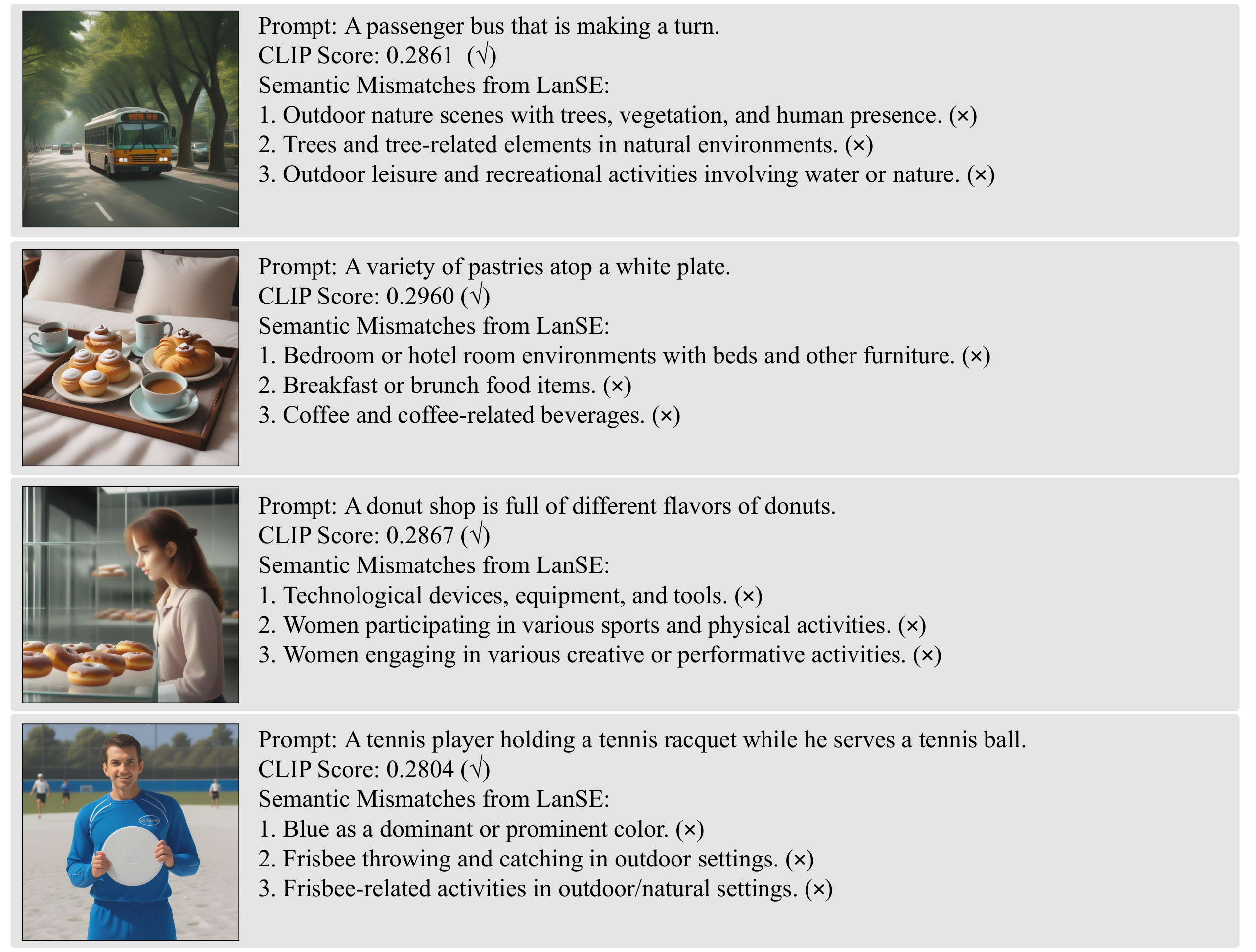}
\caption{\textbf{CLIP score's blind spots revealed through LanSE's fine-grained analysis.} Examples of images having high CLIP scores ($>0.28$) with obvious semantic mismatches. LanSE successfully detects mismatches that CLIP overlooks.}
\label{fig:misidentification}
\end{figure}

\begin{figure}[h]
\centering
\includegraphics[width=\linewidth]{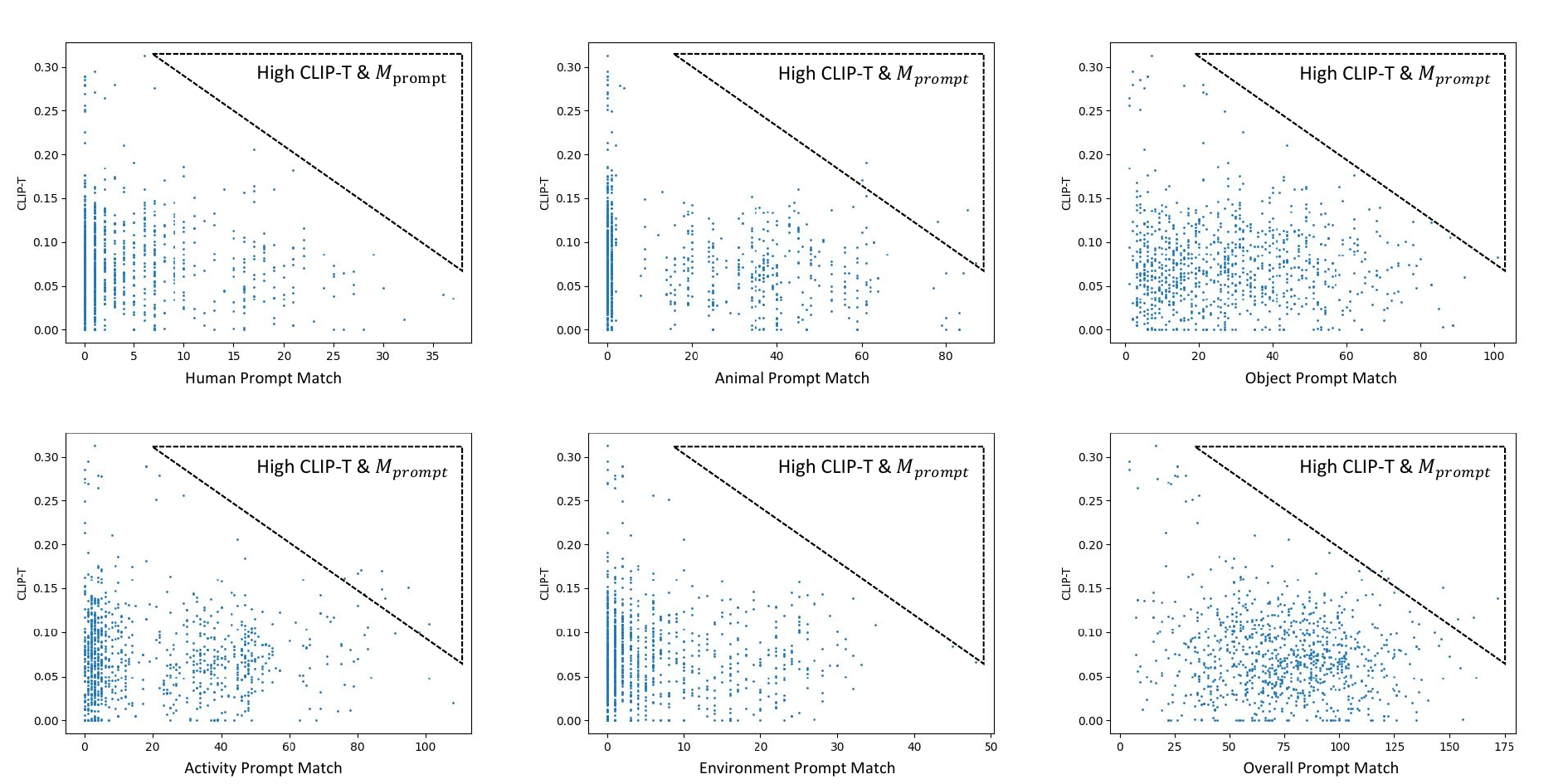}
\caption{\textbf{LanSE enables group-wise semantic mismatch detection beyond CLIP score.} Scatter plots comparing CLIP scores with LanSE's prompt match metric ($\mathcal{M}_{prompt}$) show agreement on severe misalignments (empty upper-right) but reveal LanSE's unique ability to identify specific semantic failures—distinguishing whether mismatches occur in human, animal, object, activity, or environment categories. }
\label{fig:semantic_vs_clip}
\end{figure}

Our analysis aligns with CLIP's systematic biases identified in various previous works \citep{schrodi2025effectstriggermodalitygap}. While both CLIP and our prompt match metric generally detect severe misalignments (empty upper-right regions in Figure~\ref{fig:semantic_vs_clip}), CLIP can assign high scores to outputs with obvious semantic mismatches when partial object alignment masks compositional errors, as shown in Figure~\ref{fig:misidentification}. In contrast, LanSE's prompt match evaluation is based on visual patterns and can capture subtle semantic mismatches CLIP fails to detect.

\begin{figure}[H]
\centering
\includegraphics[width=\linewidth]{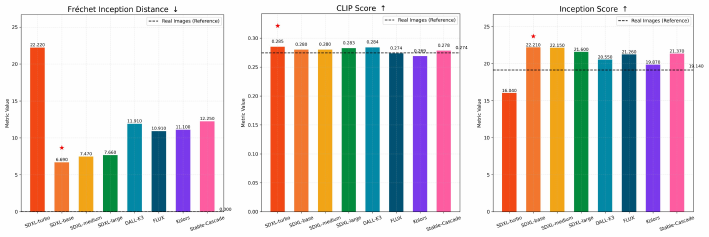}
\caption{\textbf{LanSE reveals model differences that traditional metrics fail to capture.} Evaluation results across traditional metrics (FID, CLIP, IS) show limited discriminative power and occasionally misalign with human preferences. In contrast, LanSE-derived diagnostic metrics captures these differences as shown in Figure \ref{fig:benchmark}.}
\label{fig:metric_comparison}
\end{figure}

Figure~\ref{fig:metric_comparison} further demonstrates two critical limitations of these prevalently used metrics. First, they show minimal discriminative power—CLIP scores cluster tightly (0.269-0.285) despite substantial quality differences between models. Second, they misalign with human perception; SDXL-base achieves the best FID and Inception Score while producing notably lower-quality outputs than FLUX or SDXL-medium, suggesting these metric values poorly reflect perceptual quality.

\subsection*{LanSE Beats Prevalent LMMs in Fine-grained Error Detection}
\label{results:hallucination}
In this era of generative AI, many researchers consider LMMs with prompt engineering as a quick and universal solution for diverse problems \citep{Koco__2023,bubeck2023sparksartificialgeneralintelligence,bang2023multitaskmultilingualmultimodalevaluation,tang2024demonstrationnotebookfindingsuited}. Comparatively, LanSE can detect fine-grained visual patterns that state-of-the-art LMMs miss, while requiring significantly lower computational resources during inference. We compare LanSE's specialized error detection capabilities against six state-of-the-art vision-language models on identifying physical plausibility errors. Using 3,410 human-annotated images (2,522 with physics violations, 888 without), we tasked each system with detecting anatomical impossibilities, distorted structures, and physics violations.

\begin{figure}[h]
\centering
\includegraphics[width=\linewidth]{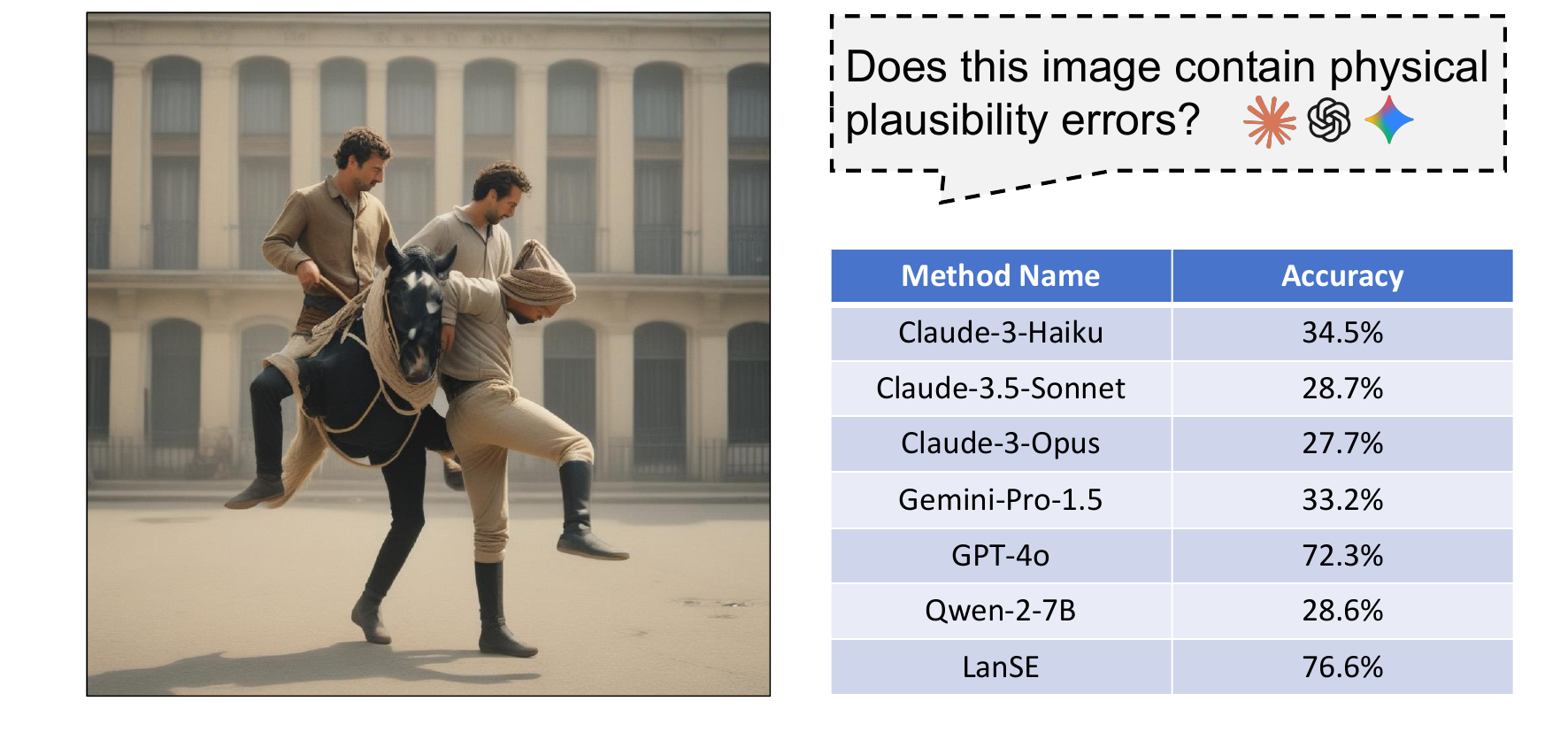}
\caption{\textbf{LanSE outperforms LMMs in detecting physical plausibility errors.} Comparison of accuracy across methods on 3,410 human-annotated images shows LanSE's specialized neurons achieve superior performance in identifying anatomical impossibilities and structural violations that LMMs consistently underperform.}
\label{fig:hallucination}
\end{figure}

As shown in Figure~\ref{fig:hallucination}, vision-language models showed limited performance on this task. While GPT-4o achieved 72.3\% accuracy, other prominent models performed poorly: Claude-3-Haiku (34.5\%), Gemini-Pro-1.5 (33.2\%), Claude-3.5-Sonnet (28.7\%), Qwen-2-7B (28.6\%), and Claude-3-Opus (27.7\%). These models consistently missed subtle anatomical errors such as extra fingers, impossible joint configurations, and localized physics violations.

LanSE achieved the best accuracy (76.6\%) using 383 specialized neurons (163 for structural errors, 220 for distortions) that explicitly encode specific failure patterns. This targeted approach demonstrates that interpretable sparse representations can capture fine-grained visual anomalies that general-purpose vision-language architectures miss, highlighting the value of domain-specific pattern detection for content analysis applications.

\section*{Discussions}
\label{discussion}

We are witnessing a profound societal transformation as AI-generated content gradually permeates nearly every domain of human content creation, fundamentally challenging how we understand, verify, and trust the materials we encounter. Research across multiple domains documents serious societal risks from AI-generated content proliferation, including homogenization of human expression \citep{suzuki2023we} and creative diversity \citep{doshi2024generative}, vulnerability to data poisoning attacks in medical systems \citep{alber2025medical}, model collapse from recursive training on synthetic data \citep{shumailov2024ai, dohmatob2024strongmodelcollapse}, contamination of scientific datasets with undetected artificial material \citep{acion2023generative}, and erosion of editorial standards in professional publishing \citep{lindsay2023llms}. These documented challenges underscore the critical need for comprehensive content analysis infrastructure that can systematically characterize AI-generated content properties, enabling informed governance and deployment of these materials across high-stakes domains. LanSE provides the systematic content analysis infrastructure this transformed landscape demands, automatically discovering thousands of interpretable visual patterns that enable comprehensive, human-understandable assessment of AI-generated content properties.

LanSE tackles these diverse challenges by combining pattern-based content analysis with distribution-level diversity assessment, enabling both granular detection of specific content characteristics and systematic measurement of variation trends. The pattern identification capability allows LanSE to detect specific synthetic signatures that indicate data contamination, identify anomalous visual patterns that suggest medical data poisoning, and recognize quality degradation markers in editorial and publishing contexts. Meanwhile, diversity assessment capabilities enable early detection of the content narrowing that precedes model collapse and systematic measurement of homogenization trends across creative outputs that threaten cultural and expressive diversity.

LanSE's systematic pattern analysis transforms how critical institutions—from hospitals to regulatory agencies—understand and govern AI-generated content. In medical diagnostics and financial decision-making, LanSE provides clinicians and analysts with language-grounded explanations of AI predictions, delivering the transparency and accountability these domains require. Regulators can customize LanSE's pipeline to detect domain-specific risks for educational, publishing, and safety-sensitive materials, enabling precise content filtering with transparent natural language justifications. For machine learning researchers, LanSE enables systematic dataset analysis by quantifying pattern distributions to reveal bias and guide targeted augmentation, while providing early warning signals for model collapse through diversity measurement \citep{10.1093/jcr/ucaf013, kazdan2025collapsethriveperilspromises}. Beyond analysis applications, LanSE enables integration into training pipelines for interpretable model feedback and supports development of pattern-based retrieval systems where users can find content sharing specific visual characteristics. 

For the mechanistic interpretability community, LanSE represents a fundamental reorientation of the field's purpose—demonstrating for the first time that interpretability modules, including sparse autoencoders and transcoders, previously confined to dissecting model circuits and scientific investigations, can be repurposed as practical infrastructure for urgent societal challenges in content verification, safety assessment, and quality control \citep{bereska2024mechanisticinterpretabilityaisafety, rai2025practicalreviewmechanisticinterpretability,sharkey2025openproblemsmechanisticinterpretability}. While the field has pursued increasingly microscopic questions about model understanding and steering \citep{chen2025personavectorsmonitoringcontrolling, turner2024steeringlanguagemodelsactivation, zou2025representationengineeringtopdownapproach,wu2025axbenchsteeringllmssimple,saini2026bridgingmechanisticinterpretabilityprompt,}, we show these same mathematical frameworks can directly address the content analysis crisis facing billions of users, establishing interpretability methods not as endpoints for understanding AI systems but as foundational tools for governing their outputs. Furthermore, constructing LanSE as one ensemble of many interpretability modules stands as a way to solve the unstability of feature sets  widely encountered in mechanistic interpretability \citep{tang2026unifiedtheorysparsedictionary}.

While LanSE demonstrates versatility across multiple applications, several avenues could strengthen and extend this method. First, expanding the scale and granularity of discovered visual patterns would enhance evaluation precision, as current assessments remain bounded by our pattern vocabulary. Second, although we successfully adapted LanSE to medical imaging through CXR-LanSE, extending the framework to additional modalities, video, 3D content, or scientific imaging, presents promising opportunities. Third, while we currently demonstrate LanSE primarily for content analysis, dataset characterization, and model correlation studies, we envision integrating LanSE directly into model training pipelines, enabling generative models to receive diverse, interpretable feedback on specific visual patterns during optimization. Fourth, developing adaptive pattern vocabularies that evolve with emerging AI capabilities could maintain LanSE's effectiveness as generative models advance. Fifth, extending LanSE toward region-level or token-level spatial localization would further enhance its actionability, enabling users to not only detect the presence of artifact or structural failures but also identify where in the image they occur — particularly valuable for debugging generative systems and dataset curation. Last, some training-free methods might be useful to help identify the interpretable features and can be integrated into LanSE's pipelines for enhanced performances (see Appendix~\ref{appendix:limitations}).

In conclusion, LanSE establishes essential infrastructure for navigating the era of ubiquitous AI-generated content. By providing systematic, interpretable analysis of visual patterns at low computational costs, LanSE transforms how society understands, governs, and deploys AI-generated material across critical domains. This pattern-based content analysis approach enables diverse institutions to make informed decisions about AI integration across applications from medical diagnostics to content moderation. As AI-generated content becomes increasingly prevalent in high-stakes applications, LanSE provides the analytical foundation necessary for navigating this transformation with deliberate oversight and control.

\section*{Method}
\label{method}

\begin{figure}[h]
    \centering
    \includegraphics[width=1\linewidth]{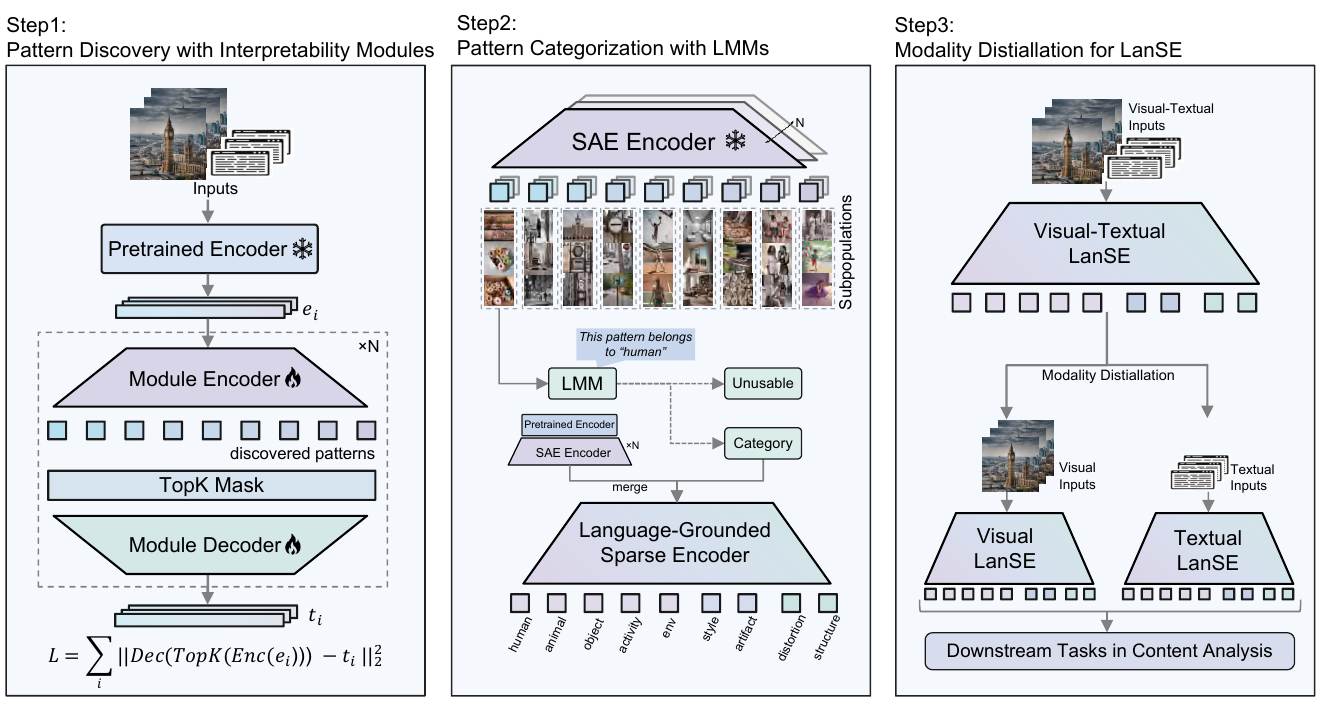}
    \caption{\textbf{Pipeline for constructing Language-Grounded Sparse Encoders.} The method proceeds through pattern discovery using sparse autoencoders and transcoders, automated categorization with LMMs, and modality distillation to enable single-input content analysis across diverse applications.}
    \label{fig:pipeline}
\end{figure}

In this section, we present LanSE's systematic content analysis pipeline (See Figure~\ref{fig:pipeline}). Our method addresses the challenge of decomposing AI-generated content into language-grounded visual patterns through a three-stage approach: (1) discovering interpretable neurons using interpretability modules (2) categorizing these neurons automatically with LMMs, and (3) building single-modality LanSEs through a modality distillation procedure. We will release our models, scripts, and human-annotated datasets at https://github.com/YimingTangible-NUS/LanSE.

\subsection*{Dataset Construction}
We use a collection of image-caption pairs from MS-COCO \citep{lin2015microsoftcococommonobjects}, Flickr8k and Flickr30k \citep{plummer2016flickr30kentitiescollectingregiontophrase}, TextCaps \citep{sidorov2020textcapsdatasetimagecaptioning}, NoCaps \citep{Agrawal_2019}, and MIMIC-CXR \citep{johnson2019mimiccxrjpglargepubliclyavailable} to build LanSE and CXR-LanSE. We augment these datasets with samples from several generative models \citep{meng2022sdeditguidedimagesynthesis, dalle3, FLUX.1-dev, Kolors-diffusers, zhang2023addingconditionalcontroltexttoimage, malik2023stable} by prompting them with captions from the original collections. We also employ human data annotators to identify erroneous images from the dataset for targeted discovery of physical plausibility patterns (detailed in Appendix~\ref{appendix:method}).

\subsection*{Visual Pattern Discovery with Interpretability Modules} 
To systematically discover interpretable visual patterns for content analysis, we leverage recent advances in mechanistic interpretability \citep{cunningham2023sparseautoencodershighlyinterpretable, bussmann2025learningmultilevelfeaturesmatryoshka, gao2024scalingevaluatingsparseautoencoders, dunefsky2024transcodersinterpretablellmfeature, paulo2025transcodersbeatsparseautoencoders,lindsey2024sparse}. Studies have shown that enforcing sparsity during neural network training produces monosemantic neurons—individual units that respond to specific, interpretable concepts \citep{makhzani2014ksparseautoencoders, farrell2024applyingsparseautoencodersunlearn, elhage2022toymodelssuperposition, bricken2023monosemanticity,zhao2026rep2textdecodingtextsingle}, motivating us to adapt these techniques for content analysis. 

In this work, we employ sparse autoencoders and transcoders to discover visual patterns that are interpretable in natural language, enabling systematic content analysis. We first encode image-caption pairs into multimodal embeddings by concatenating image and text representations, $\mathbf{e_i} = [\mathrm{E_x}(\mathbf{x_i}),\ \mathrm{E_y}(\mathbf{y_i})]$, then utilize these embeddings to train the pattern discovery modules. 

\textbf{Sparse autoencoders} are sparsity-driven interpretability modules that can comprehensively discover a diverse set of general visual patterns through the reconstruction of one model embedding. We develop multiple sparse autoencoders, each comprising a single-layer encoder $\mathbf{z_i} = \mathrm{Enc}(\mathbf{e_i})=\mathrm{ReLU}(\mathbf{w_{enc}}\cdot \mathbf{e_i} +\mathbf{b_{enc}})$ and a single-layer decoder $\mathrm{Dec}(\mathbf{z_i}) = \mathrm{ReLU}(\mathbf{w_{dec}}\cdot \mathbf{z_i} +\mathbf{b_{dec}})$. A top-$k$ activation scheme \citep{gao2024scalingevaluatingsparseautoencoders} is used to encourage sparsity in the latent space: for each $\mathbf{z_i}$, we retain only the $k$ highest-magnitude activations and set the rest to zero. The training objective is as follows:
\begin{equation}
\mathcal{L}_{\text{sae}} = \mathbb{E}_{(\mathbf{x_i}, \mathbf{y_i})}\left[\left\|\mathrm{Dec}(\mathrm{TopK}(\mathrm{Enc}(\mathbf{e_i}))) - \mathbf{e_i}\right\|_2^2\right].
\end{equation}

\textbf{Transcoders} are sparsity-driven interpretability modules that enable targeted discovery of specific pattern groups by learning sparse mappings from task-specific classifier representations to target outputs. We train transcoders to predict the hidden representations of lightweight classifiers designed to detect specific content properties (e.g., physics violations). This targeted approach complements sparse autoencoders by discovering rare but critical visual patterns that might be underrepresented in general reconstruction-based discovery. The training objective follows:
\begin{equation}
\mathcal{L}_{\text{transcoder}} = \mathbb{E}_{(\mathbf{x_i}, \mathbf{y_i})}\left[\left\|\mathrm{Dec}(\mathrm{TopK}(\mathrm{Enc}(\mathbf{e_i}))) - \mathbf{t_i}\right\|_2^2\right].
\end{equation}
where $\mathbf{t_i}$ represents classifier hidden activations that contain task-relevant information.

The neurons discovered by these interpretability modules are initially raw and noisy, requiring systematic curation to identify reliably interpretable patterns. To maximize pattern discovery coverage while enabling quality filtering, we train multiple modules on different data subsets. In total, we train 200 sparse autoencoders and 50 transcoders for LanSE, plus 100 sparse autoencoders for CXR-LanSE, with each module containing 15,000 latent neurons and k=32 for the top-k sparsity constraint. From this large pool of candidate neurons, we apply systematic filtering criteria to identify those with consistent, interpretable activations. We adopt hyperparameters similar to recent developments in mechanistic interpretability \citep{sun2025high, minegishi2025rethinkingevaluationsparseautoencoders} and the approach can be scaled with additional compute and data.

\subsection*{Visual Pattern Categorization with LMMs}
After training the interpretability modules, we obtain a collection of latent neurons, each of them potentially interpretable in natural language. We then employ LMMs \citep{claudehaiku, llama3} to analyze and interpret these latent neurons.

For each neuron, we collect a set of samples that trigger strong activations for this neuron and prompt Claude-3.5-Haiku \citep{claudehaiku} (a large multimodal model) to summarize the common characteristics within these samples. The prompt we use to summarize the visual pattern in natural language is given below (more prompts in Appendix~\ref{appendix:llm_analysis}):

\begin{tcolorbox}[title=Pattern Summarization Prompt ($\text{Prompt}_{\text{sum}}$),fonttitle=\bfseries,colback=gray!5!white,colframe=gray!75!black]
You are an expert in visual pattern analysis. Given the following images and texts: \{samples\}, analyze the commonalities among them. Concisely summarize one common visual pattern that is possessed by all the instances. Wrap the commonality in square brackets.
\end{tcolorbox}

Based on the natural language summarizations of the visual patterns, we categorize these visual patterns into nine groups, $\mathbf{C}$, including five groups designed for semantics: \textbf{human}, \textbf{animal}, \textbf{object}, \textbf{activity}, and \textbf{environment}; two groups for visual realism: \textbf{style} and \textbf{artifact}; and two groups for physical plausibility: \textbf{distortion} and \textbf{structure} (See Appendix~\ref{appendix:neuron_groups}). We use the following prompt to categorize the visual pattern based on their explanations.

\begin{tcolorbox}[title=Pattern Categorization Prompt ($\text{Prompt}_{\text{cat}}$),fonttitle=\bfseries,colback=gray!5!white,colframe=gray!75!black]
We have a group of datapoints possessing one common visual pattern. Given the description of their explanation: \{explanation\}, which of the five categories does the visual pattern best belong to, human, animal, object, activity, or environment? Only output the category and wrap the answer in brackets.
\end{tcolorbox}

These visual pattern analysis procedures can be formulated as below \citep{luo2023promptengineeringlensoptimal}, noticing that the summarizations are obtained automatically during the construction of LanSEs, no assistance from LMMs are used in inference.

\begin{equation}
\epsilon_{\text{n}} = \text{LMM}(\text{Samples}| \text{Prompt}_{\text{sum}}).
\end{equation}

\begin{equation}
\text{c}_{\text{n}} = \text{LMM}(\epsilon_{\text{n}}| \text{Prompt}_{\text{cat}}) \in \mathbf{C}.
\end{equation}

\subsection*{Modality Distillation for Language-Ground Sparse Encoders}
LanSE supports two pathways for caption-free evaluation: direct training on visual-only datasets, or modality distillation from visual-textual LanSE models. For visual-textual LanSEs, we develop a modality distillation procedure as an inherent feature that enables applications to image-only datasets. 

We denote one visual-textual LanSE as $\mathrm{S_{\text{c}}}:(\mathcal{X}, \mathcal{Y})\rightarrow\mathbb{R}^{d_\text{c}}$, ($\text{c}\in \mathbf{C}$), with its language-grounded sparse layer obtained by concatenating neurons in the interpretability modules: $\mathbf{W_{\text{c}}} = [\mathbf{w_{\text{n}}}]$, $\mathbf{b_{\text{c}}} = [b_{\text{n}}]$ ($n\in N_\text{c}$).

\begin{equation}
\mathrm{S_{\text{c}}} (\mathbf{x_i}, \mathbf{y_i}) =  \mathrm{ReLU}(\mathbf{W_{\text{c}}} \cdot [\mathrm{E_x}(\mathbf{x_i}),\ \mathrm{E_y}(\mathbf{y_i})]+ \mathbf{b_{\text{c}}}).
\end{equation}

Modality distillation refers to training single-modality LanSEs, $\mathrm{S^{img}_{\text{c}}}$ and $\mathrm{S^{txt}_{\text{c}}}$, using outputs from $\mathrm{S_{\text{c}}} (\mathbf{x}, \mathbf{y})$ \citep{hinton2015distillingknowledgeneuralnetwork}. The LanSEs are initialized with pretrained encoders and tuned with parameter efficient fine-tuneing \citep{xu2023parameterefficientfinetuningmethodspretrained}. The optimization target is to minimize the distances between the outputs of the single-modality LanSEs and $\mathrm{S_{\text{c}}} (\mathbf{x}, \mathbf{y})$:

\begin{equation}
\left\{
\begin{aligned}
    \mathcal{L}_{\text{c}}^{\text{img}} &= \sum_i \left\|\mathrm{S_{\text{c}}^{img}}(\mathbf{x_i}) - \mathrm{S_{\text{c}}}(\mathbf{x_i},\mathbf{y_i})\right\|^2_2, \text{c}\in\mathbf{C}.\\
    \mathcal{L}_{\text{c}}^{\text{txt}} &= \sum_i \left\|\mathrm{S_{\text{c}}^{txt}}(\mathbf{y_i}) - \mathrm{S_{\text{c}}}(\mathbf{x_i},\mathbf{y_i})\right\|^2_2, \text{c}\in\mathbf{C}.\\
\end{aligned}
\right.
\end{equation}

This modality distillation procedure results in LanSEs that only require single-modality inputs, enabling us to apply LanSE when only images are available, and to analyze which patterns are encoded in one modality but not the other.

\newpage
\section*{Acknowledgements}
\label{acknowledgements}
We are deeply grateful for the computational resources provided by the Cognitive AI for Science lab (https://www.cogai4sci.com/).
Great thanks to the human annotators who contributed their time and expertise to this project, including Yiming Tang, Qiran Zou, Arash Lagzian, Samson Yu, Ye Zhang and Yingtao Zhu. Thanks to Xuming Ran for his valuable advice and guidance during the early beginning of the project. We are also grateful to Vaishnavh Nagarajan for his insightful feedback and constructive suggestions that helped shape the direction of this work. Finally, we acknowledge the use of Claude (Anthropic) for language editing and improving readability.

\section*{Funding}
This work was supported by Singapore MOE Tier 2 grant T2EP20125-0026 and Singapore MOH SIMFONI grant to D.L.

\section*{Declarations}

\subsection*{Author Contributions}
Yiming T. and D.L. conceived the project. Yiming T. developed the methodology, conducted experiments, and performed data analysis. A.L. and Q.Z. contributed to experimental validation and manuscript drafting. Yingtao Z. and Ye Z. assisted with experimental validation. S.A., T.N., Yih-Chung T., E.A., C.C., and Y.D. provided critical feedback and helped conceive key experiments. D.L. supervised the project. All authors contributed to manuscript preparation and approved the final version.

\subsection*{Competing Interests}
The authors declare no competing interests.

\bibliography{main}

@article{porter2024ai,
  title={AI-generated poetry is indistinguishable from human-written poetry and is rated more favorably},
  author={Porter, Brian and Machery, Edouard},
  journal={Scientific Reports},
  volume={14},
  number={1},
  pages={26133},
  year={2024},
  publisher={Nature Publishing Group UK London}
}

@article{nightingale2022ai,
  title={AI-synthesized faces are indistinguishable from real faces and more trustworthy},
  author={Nightingale, Sophie J and Farid, Hany},
  journal={Proceedings of the National Academy of Sciences},
  volume={119},
  number={8},
  pages={e2120481119},
  year={2022},
  publisher={National Academy of Sciences}
}

@article{morriss2010evolution,
  title={The evolution of human artistic creativity},
  author={Morriss-Kay, Gillian M},
  journal={Journal of anatomy},
  volume={216},
  number={2},
  pages={158--176},
  year={2010},
  publisher={Wiley Online Library}
}

@book{raaflaub2013thinking,
  title={Thinking, recording, and writing history in the Ancient World},
  author={Raaflaub, Kurt A},
  year={2013},
  publisher={John Wiley \& Sons}
}

@article{gripshover2020writing,
  title={Writing Systems as a Reflection of Ancient Societies},
  author={Gripshover, Sarah M},
  year={2020},
  publisher={Encompass Digital Archive, Eastern Kentucky University}
}

@misc{luo2023promptengineeringlensoptimal,
      title={Prompt Engineering Through the Lens of Optimal Control}, 
      author={Yifan Luo and Yiming Tang and Chengfeng Shen and Zhennan Zhou and Bin Dong},
      year={2023},
      eprint={2310.14201},
      archivePrefix={arXiv},
      primaryClass={cs.LG},
      url={https://arxiv.org/abs/2310.14201}, 
}

@misc{tang2025doesmodelfailautomatic,
      title={How does My Model Fail? Automatic Identification and Interpretation of Physical Plausibility Failure Modes with Matryoshka Transcoders}, 
      author={Yiming Tang and Abhijeet Sinha and Dianbo Liu},
      year={2025},
      eprint={2511.10094},
      archivePrefix={arXiv},
      primaryClass={cs.LG},
      url={https://arxiv.org/abs/2511.10094}, 
}

@misc{tang2026unifiedtheorysparsedictionary,
      title={A Unified Theory of Sparse Dictionary Learning in Mechanistic Interpretability: Piecewise Biconvexity and Spurious Minima}, 
      author={Yiming Tang and Harshvardhan Saini and Zhaoqian Yao and Zheng Lin and Yizhen Liao and Qianxiao Li and Mengnan Du and Dianbo Liu},
      year={2026},
      eprint={2512.05534},
      archivePrefix={arXiv},
      primaryClass={cs.LG},
      url={https://arxiv.org/abs/2512.05534}, 
}

@misc{tang2024demonstrationnotebookfindingsuited,
      title={Demonstration Notebook: Finding the Most Suited In-Context Learning Example from Interactions}, 
      author={Yiming Tang and Bin Dong},
      year={2024},
      eprint={2406.10878},
      archivePrefix={arXiv},
      primaryClass={cs.AI},
      url={https://arxiv.org/abs/2406.10878}, 
}

@misc{zhao2026rep2textdecodingtextsingle,
      title={Rep2Text: Decoding Full Text from a Single LLM Token Representation}, 
      author={Haiyan Zhao and Zirui He and Yiming Tang and Fan Yang and Ali Payani and Dianbo Liu and Mengnan Du},
      year={2026},
      eprint={2511.06571},
      archivePrefix={arXiv},
      primaryClass={cs.CL},
      url={https://arxiv.org/abs/2511.06571}, 
}

@misc{zou2026fmlbenchbenchmarkingmachinelearning,
      title={FML-bench: Benchmarking Machine Learning Agents for Scientific Research}, 
      author={Qiran Zou and Hou Hei Lam and Wenhao Zhao and Yiming Tang and Tingting Chen and Samson Yu and Tianyi Zhang and Chang Liu and Xiangyang Ji and Dianbo Liu},
      year={2026},
      eprint={2510.10472},
      archivePrefix={arXiv},
      primaryClass={cs.CL},
      url={https://arxiv.org/abs/2510.10472}, 
}

@misc{cunningham2023sparseautoencodershighlyinterpretable,
      title={Sparse Autoencoders Find Highly Interpretable Features in Language Models}, 
      author={Hoagy Cunningham and Aidan Ewart and Logan Riggs and Robert Huben and Lee Sharkey},
      year={2023},
      eprint={2309.08600},
      archivePrefix={arXiv},
      primaryClass={cs.LG},
      url={https://arxiv.org/abs/2309.08600}, 
}

@misc{gao2024scalingevaluatingsparseautoencoders,
      title={Scaling and evaluating sparse autoencoders}, 
      author={Leo Gao and Tom Dupré la Tour and Henk Tillman and Gabriel Goh and Rajan Troll and Alec Radford and Ilya Sutskever and Jan Leike and Jeffrey Wu},
      year={2024},
      eprint={2406.04093},
      archivePrefix={arXiv},
      primaryClass={cs.LG},
      url={https://arxiv.org/abs/2406.04093}, 
}

@misc{dunefsky2024transcodersinterpretablellmfeature,
      title={Transcoders Find Interpretable LLM Feature Circuits}, 
      author={Jacob Dunefsky and Philippe Chlenski and Neel Nanda},
      year={2024},
      eprint={2406.11944},
      archivePrefix={arXiv},
      primaryClass={cs.LG},
      url={https://arxiv.org/abs/2406.11944}, 
}

@misc{chen2025contrastivelocalizedlanguageimagepretraining,
      title={Contrastive Localized Language-Image Pre-Training}, 
      author={Hong-You Chen and Zhengfeng Lai and Haotian Zhang and Xinze Wang and Marcin Eichner and Keen You and Meng Cao and Bowen Zhang and Yinfei Yang and Zhe Gan},
      year={2025},
      eprint={2410.02746},
      archivePrefix={arXiv},
      primaryClass={cs.CV},
      url={https://arxiv.org/abs/2410.02746}, 
}

@misc{makhzani2014ksparseautoencoders,
      title={k-Sparse Autoencoders}, 
      author={Alireza Makhzani and Brendan Frey},
      year={2014},
      eprint={1312.5663},
      archivePrefix={arXiv},
      primaryClass={cs.LG},
      url={https://arxiv.org/abs/1312.5663}, 
}

@misc{lin2015microsoftcococommonobjects,
      title={Microsoft COCO: Common Objects in Context}, 
      author={Tsung-Yi Lin and Michael Maire and Serge Belongie and Lubomir Bourdev and Ross Girshick and James Hays and Pietro Perona and Deva Ramanan and C. Lawrence Zitnick and Piotr Dollár},
      year={2015},
      eprint={1405.0312},
      archivePrefix={arXiv},
      primaryClass={cs.CV},
      url={https://arxiv.org/abs/1405.0312}, 
}

@misc{rombach2022highresolutionimagesynthesislatent,
      title={High-Resolution Image Synthesis with Latent Diffusion Models}, 
      author={Robin Rombach and Andreas Blattmann and Dominik Lorenz and Patrick Esser and Björn Ommer},
      year={2022},
      eprint={2112.10752},
      archivePrefix={arXiv},
      primaryClass={cs.CV},
      url={https://arxiv.org/abs/2112.10752}, 
}

@misc{stablecascade2025,
  title={Stable Cascade: Efficient Text-to-Image Generation in Highly Compressed Latent Spaces},
  author={Stability AI},
  year={2025},
  howpublished={\url{https://huggingface.co/stabilityai/stable-cascade}},
  note={Technical Report}
}

@misc{zhang2023addingconditionalcontroltexttoimage,
      title={Adding Conditional Control to Text-to-Image Diffusion Models}, 
      author={Lvmin Zhang and Anyi Rao and Maneesh Agrawala},
      year={2023},
      eprint={2302.05543},
      archivePrefix={arXiv},
      primaryClass={cs.CV},
      url={https://arxiv.org/abs/2302.05543}, 
}

@article{bricken2023monosemanticity,
    title={Towards Monosemanticity: Decomposing Language Models With Dictionary Learning},
    author={Bricken, Trenton and Templeton, Adly and Batson, Joshua and Chen, Brian and Jermyn, Adam and Conerly, Tom and Turner, Nick and Anil, Cem and Denison, Carson and Askell, Amanda and Lasenby, Robert and Wu, Yifan and Kravec, Shauna and Schiefer, Nicholas and Maxwell, Tim and Joseph, Nicholas and Hatfield-Dodds, Zac and Tamkin, Alex and Nguyen, Karina and McLean, Brayden and Burke, Josiah E and Hume, Tristan and Carter, Shan and Henighan, Tom and Olah, Christopher},
    year={2023},
    journal={Transformer Circuits Thread},
    note={https://transformer-circuits.pub/2023/monosemantic-features/index.html}
}

@article{templeton2024scaling,
    title={Scaling Monosemanticity: Extracting Interpretable Features from Claude 3 Sonnet},
    author={Templeton, Adly and Conerly, Tom and Marcus, Jonathan and Lindsey, Jack and Bricken, Trenton and Chen, Brian and Pearce, Adam and Citro, Craig and Ameisen, Emmanuel and Jones, Andy and Cunningham, Hoagy and Turner, Nicholas L and McDougall, Callum and MacDiarmid, Monte and Freeman, C. Daniel and Sumers, Theodore R. and Rees, Edward and Batson, Joshua and Jermyn, Adam and Carter, Shan and Olah, Chris and Henighan, Tom},
    year={2024},
    journal={Transformer Circuits Thread},
    url={https://transformer-circuits.pub/2024/scaling-monosemanticity/index.html}
}

@misc{jayasumana2024rethinkingfidbetterevaluation,
      title={Rethinking FID: Towards a Better Evaluation Metric for Image Generation}, 
      author={Sadeep Jayasumana and Srikumar Ramalingam and Andreas Veit and Daniel Glasner and Ayan Chakrabarti and Sanjiv Kumar},
      year={2024},
      eprint={2401.09603},
      archivePrefix={arXiv},
      primaryClass={cs.CV},
      url={https://arxiv.org/abs/2401.09603}, 
}

@misc{claudehaiku,
  author       = {Anthropic},
  title        = {Claude Haiku},
  year         = {2024},
  howpublished = {\url{https://www.anthropic.com/index/claude}},
  note         = {Accessed: 2025-05-04}
}

@misc{llama3,
  author       = {{Meta AI}},
  title        = {LLaMA 3 Instruct},
  year         = {2024},
  howpublished = {\url{https://ai.meta.com/llama/}},
  note         = {Accessed: 2025-05-04}
}

@misc{sun2025high,
  title={High Frequency Latents Are Features, Not Bugs},
  author={Xiaoqing Sun and Joshua Engels and Max Tegmark},
  year={2025},
  howpublished={\url{https://openreview.net/forum?id=IT5fRjnGr0}},
  note={Sparsity in LLMs (SLLM): Deep Dive into Mixture of Experts, Quantization, Hardware, and Inference}
}

@misc{sampaio2024typescoretextfidelitymetric,
      title={TypeScore: A Text Fidelity Metric for Text-to-Image Generative Models}, 
      author={Georgia Gabriela Sampaio and Ruixiang Zhang and Shuangfei Zhai and Jiatao Gu and Josh Susskind and Navdeep Jaitly and Yizhe Zhang},
      year={2024},
      eprint={2411.02437},
      archivePrefix={arXiv},
      primaryClass={cs.CV},
      url={https://arxiv.org/abs/2411.02437}, 
}

@misc{sidorov2020textcapsdatasetimagecaptioning,
      title={TextCaps: a Dataset for Image Captioning with Reading Comprehension}, 
      author={Oleksii Sidorov and Ronghang Hu and Marcus Rohrbach and Amanpreet Singh},
      year={2020},
      eprint={2003.12462},
      archivePrefix={arXiv},
      primaryClass={cs.CV},
      url={https://arxiv.org/abs/2003.12462}, 
}

@misc{Agrawal_2019,
  title        = {nocaps: novel object captioning at scale},
  author       = {Agrawal, Harsh and Desai, Karan and Wang, Yufei and Chen, Xinlei and Jain, Rishabh and Johnson, Mark and Batra, Dhruv and Parikh, Devi and Lee, Stefan and Anderson, Peter},
  year         = {2019},
  month        = {oct},
  howpublished = {\url{http://dx.doi.org/10.1109/ICCV.2019.00904}},
  note         = {2019 IEEE/CVF International Conference on Computer Vision (ICCV), IEEE}
}

@misc{plummer2016flickr30kentitiescollectingregiontophrase,
      title={Flickr30k Entities: Collecting Region-to-Phrase Correspondences for Richer Image-to-Sentence Models}, 
      author={Bryan A. Plummer and Liwei Wang and Chris M. Cervantes and Juan C. Caicedo and Julia Hockenmaier and Svetlana Lazebnik},
      year={2016},
      eprint={1505.04870},
      archivePrefix={arXiv},
      primaryClass={cs.CV},
      url={https://arxiv.org/abs/1505.04870}, 
}

@misc{schrodi2025effectstriggermodalitygap,
      title={Two Effects, One Trigger: On the Modality Gap, Object Bias, and Information Imbalance in Contrastive Vision-Language Models}, 
      author={Simon Schrodi and David T. Hoffmann and Max Argus and Volker Fischer and Thomas Brox},
      year={2025},
      eprint={2404.07983},
      archivePrefix={arXiv},
      primaryClass={cs.CV},
      url={https://arxiv.org/abs/2404.07983}, 
}

@misc{hinton2015distillingknowledgeneuralnetwork,
      title={Distilling the Knowledge in a Neural Network}, 
      author={Geoffrey Hinton and Oriol Vinyals and Jeff Dean},
      year={2015},
      eprint={1503.02531},
      archivePrefix={arXiv},
      primaryClass={stat.ML},
      url={https://arxiv.org/abs/1503.02531}, 
}

@misc{xu2023parameterefficientfinetuningmethodspretrained,
      title={Parameter-Efficient Fine-Tuning Methods for Pretrained Language Models: A Critical Review and Assessment}, 
      author={Lingling Xu and Haoran Xie and Si-Zhao Joe Qin and Xiaohui Tao and Fu Lee Wang},
      year={2023},
      eprint={2312.12148},
      archivePrefix={arXiv},
      primaryClass={cs.CL},
      url={https://arxiv.org/abs/2312.12148}, 
}

@misc{elhage2022toymodelssuperposition,
      title={Toy Models of Superposition}, 
      author={Nelson Elhage and Tristan Hume and Catherine Olsson and Nicholas Schiefer and Tom Henighan and Shauna Kravec and Zac Hatfield-Dodds and Robert Lasenby and Dawn Drain and Carol Chen and Roger Grosse and Sam McCandlish and Jared Kaplan and Dario Amodei and Martin Wattenberg and Christopher Olah},
      year={2022},
      eprint={2209.10652},
      archivePrefix={arXiv},
      primaryClass={cs.LG},
      url={https://arxiv.org/abs/2209.10652}, 
}

@misc{lieberum2024gemmascopeopensparse,
      title={Gemma Scope: Open Sparse Autoencoders Everywhere All At Once on Gemma 2}, 
      author={Tom Lieberum and Senthooran Rajamanoharan and Arthur Conmy and Lewis Smith and Nicolas Sonnerat and Vikrant Varma and János Kramár and Anca Dragan and Rohin Shah and Neel Nanda},
      year={2024},
      eprint={2408.05147},
      archivePrefix={arXiv},
      primaryClass={cs.LG},
      url={https://arxiv.org/abs/2408.05147}, 
}

@misc{bussmann2025learningmultilevelfeaturesmatryoshka,
      title={Learning Multi-Level Features with Matryoshka Sparse Autoencoders}, 
      author={Bart Bussmann and Noa Nabeshima and Adam Karvonen and Neel Nanda},
      year={2025},
      eprint={2503.17547},
      archivePrefix={arXiv},
      primaryClass={cs.LG},
      url={https://arxiv.org/abs/2503.17547}, 
}

@misc{hessel2022clipscorereferencefreeevaluationmetric,
      title={CLIPScore: A Reference-free Evaluation Metric for Image Captioning}, 
      author={Jack Hessel and Ari Holtzman and Maxwell Forbes and Ronan Le Bras and Yejin Choi},
      year={2022},
      eprint={2104.08718},
      archivePrefix={arXiv},
      primaryClass={cs.CV},
      url={https://arxiv.org/abs/2104.08718}, 
}

@article{zeki1999art,
  title={Art and the brain},
  author={Zeki, Semir},
  journal={Journal of Consciousness Studies},
  volume={6},
  number={6-7},
  pages={76--96},
  year={1999},
  publisher={Imprint Academic}
}

@book{clark1996using,
  title={Using language},
  author={Clark, Herbert H},
  year={1996},
  publisher={Cambridge university press}
}

@misc{paulo2025transcodersbeatsparseautoencoders,
      title={Transcoders Beat Sparse Autoencoders for Interpretability}, 
      author={Gonçalo Paulo and Stepan Shabalin and Nora Belrose},
      year={2025},
      eprint={2501.18823},
      archivePrefix={arXiv},
      primaryClass={cs.LG},
      url={https://arxiv.org/abs/2501.18823}, 
}

@misc{meng2022sdeditguidedimagesynthesis,
      title={SDEdit: Guided Image Synthesis and Editing with Stochastic Differential Equations}, 
      author={Chenlin Meng and Yutong He and Yang Song and Jiaming Song and Jiajun Wu and Jun-Yan Zhu and Stefano Ermon},
      year={2022},
      eprint={2108.01073},
      archivePrefix={arXiv},
      primaryClass={cs.CV},
      url={https://arxiv.org/abs/2108.01073}, 
}

@misc{dalle3,
  author       = {Evgeniy Hristoforu},
  title        = {DALL·E 3},
  year         = {2023},
  howpublished = {\url{https://huggingface.co/ehristoforu/dalle-3-xl-v2}},
  note         = {Accessed: 2025-03-14}
}

@misc{FLUX.1-dev,
  author       = {Black Forest Labs},
  title        = {FLUX.1-dev},
  year         = {2024},
  howpublished = {\url{https://huggingface.co/black-forest-labs/FLUX.1-dev}},
  note         = {Accessed: 2025-03-14}
}

@misc{Kolors-diffusers,
  author       = {Kwai-Kolors},
  title        = {Kolors-diffusers},
  year         = {2024},
  howpublished = {\url{https://huggingface.co/Kwai-Kolors/Kolors-diffusers}},
  note         = {Accessed: 2025-03-14}
}

@misc{lagzian2025multinoveltyimprovediversitynovelty,
      title={Multi-Novelty: Improve the Diversity and Novelty of Contents Generated by Large Language Models via inference-time Multi-Views Brainstorming}, 
      author={Arash Lagzian and Srinivas Anumasa and Dianbo Liu},
      year={2025},
      eprint={2502.12700},
      archivePrefix={arXiv},
      primaryClass={cs.CL},
      url={https://arxiv.org/abs/2502.12700}, 
}

@misc{chakrabarty2024artartificelargelanguage,
      title={Art or Artifice? Large Language Models and the False Promise of Creativity}, 
      author={Tuhin Chakrabarty and Philippe Laban and Divyansh Agarwal and Smaranda Muresan and Chien-Sheng Wu},
      year={2024},
      eprint={2309.14556},
      archivePrefix={arXiv},
      primaryClass={cs.CL},
      url={https://arxiv.org/abs/2309.14556}, 
}

@book{baer2014creativity,
  title={Creativity and divergent thinking: A task-specific approach},
  author={Baer, John},
  year={2014},
  publisher={Psychology Press}
}

@misc{chen2023understandingvulnerabilityclipimage,
      title={Understanding the Vulnerability of CLIP to Image Compression}, 
      author={Cangxiong Chen and Vinay P. Namboodiri and Julian Padget},
      year={2023},
      eprint={2311.14029},
      archivePrefix={arXiv},
      primaryClass={cs.CV},
      url={https://arxiv.org/abs/2311.14029}, 
}

@misc{farrell2024applyingsparseautoencodersunlearn,
      title={Applying sparse autoencoders to unlearn knowledge in language models}, 
      author={Eoin Farrell and Yeu-Tong Lau and Arthur Conmy},
      year={2024},
      eprint={2410.19278},
      archivePrefix={arXiv},
      primaryClass={cs.LG},
      url={https://arxiv.org/abs/2410.19278}, 
}

@article{lindsey2024sparse,
  title={Sparse Crosscoders for Cross-Layer Features and Model Diffing},
  author={Lindsey, J. and Templeton, A. and Marcus, J. and Conerly, T. and Baston, J. and Olah, C.},
  year={2024},
  journal={arXiv preprint arXiv:2402.14634},
  note={Accessed June 2025}
}

@misc{ramesh2022hierarchicaltextconditionalimagegeneration,
      title={Hierarchical Text-Conditional Image Generation with CLIP Latents}, 
      author={Aditya Ramesh and Prafulla Dhariwal and Alex Nichol and Casey Chu and Mark Chen},
      year={2022},
      eprint={2204.06125},
      archivePrefix={arXiv},
      primaryClass={cs.CV},
      url={https://arxiv.org/abs/2204.06125}, 
}

@misc{lim2025evaluatingimagehallucinationtexttoimage,
      title={Evaluating Image Hallucination in Text-to-Image Generation with Question-Answering}, 
      author={Youngsun Lim and Hojun Choi and Hyunjung Shim},
      year={2025},
      eprint={2409.12784},
      archivePrefix={arXiv},
      primaryClass={cs.CV},
      url={https://arxiv.org/abs/2409.12784}, 
}

@misc{kondapaneni2024textimagealignmentdiffusionbasedperception,
      title={Text-image Alignment for Diffusion-based Perception}, 
      author={Neehar Kondapaneni and Markus Marks and Manuel Knott and Rogerio Guimaraes and Pietro Perona},
      year={2024},
      eprint={2310.00031},
      archivePrefix={arXiv},
      primaryClass={cs.CV},
      url={https://arxiv.org/abs/2310.00031}, 
}

@misc{inoue2018crossdomainweaklysupervisedobjectdetection,
      title={Cross-Domain Weakly-Supervised Object Detection through Progressive Domain Adaptation}, 
      author={Naoto Inoue and Ryosuke Furuta and Toshihiko Yamasaki and Kiyoharu Aizawa},
      year={2018},
      eprint={1803.11365},
      archivePrefix={arXiv},
      primaryClass={cs.CV},
      url={https://arxiv.org/abs/1803.11365}, 
}

@misc{kwon2024graspdiffusionsynthesizingrealisticwholebody,
      title={GraspDiffusion: Synthesizing Realistic Whole-body Hand-Object Interaction}, 
      author={Patrick Kwon and Hanbyul Joo},
      year={2024},
      eprint={2410.13911},
      archivePrefix={arXiv},
      primaryClass={cs.CV},
      url={https://arxiv.org/abs/2410.13911}, 
}

@misc{Narasimhaswamy_2024,
  author       = {Narasimhaswamy, Supreeth and Bhattacharya, Uttaran and Chen, Xiang and Dasgupta, Ishita and Mitra, Saayan and Hoai, Minh},
  title        = {HanDiffuser: Text-to-Image Generation with Realistic Hand Appearances},
  howpublished = {Presented at CVPR 2024},
  year         = {2024},
  month        = jun,
  note         = {Available at \url{https://doi.org/10.1109/CVPR52733.2024.00239}}
}

@misc{rai2025practicalreviewmechanisticinterpretability,
      title={A Practical Review of Mechanistic Interpretability for Transformer-Based Language Models}, 
      author={Daking Rai and Yilun Zhou and Shi Feng and Abulhair Saparov and Ziyu Yao},
      year={2025},
      eprint={2407.02646},
      archivePrefix={arXiv},
      primaryClass={cs.AI},
      url={https://arxiv.org/abs/2407.02646}, 
}

@article{biederman1987recognition,
  title={Recognition-by-components: a theory of human image understanding.},
  author={Biederman, Irving},
  journal={Psychological review},
  volume={94},
  number={2},
  pages={115},
  year={1987},
  publisher={American Psychological Association}
}

@article{nightingale2017can,
  title={Can people identify original and manipulated photos of real-world scenes?},
  author={Nightingale, Sophie J and Wade, Kimberley A and Watson, Derrick G},
  journal={Cognitive research: principles and implications},
  volume={2},
  pages={1--21},
  year={2017},
  publisher={Springer}
}

@article{farid2022creating,
  title={Creating, using, misusing, and detecting deep fakes},
  author={Farid, Hany},
  journal={Journal of Online Trust and Safety},
  volume={1},
  number={4},
  year={2022}
}

@article{landy2013texture,
  title={Texture analysis and perception},
  author={Landy, Michael S},
  journal={The new visual neurosciences},
  volume={476},
  pages={639--652},
  year={2013},
  publisher={MIT Press Cambridge}
}

@article{Santos03042018,
    author = {Paulo E. Santos and Roberto Casati and Patrick Cavanagh and},
    title = {Perception, cognition and reasoning about shadows},
    journal = {Spatial Cognition \& Computation},
    volume = {18},
    number = {2},
    pages = {78--85},
    year = {2018},
    publisher = {Taylor \& Francis},
    doi = {10.1080/13875868.2017.1377204},
    URL = {
            https://doi.org/10.1080/13875868.2017.1377204
    },
    eprint = { 
            https://doi.org/10.1080/13875868.2017.1377204
    }
}

@article{rhodes1998facial,
  title={Facial symmetry and the perception of beauty},
  author={Rhodes, Gillian and Proffitt, Fiona and Grady, Jonathon M and Sumich, Alex},
  journal={Psychonomic Bulletin \& Review},
  volume={5},
  pages={659--669},
  year={1998},
  publisher={Springer}
}

@misc{huang2025t2icompbenchenhancedcomprehensivebenchmark,
      title={T2I-CompBench++: An Enhanced and Comprehensive Benchmark for Compositional Text-to-image Generation}, 
      author={Kaiyi Huang and Chengqi Duan and Kaiyue Sun and Enze Xie and Zhenguo Li and Xihui Liu},
      year={2025},
      eprint={2307.06350},
      archivePrefix={arXiv},
      primaryClass={cs.CV},
      url={https://arxiv.org/abs/2307.06350}, 
}

@misc{kynkäänniemi2019improvedprecisionrecallmetric,
      title={Improved Precision and Recall Metric for Assessing Generative Models}, 
      author={Tuomas Kynkäänniemi and Tero Karras and Samuli Laine and Jaakko Lehtinen and Timo Aila},
      year={2019},
      eprint={1904.06991},
      archivePrefix={arXiv},
      primaryClass={stat.ML},
      url={https://arxiv.org/abs/1904.06991}, 
}

@misc{yang2018physicsinformeddeepgenerativemodels,
      title={Physics-informed deep generative models}, 
      author={Yibo Yang and Paris Perdikaris},
      year={2018},
      eprint={1812.03511},
      archivePrefix={arXiv},
      primaryClass={stat.ML},
      url={https://arxiv.org/abs/1812.03511}, 
}

@misc{gerstgrasser2024modelcollapseinevitablebreaking,
      title={Is Model Collapse Inevitable? Breaking the Curse of Recursion by Accumulating Real and Synthetic Data}, 
      author={Matthias Gerstgrasser and Rylan Schaeffer and Apratim Dey and Rafael Rafailov and Henry Sleight and John Hughes and Tomasz Korbak and Rajashree Agrawal and Dhruv Pai and Andrey Gromov and Daniel A. Roberts and Diyi Yang and David L. Donoho and Sanmi Koyejo},
      year={2024},
      eprint={2404.01413},
      archivePrefix={arXiv},
      primaryClass={cs.LG},
      url={https://arxiv.org/abs/2404.01413}, 
}

@misc{barratt2018noteinceptionscore,
      title={A Note on the Inception Score}, 
      author={Shane Barratt and Rishi Sharma},
      year={2018},
      eprint={1801.01973},
      archivePrefix={arXiv},
      primaryClass={stat.ML},
      url={https://arxiv.org/abs/1801.01973}, 
}

@misc{minegishi2025rethinkingevaluationsparseautoencoders,
      title={Rethinking Evaluation of Sparse Autoencoders through the Representation of Polysemous Words}, 
      author={Gouki Minegishi and Hiroki Furuta and Yusuke Iwasawa and Yutaka Matsuo},
      year={2025},
      eprint={2501.06254},
      archivePrefix={arXiv},
      primaryClass={cs.CL},
      url={https://arxiv.org/abs/2501.06254}, 
}

@misc{johnson2019mimiccxrjpglargepubliclyavailable,
      title={MIMIC-CXR-JPG, a large publicly available database of labeled chest radiographs}, 
      author={Alistair E. W. Johnson and Tom J. Pollard and Nathaniel R. Greenbaum and Matthew P. Lungren and Chih-ying Deng and Yifan Peng and Zhiyong Lu and Roger G. Mark and Seth J. Berkowitz and Steven Horng},
      year={2019},
      eprint={1901.07042},
      archivePrefix={arXiv},
      primaryClass={cs.CV},
      url={https://arxiv.org/abs/1901.07042}, 
}

@misc{malik2023stable,
  author = {Malik, Danyal},
  title = {Stable Diffusion Chest X-ray: DreamBooth model trained on chest-xray14 dataset},
  year = {2023},
  url = {https://huggingface.co/danyalmalik/stable-diffusion-chest-xray},
  note = {Hugging Face model repository}
}

@book{review2024generative,
  title={Generative AI: The Insights You Need from Harvard Business Review},
  author={Review, H.B. and Mollick, E. and De Cremer, D. and Neeley, T. and Sinha, P.},
  isbn={9781647826406},
  lccn={2023029122},
  series={HBR Insights Series},
  url={https://books.google.com.sg/books?id=MIvGEAAAQBAJ},
  year={2024},
  publisher={Harvard Business Review Press}
}

@article{doshi2024generative,
  title={Generative AI enhances individual creativity but reduces the collective diversity of novel content},
  author={Doshi, Anil R and Hauser, Oliver P},
  journal={Science advances},
  volume={10},
  number={28},
  pages={eadn5290},
  year={2024},
  publisher={American Association for the Advancement of Science}
}

@article{gil2022will,
  title={Will AI write scientific papers in the future?},
  author={Gil, Yolanda},
  journal={AI Magazine},
  volume={42},
  number={4},
  pages={3--15},
  year={2022}
}

@article{mahmood2019deep,
  title={Deep adversarial training for multi-organ nuclei segmentation in histopathology images},
  author={Mahmood, Faisal and Borders, Daniel and Chen, Richard J and McKay, Gregory N and Salimian, Kevan J and Baras, Alexander and Durr, Nicholas J},
  journal={IEEE transactions on medical imaging},
  volume={39},
  number={11},
  pages={3257--3267},
  year={2019},
  publisher={IEEE}
}

@misc{teixeira2018generatingsyntheticxrayimages,
      title={Generating Synthetic X-ray Images of a Person from the Surface Geometry}, 
      author={Brian Teixeira and Vivek Singh and Terrence Chen and Kai Ma and Birgi Tamersoy and Yifan Wu and Elena Balashova and Dorin Comaniciu},
      year={2018},
      eprint={1805.00553},
      archivePrefix={arXiv},
      primaryClass={cs.CV},
      url={https://arxiv.org/abs/1805.00553}, 
}

@article{kitano2021nobel,
  title={Nobel Turing Challenge: creating the engine for scientific discovery},
  author={Kitano, Hiroaki},
  journal={NPJ systems biology and applications},
  volume={7},
  number={1},
  pages={29},
  year={2021},
  publisher={Nature Publishing Group UK London}
}

@article{LivingBraveAI2023,
  title   = {Living in a brave new AI era},
  journal = {Nature Human Behaviour},
  volume  = {7},
  pages   = {1799},
  year    = {2023},
  doi     = {10.1038/s41562-023-01775-7},
  url     = {https://doi.org/10.1038/s41562-023-01775-7},
  note    = {Published 20 November 2023},
}

@article{kottapallitransforming,
  title={Transforming Real Images to Studio Ghibli Style: A Comparative Study of Deep Learning Approaches},
  author={Kottapalli, Venkatalakshmi and Kundargi, Satvik}
}

@article{lindsay2023llms,
  title={LLMs are not ready for editorial work},
  author={Lindsay, Grace W},
  journal={Nature human behaviour},
  volume={7},
  number={11},
  pages={1814--1815},
  year={2023},
  publisher={Nature Publishing Group UK London}
}

@article{suzuki2023we,
  title={We need a culturally aware approach to AI},
  author={Suzuki, Shoko},
  journal={Nature Human Behaviour},
  volume={7},
  number={11},
  pages={1816--1817},
  year={2023},
  publisher={Nature Publishing Group UK London}
}

@article{shumailov2024ai,
  title={AI models collapse when trained on recursively generated data},
  author={Shumailov, Ilia and Shumaylov, Zakhar and Zhao, Yiren and Papernot, Nicolas and Anderson, Ross and Gal, Yarin},
  journal={Nature},
  volume={631},
  number={8022},
  pages={755--759},
  year={2024},
  publisher={Nature Publishing Group UK London}
}

@article{smith2024ai,
  title={Ai model collapse might be prevented by studying human language transmission},
  author={Smith, Kenny and Kirby, Simon and Guo, Shangmin and Griffiths, Thomas L},
  journal={Nature},
  volume={633},
  number={8030},
  pages={525},
  year={2024}
}

@article{alber2025medical,
  title={Medical large language models are vulnerable to data-poisoning attacks},
  author={Alber, Daniel Alexander and Yang, Zihao and Alyakin, Anton and Yang, Eunice and Rai, Sumedha and Valliani, Aly A and Zhang, Jeff and Rosenbaum, Gabriel R and Amend-Thomas, Ashley K and Kurland, David B and others},
  journal={Nature Medicine},
  volume={31},
  number={2},
  pages={618--626},
  year={2025},
  publisher={Nature Publishing Group US New York}
}

@article{hager2024evaluation,
  title={Evaluation and mitigation of the limitations of large language models in clinical decision-making},
  author={Hager, Paul and Jungmann, Friederike and Holland, Robbie and Bhagat, Kunal and Hubrecht, Inga and Knauer, Manuel and Vielhauer, Jakob and Makowski, Marcus and Braren, Rickmer and Kaissis, Georgios and others},
  journal={Nature medicine},
  volume={30},
  number={9},
  pages={2613--2622},
  year={2024},
  publisher={Nature Publishing Group US New York}
}

@article{choudhury2023generative,
  title={Generative AI has a language problem},
  author={Choudhury, Monojit},
  journal={Nature human behaviour},
  volume={7},
  number={11},
  pages={1802--1803},
  year={2023},
  publisher={Nature Publishing Group UK London}
}

@article{kundu2023measuring,
  title={Measuring trustworthiness is crucial for medical AI tools},
  author={Kundu, Shinjini},
  journal={Nature Human Behaviour},
  volume={7},
  number={11},
  pages={1812--1813},
  year={2023},
  publisher={Nature Publishing Group UK London}
}

@misc{feng2025unravelingmisinformationpropagationllm,
      title={Unraveling Misinformation Propagation in LLM Reasoning}, 
      author={Yiyang Feng and Yichen Wang and Shaobo Cui and Boi Faltings and Mina Lee and Jiawei Zhou},
      year={2025},
      eprint={2505.18555},
      archivePrefix={arXiv},
      primaryClass={cs.CL},
      url={https://arxiv.org/abs/2505.18555}, 
}

@misc{li2023haluevallargescalehallucinationevaluation,
      title={HaluEval: A Large-Scale Hallucination Evaluation Benchmark for Large Language Models}, 
      author={Junyi Li and Xiaoxue Cheng and Wayne Xin Zhao and Jian-Yun Nie and Ji-Rong Wen},
      year={2023},
      eprint={2305.11747},
      archivePrefix={arXiv},
      primaryClass={cs.CL},
      url={https://arxiv.org/abs/2305.11747}, 
}

@misc{fang2024humanrefinerbenchmarkingabnormalhuman,
      title={HumanRefiner: Benchmarking Abnormal Human Generation and Refining with Coarse-to-fine Pose-Reversible Guidance}, 
      author={Guian Fang and Wenbiao Yan and Yuanfan Guo and Jianhua Han and Zutao Jiang and Hang Xu and Shengcai Liao and Xiaodan Liang},
      year={2024},
      eprint={2407.06937},
      archivePrefix={arXiv},
      primaryClass={cs.CV},
      url={https://arxiv.org/abs/2407.06937}, 
}

@misc{yao2024hifi,
  title        = {HiFi-Score: Fine-Grained Image Description Evaluation with Hierarchical Parsing Graphs},
  author       = {Yao, Ziwei and Wang, Ruiping and Chen, Xilin},
  year         = {2024},
  howpublished = {European Conference on Computer Vision (ECCV), pp. 441--458. Springer},
}

@article{acion2023generative,
  title={Generative AI poses ethical challenges for open science},
  author={Acion, Laura and Rajngewerc, Mariela and Randall, Gregory and Etcheverry, Lorena},
  journal={Nature Human Behaviour},
  volume={7},
  number={11},
  pages={1800--1801},
  year={2023},
  publisher={Nature Publishing Group UK London}
}

@article{Koco__2023,
   title={ChatGPT: Jack of all trades, master of none},
   volume={99},
   ISSN={1566-2535},
   url={http://dx.doi.org/10.1016/j.inffus.2023.101861},
   DOI={10.1016/j.inffus.2023.101861},
   journal={Information Fusion},
   publisher={Elsevier BV},
   author={Kocoń, Jan and Cichecki, Igor and Kaszyca, Oliwier and Kochanek, Mateusz and Szydło, Dominika and Baran, Joanna and Bielaniewicz, Julita and Gruza, Marcin and Janz, Arkadiusz and Kanclerz, Kamil and Kocoń, Anna and Koptyra, Bartłomiej and Mieleszczenko-Kowszewicz, Wiktoria and Miłkowski, Piotr and Oleksy, Marcin and Piasecki, Maciej and Radliński, Łukasz and Wojtasik, Konrad and Woźniak, Stanisław and Kazienko, Przemysław},
   year={2023},
   month=nov, pages={101861} }

@misc{bubeck2023sparksartificialgeneralintelligence,
      title={Sparks of Artificial General Intelligence: Early experiments with GPT-4}, 
      author={Sébastien Bubeck and Varun Chandrasekaran and Ronen Eldan and Johannes Gehrke and Eric Horvitz and Ece Kamar and Peter Lee and Yin Tat Lee and Yuanzhi Li and Scott Lundberg and Harsha Nori and Hamid Palangi and Marco Tulio Ribeiro and Yi Zhang},
      year={2023},
      eprint={2303.12712},
      archivePrefix={arXiv},
      primaryClass={cs.CL},
      url={https://arxiv.org/abs/2303.12712}, 
}

@misc{bang2023multitaskmultilingualmultimodalevaluation,
      title={A Multitask, Multilingual, Multimodal Evaluation of ChatGPT on Reasoning, Hallucination, and Interactivity}, 
      author={Yejin Bang and Samuel Cahyawijaya and Nayeon Lee and Wenliang Dai and Dan Su and Bryan Wilie and Holy Lovenia and Ziwei Ji and Tiezheng Yu and Willy Chung and Quyet V. Do and Yan Xu and Pascale Fung},
      year={2023},
      eprint={2302.04023},
      archivePrefix={arXiv},
      primaryClass={cs.CL},
      url={https://arxiv.org/abs/2302.04023}, 
}

@misc{chen2025personavectorsmonitoringcontrolling,
      title={Persona Vectors: Monitoring and Controlling Character Traits in Language Models}, 
      author={Runjin Chen and Andy Arditi and Henry Sleight and Owain Evans and Jack Lindsey},
      year={2025},
      eprint={2507.21509},
      archivePrefix={arXiv},
      primaryClass={cs.CL},
      url={https://arxiv.org/abs/2507.21509}, 
}

@misc{turner2024steeringlanguagemodelsactivation,
      title={Steering Language Models With Activation Engineering}, 
      author={Alexander Matt Turner and Lisa Thiergart and Gavin Leech and David Udell and Juan J. Vazquez and Ulisse Mini and Monte MacDiarmid},
      year={2024},
      eprint={2308.10248},
      archivePrefix={arXiv},
      primaryClass={cs.CL},
      url={https://arxiv.org/abs/2308.10248}, 
}

@misc{zou2025representationengineeringtopdownapproach,
      title={Representation Engineering: A Top-Down Approach to AI Transparency}, 
      author={Andy Zou and Long Phan and Sarah Chen and James Campbell and Phillip Guo and Richard Ren and Alexander Pan and Xuwang Yin and Mantas Mazeika and Ann-Kathrin Dombrowski and Shashwat Goel and Nathaniel Li and Michael J. Byun and Zifan Wang and Alex Mallen and Steven Basart and Sanmi Koyejo and Dawn Song and Matt Fredrikson and J. Zico Kolter and Dan Hendrycks},
      year={2025},
      eprint={2310.01405},
      archivePrefix={arXiv},
      primaryClass={cs.LG},
      url={https://arxiv.org/abs/2310.01405}, 
}

@misc{wu2025axbenchsteeringllmssimple,
      title={AxBench: Steering LLMs? Even Simple Baselines Outperform Sparse Autoencoders}, 
      author={Zhengxuan Wu and Aryaman Arora and Atticus Geiger and Zheng Wang and Jing Huang and Dan Jurafsky and Christopher D. Manning and Christopher Potts},
      year={2025},
      eprint={2501.17148},
      archivePrefix={arXiv},
      primaryClass={cs.CL},
      url={https://arxiv.org/abs/2501.17148}, 
}

@article{10.1093/polsoc/puaf001,
    author = {Taeihagh, Araz},
    title = {Governance of Generative AI},
    journal = {Policy and Society},
    volume = {44},
    number = {1},
    pages = {1-22},
    year = {2025},
    month = {02},
    issn = {1449-4035},
    doi = {10.1093/polsoc/puaf001},
    url = {https://doi.org/10.1093/polsoc/puaf001},
    eprint = {https://academic.oup.com/policyandsociety/article-pdf/44/1/1/61741705/puaf001.pdf},
}

@article{KRAKOWSKI2025100560,
title = {Human-AI agency in the age of generative AI},
journal = {Information and Organization},
volume = {35},
number = {1},
pages = {100560},
year = {2025},
issn = {1471-7727},
doi = {https://doi.org/10.1016/j.infoandorg.2025.100560},
url = {https://www.sciencedirect.com/science/article/pii/S1471772725000065},
author = {Sebastian Krakowski}
}

@misc{dohmatob2024strongmodelcollapse,
      title={Strong Model Collapse}, 
      author={Elvis Dohmatob and Yunzhen Feng and Arjun Subramonian and Julia Kempe},
      year={2024},
      eprint={2410.04840},
      archivePrefix={arXiv},
      primaryClass={cs.LG},
      url={https://arxiv.org/abs/2410.04840}, 
}

@misc{kazdan2025collapsethriveperilspromises,
      title={Collapse or Thrive? Perils and Promises of Synthetic Data in a Self-Generating World}, 
      author={Joshua Kazdan and Rylan Schaeffer and Apratim Dey and Matthias Gerstgrasser and Rafael Rafailov and David L. Donoho and Sanmi Koyejo},
      year={2025},
      eprint={2410.16713},
      archivePrefix={arXiv},
      primaryClass={cs.LG},
      url={https://arxiv.org/abs/2410.16713}, 
}

@article{10.1093/jcr/ucaf013,
    author = {Huang, Ming-Hui and Rust, Roland T},
    title = {The GenAI Future of Consumer Research},
    journal = {Journal of Consumer Research},
    volume = {52},
    number = {1},
    pages = {4-17},
    year = {2025},
    month = {05},
    issn = {0093-5301},
    doi = {10.1093/jcr/ucaf013},
    url = {https://doi.org/10.1093/jcr/ucaf013},
    eprint = {https://academic.oup.com/jcr/article-pdf/52/1/4/63186646/ucaf013.pdf},
}

@misc{geng2024impactlargelanguagemodels,
      title={The Impact of Large Language Models in Academia: from Writing to Speaking}, 
      author={Mingmeng Geng and Caixi Chen and Yanru Wu and Dongping Chen and Yao Wan and Pan Zhou},
      year={2024},
      eprint={2409.13686},
      archivePrefix={arXiv},
      primaryClass={cs.CL},
      url={https://arxiv.org/abs/2409.13686}, 
}

@misc{bereska2024mechanisticinterpretabilityaisafety,
      title={Mechanistic Interpretability for AI Safety -- A Review}, 
      author={Leonard Bereska and Efstratios Gavves},
      year={2024},
      eprint={2404.14082},
      archivePrefix={arXiv},
      primaryClass={cs.AI},
      url={https://arxiv.org/abs/2404.14082}, 
}

@misc{sharkey2025openproblemsmechanisticinterpretability,
      title={Open Problems in Mechanistic Interpretability}, 
      author={Lee Sharkey and Bilal Chughtai and Joshua Batson and Jack Lindsey and Jeff Wu and Lucius Bushnaq and Nicholas Goldowsky-Dill and Stefan Heimersheim and Alejandro Ortega and Joseph Bloom and Stella Biderman and Adria Garriga-Alonso and Arthur Conmy and Neel Nanda and Jessica Rumbelow and Martin Wattenberg and Nandi Schoots and Joseph Miller and Eric J. Michaud and Stephen Casper and Max Tegmark and William Saunders and David Bau and Eric Todd and Atticus Geiger and Mor Geva and Jesse Hoogland and Daniel Murfet and Tom McGrath},
      year={2025},
      eprint={2501.16496},
      archivePrefix={arXiv},
      primaryClass={cs.LG},
      url={https://arxiv.org/abs/2501.16496}, 
}

@misc{saini2026bridgingmechanisticinterpretabilityprompt,
    title={Bridging Mechanistic Interpretability and Prompt Engineering with Gradient Ascent for Interpretable Persona Control},
    author={Harshvardhan Saini and Yiming Tang and Dianbo Liu},
    year={2026},
    eprint={2601.02896},
    archivePrefix={arXiv},
    primaryClass={cs.LG},
    url={https://arxiv.org/abs/2601.02896},
}

@misc{tang2023integratedforwardforwardalgorithmintegrating,
title={The Integrated Forward-Forward Algorithm: Integrating Forward-Forward and Shallow Backpropagation With Local Losses},
author={Desmond Y. M. Tang},
year={2023},
eprint={2305.12960},
archivePrefix={arXiv},
primaryClass={[cs.NE](http://cs.ne/)},
url={https://arxiv.org/abs/2305.12960},
}

@misc{anumasa2026navigatingheterogeneousproteinlandscapes,
    title={Navigating heterogeneous protein landscapes through geometry-aware smoothing},
    author={Srinivas Anumasa and Barath Chandran and Tingting Chen and Nuwaisir Mohammad Rahman and Yingtao Zhu and Rushi Shah and Hongyu He and Peisong Zhang and Yizhen Liao and Yiming Tang and Yong Shen and Tianfan Fu and Rui Qing and Xiao Li and Sebastian Maurer-Stroh and Xinyi Su and Zhizhuo Zhang and Dianbo Liu},
    year={2026},
    eprint={2602.10422},
    archivePrefix={arXiv},
    primaryClass={cs.CE},
    url={https://arxiv.org/abs/2602.10422},
}

@misc{dai2025sanhypothesizinglongtermsynaptic,
    title={SAN: Hypothesizing Long-Term Synaptic Development and Neural Engram Mechanism in Scalable Model's Parameter-Efficient Fine-Tuning},
    author={Gaole Dai and Chun-Kai Fan and Yiming Tang and Zhi Zhang and Yuan Zhang and Yulu Gan and Qizhe Zhang and Cheng-Ching Tseng and Shanghang Zhang and Tiejun Huang},
    year={2025},
    eprint={2409.06706},
    archivePrefix={arXiv},
    primaryClass={[cs.NE](http://cs.ne/)},
    url={https://arxiv.org/abs/2409.06706},
}

\newpage
\renewcommand{\thesection}{\Alph{section}}
\renewcommand{\thefigure}{A\arabic{figure}}
\setcounter{section}{0}
\setcounter{figure}{0}

\section*{Appendix A: Qualitative Examples of LanSE Neurons}
\addcontentsline{toc}{section}{Appendix A}
\refstepcounter{section}
\label{appendix:example_neurons}

\begin{figure}[H]
    \centering
    \includegraphics[width=0.9\linewidth]{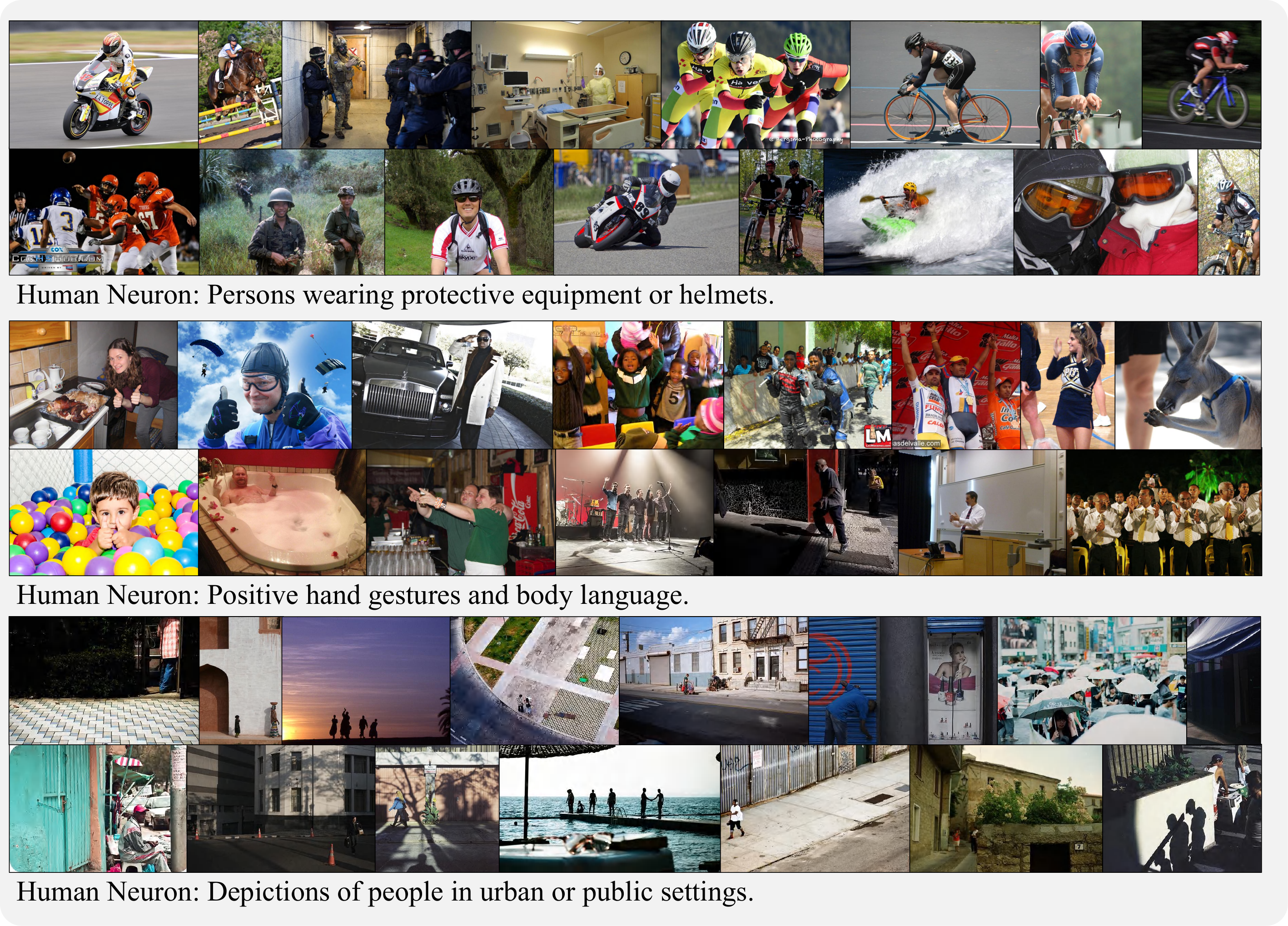}
    \caption{Visual pattern examples for the human subgroup of semantic meaning.}
    \label{fig:A1}
\end{figure}

\begin{figure}[H]
    \centering
    \includegraphics[width=0.9\linewidth]{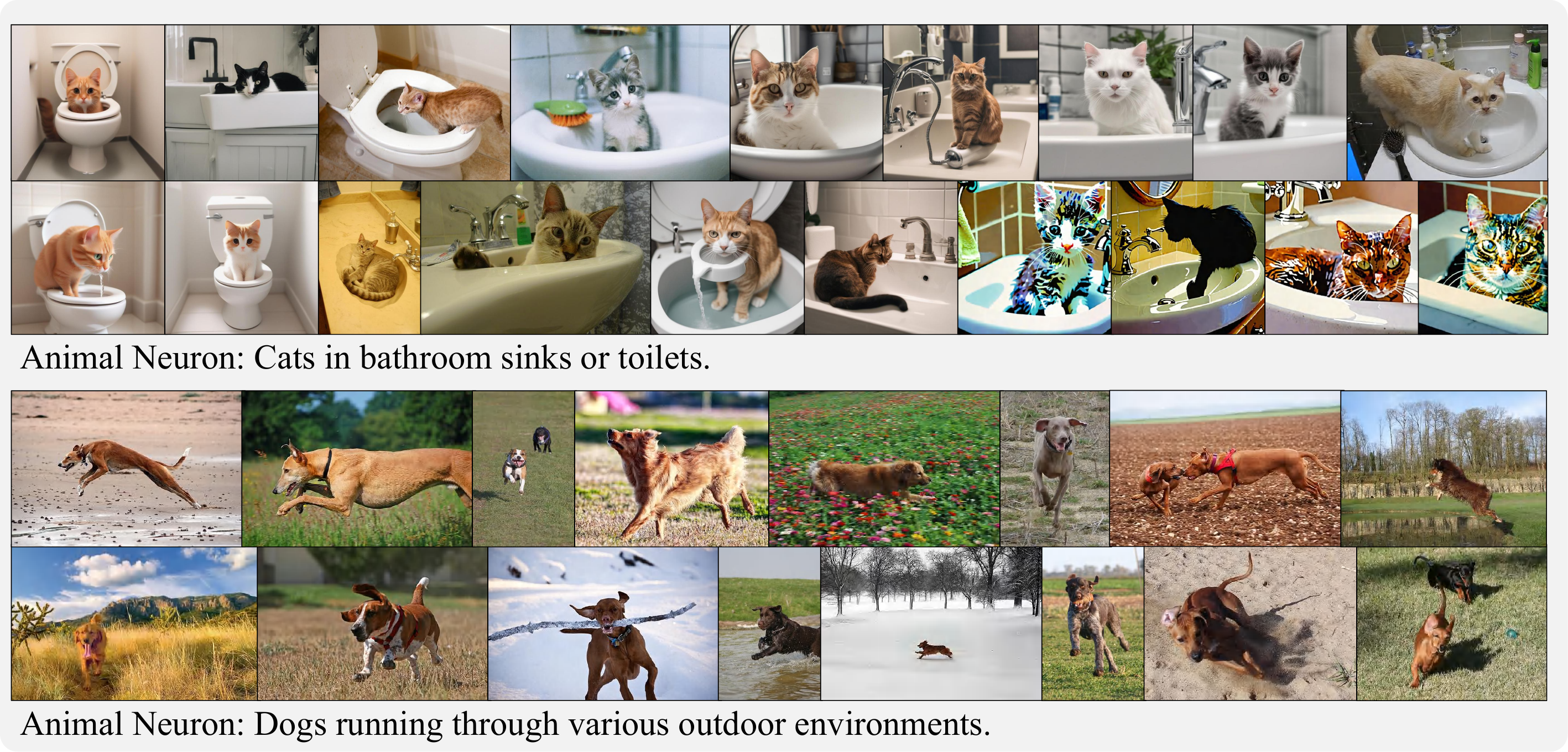}
    \caption{Visual pattern examples for the animal subgroup of semantic meaning.}
    \label{fig:A2}
\end{figure}

\begin{figure}[H]
    \centering
    \includegraphics[width=0.9\linewidth]{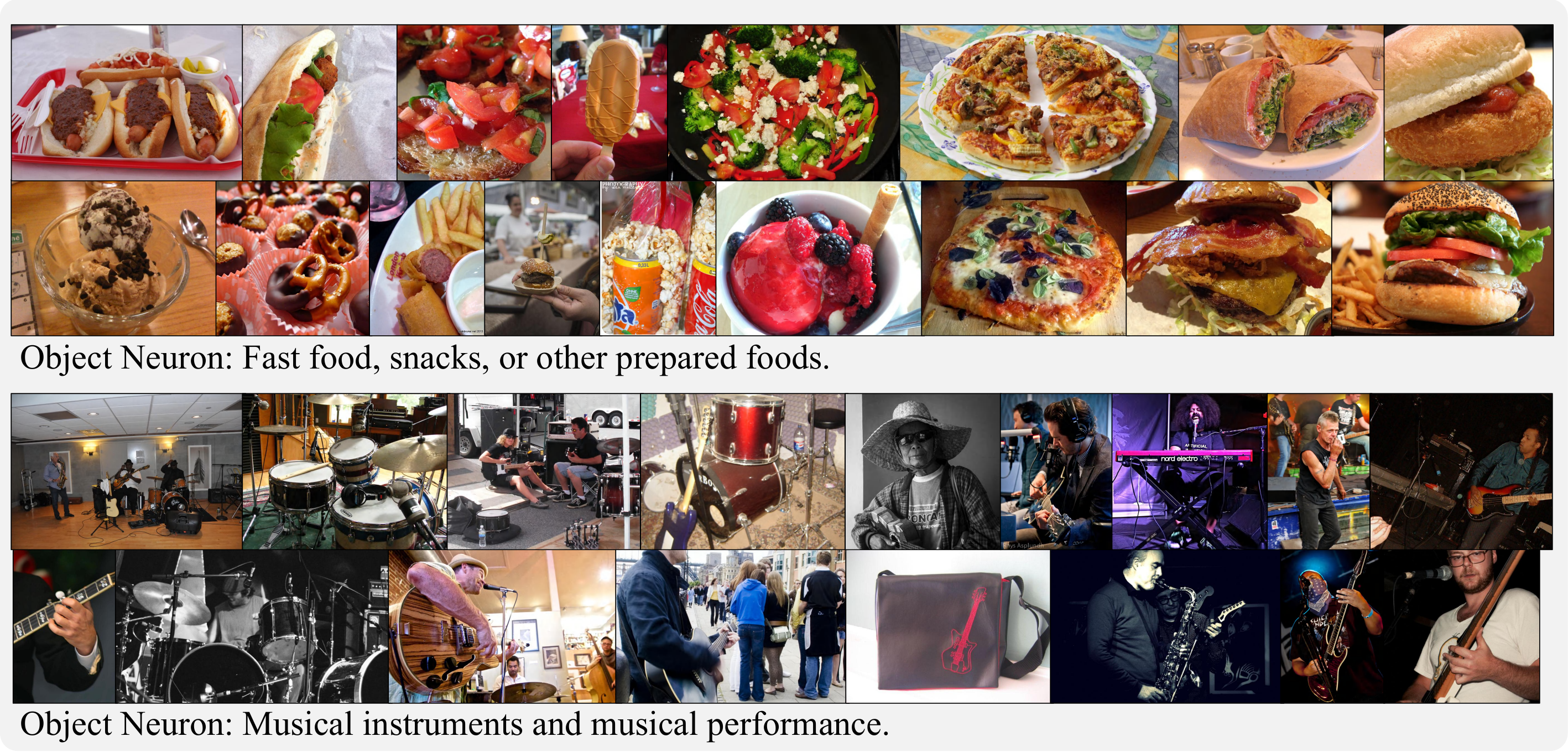}
    \caption{Visual pattern examples for the object subgroup of semantic meaning.}
    \label{fig:A3}
\end{figure}

\begin{figure}[H]
    \centering
    \includegraphics[width=0.9\linewidth]{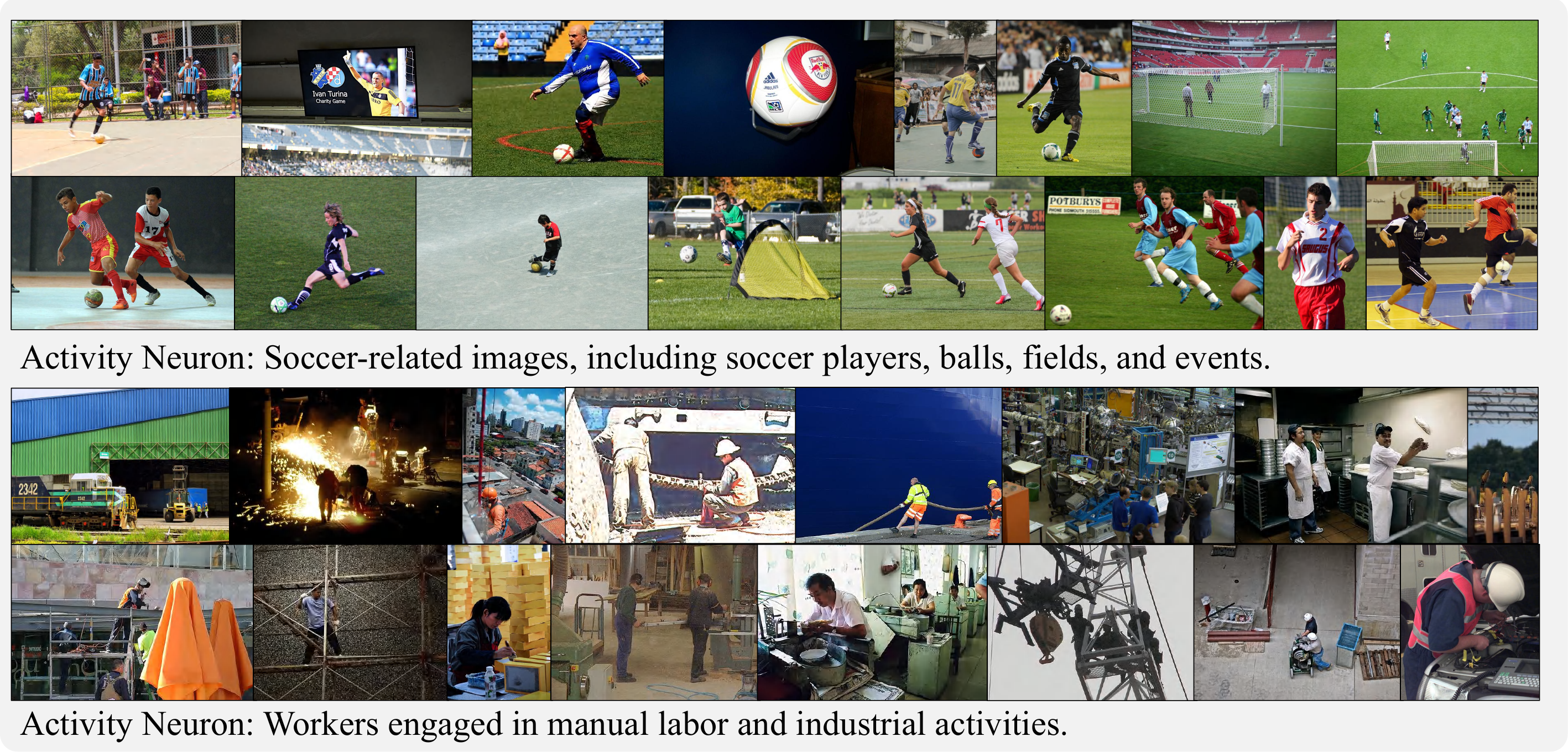}
    \caption{Visual pattern examples for the activity subgroup of semantic meaning.}
    \label{fig:A4}
\end{figure}

\begin{figure}[H]
    \centering
    \includegraphics[width=0.9\linewidth]{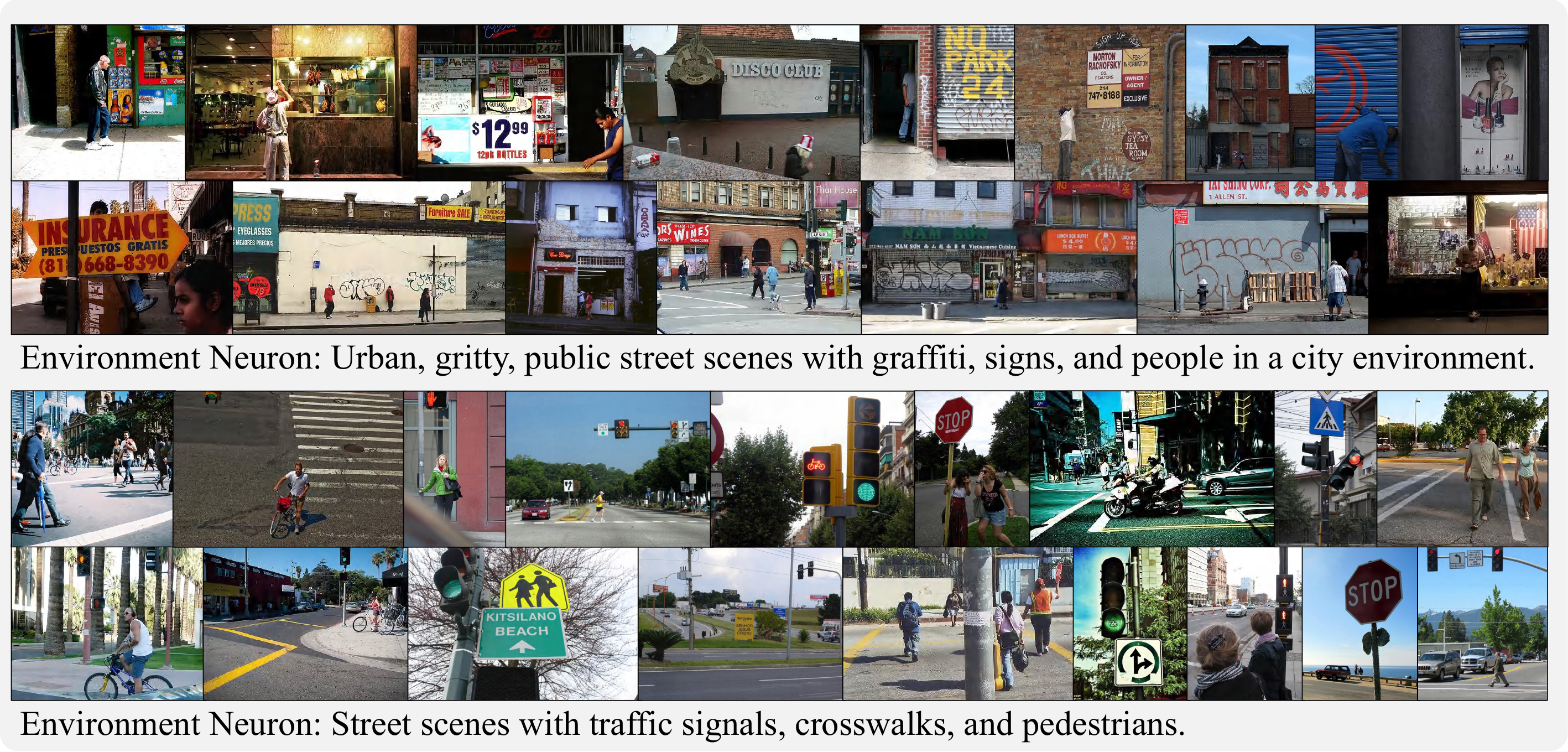}
    \caption{Visual pattern examples for the environment subgroup of semantic meaning.}
    \label{fig:A5}
\end{figure}

\begin{figure}[H]
    \centering
    \includegraphics[width=0.9\linewidth]{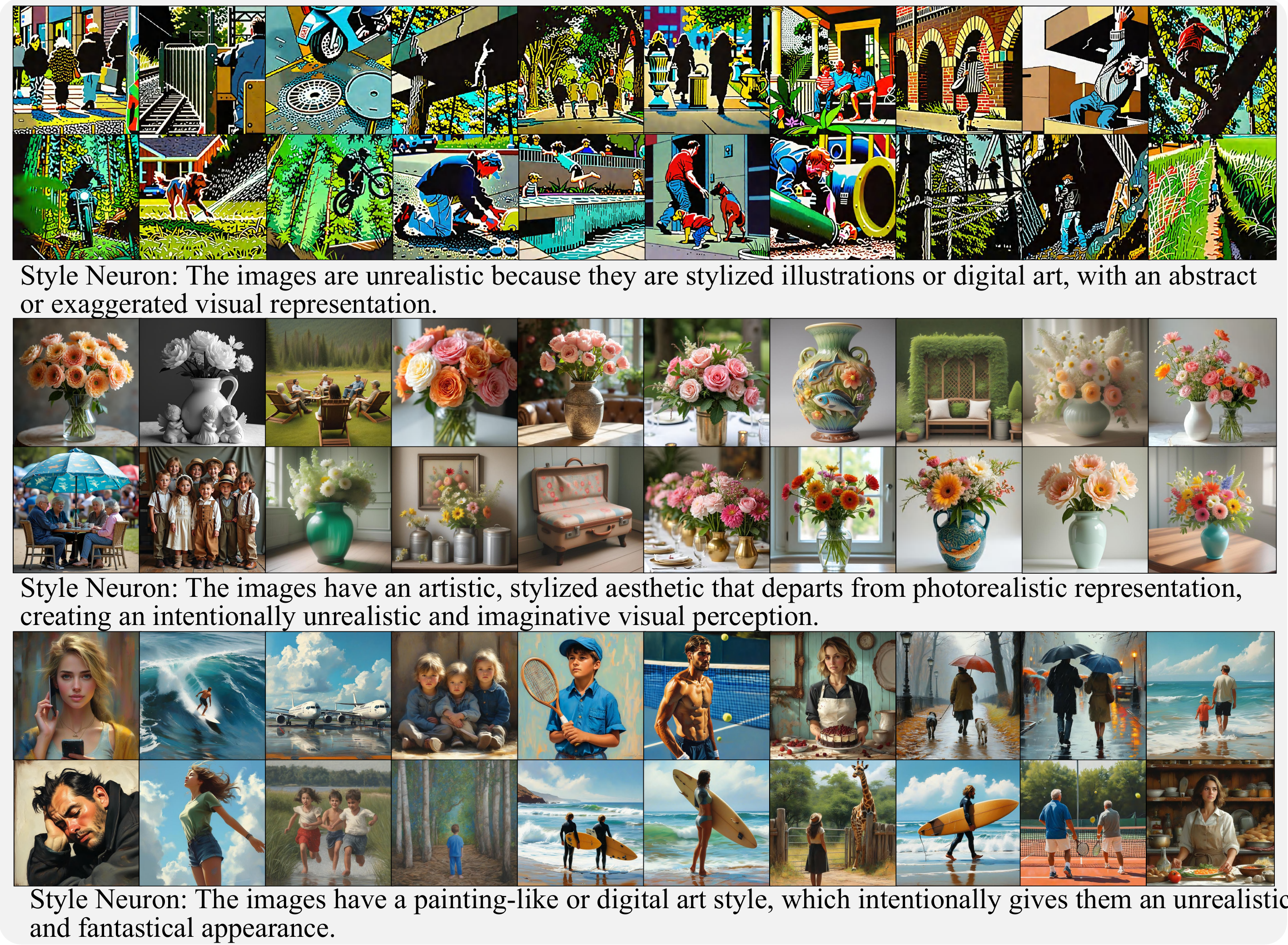}
    \caption{Visual pattern examples for the style subgroup of visual realism.}
    \label{fig:A6}
\end{figure}

\begin{figure}[H]
    \centering
    \includegraphics[width=0.9\linewidth]{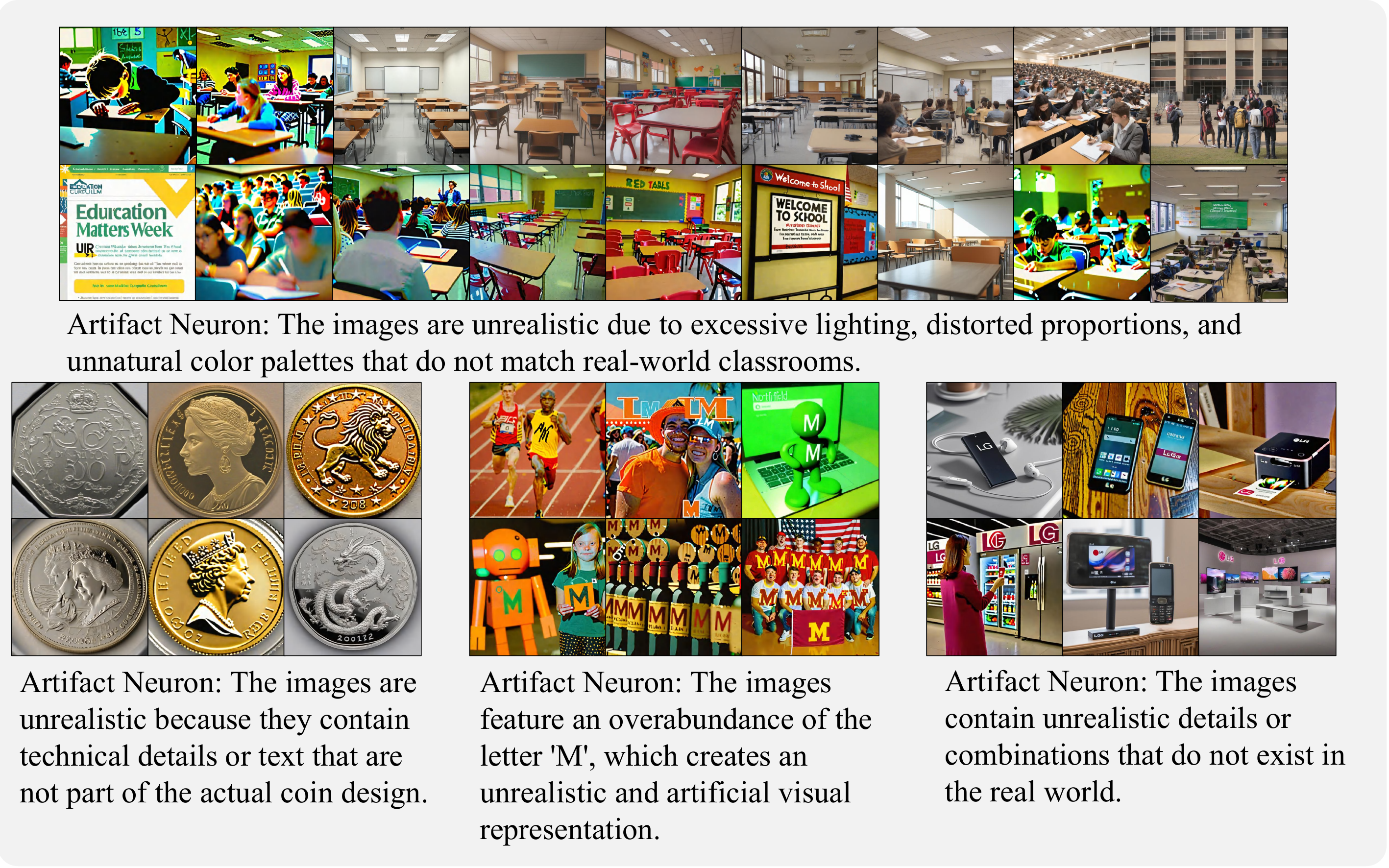}
    \caption{Visual pattern examples for the artifact subgroup of visual realism.}
    \label{fig:A7}
\end{figure}

\begin{figure}[H]
    \centering
    \includegraphics[width=\linewidth]{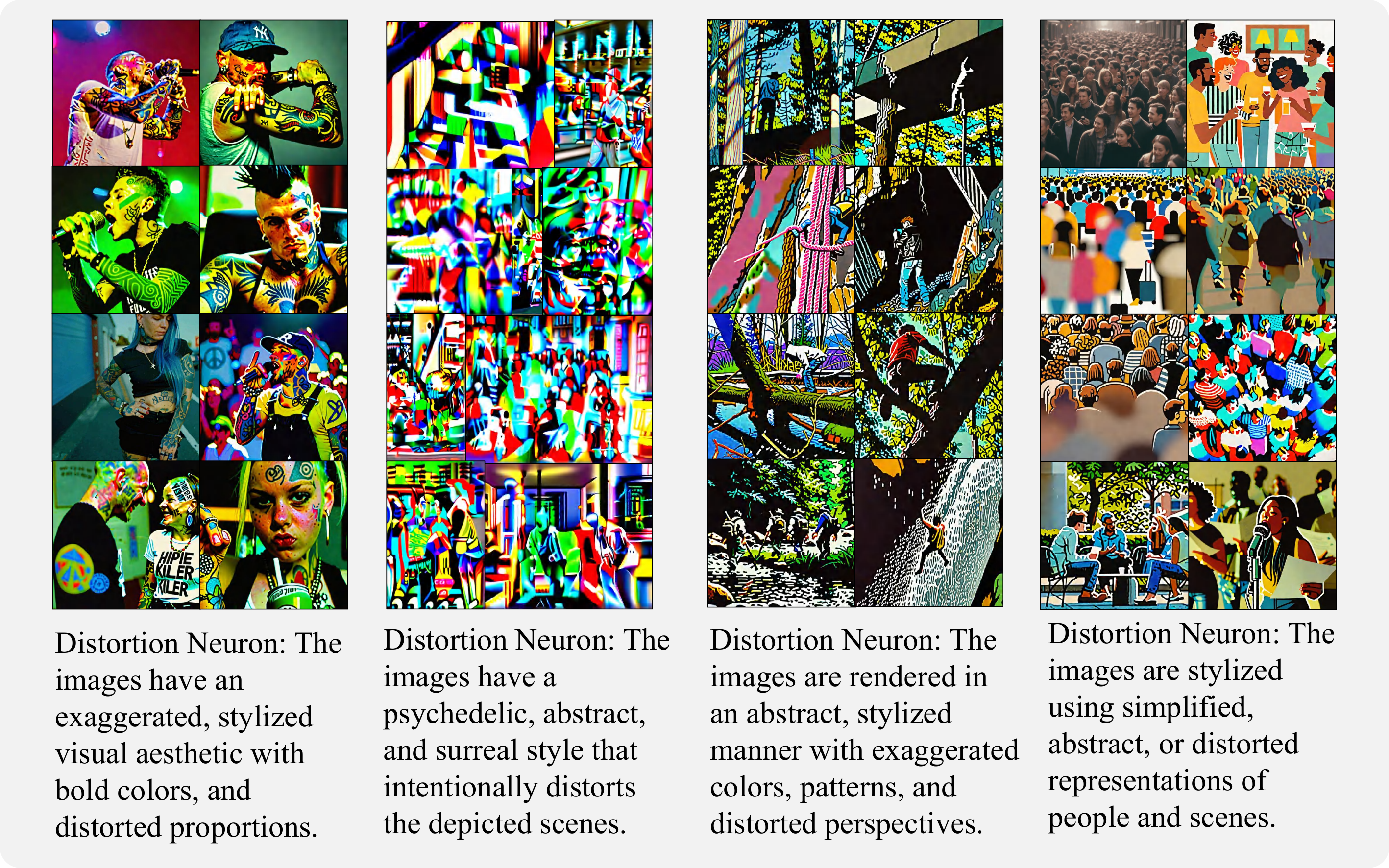}
    \caption{Visual pattern examples for the distortion subgroup of physical plausibility.}
    \label{fig:A8}
\end{figure}

\begin{figure}[H]
    \centering
    \includegraphics[width=\linewidth]{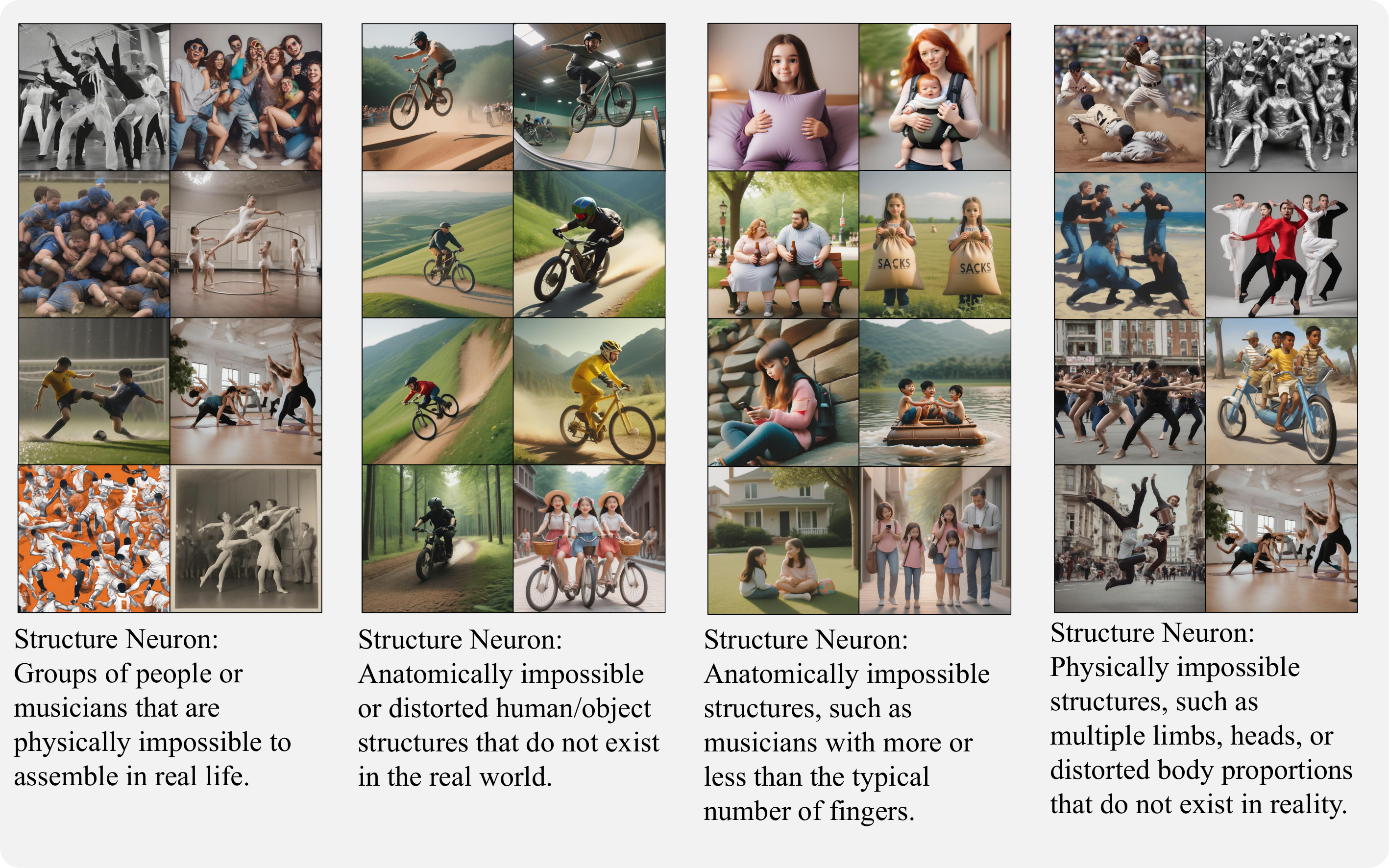}
    \caption{Visual pattern examples for the structure subgroup of physical plausibility.}
    \label{fig:A9}
\end{figure}

\begin{figure}[H]
    \centering
    \includegraphics[width=\linewidth]{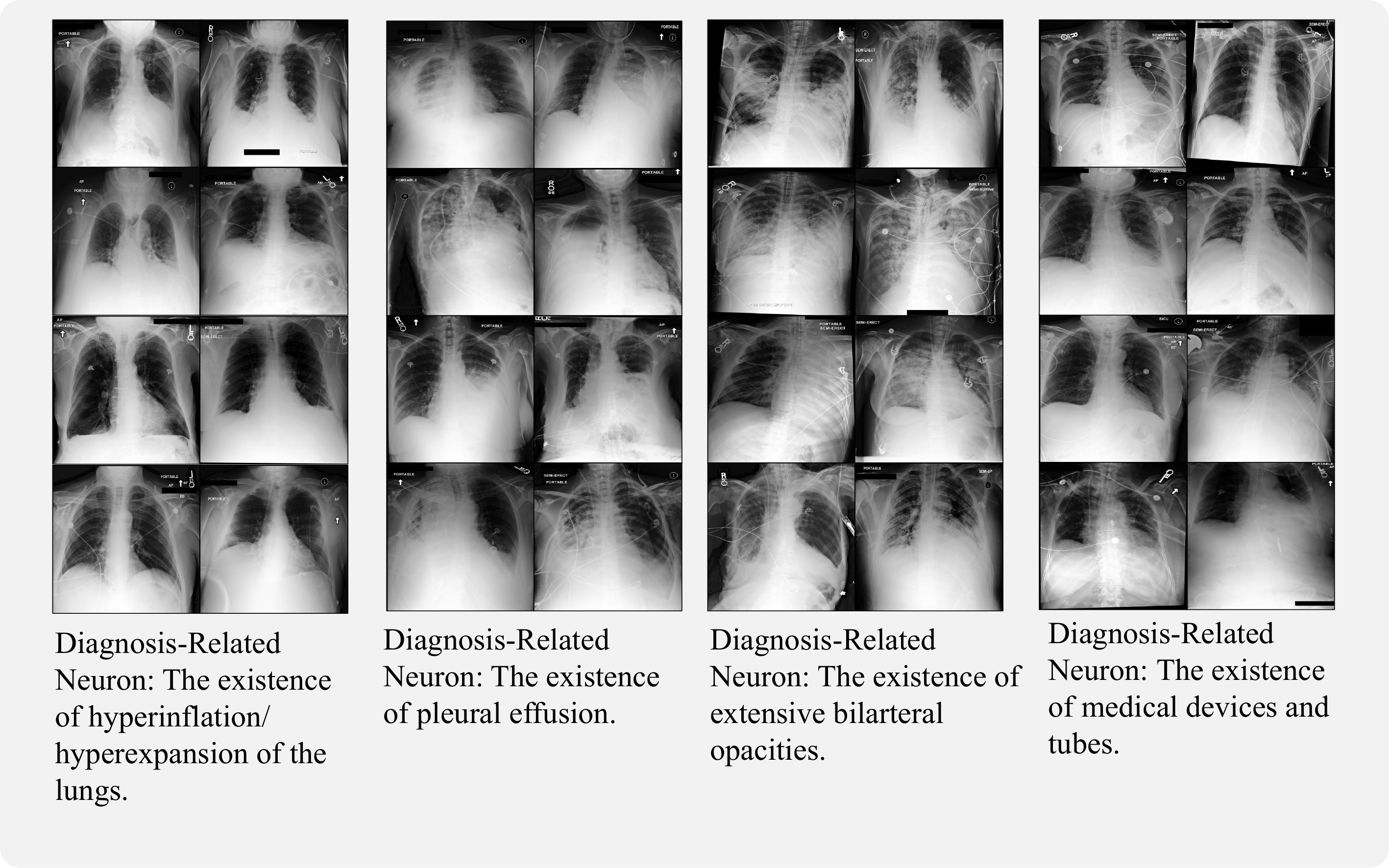}
    \caption{Visual pattern examples for the CXR-LanSE built on chest X-ray images.}
    \label{fig:A10}
\end{figure}

\noindent In this section, we provide qualitative examples of the interpretable visual patterns used in our evaluation method, LanSE (Figs.~\ref{fig:A1}--\ref{fig:A10}). These visualizations contain individual visual patterns that encodes meaningful explanations and the set of images that can activate them, including both photos and generated images.
\vspace{-0.3cm}
\vspace{-0.7cm}

\vspace{-1cm}
\vspace{-1cm}

\vspace{-0.5cm}

\vspace{-0.5cm}

\newpage
\section*{Appendix B. Qualitative Examples of Image-caption Pairs and Activated Neurons}
\addcontentsline{toc}{section}{Appendix B}
\refstepcounter{section}
\label{appendix:example_images}

\begin{figure}[H]
    \centering
    \includegraphics[width=0.97\linewidth]{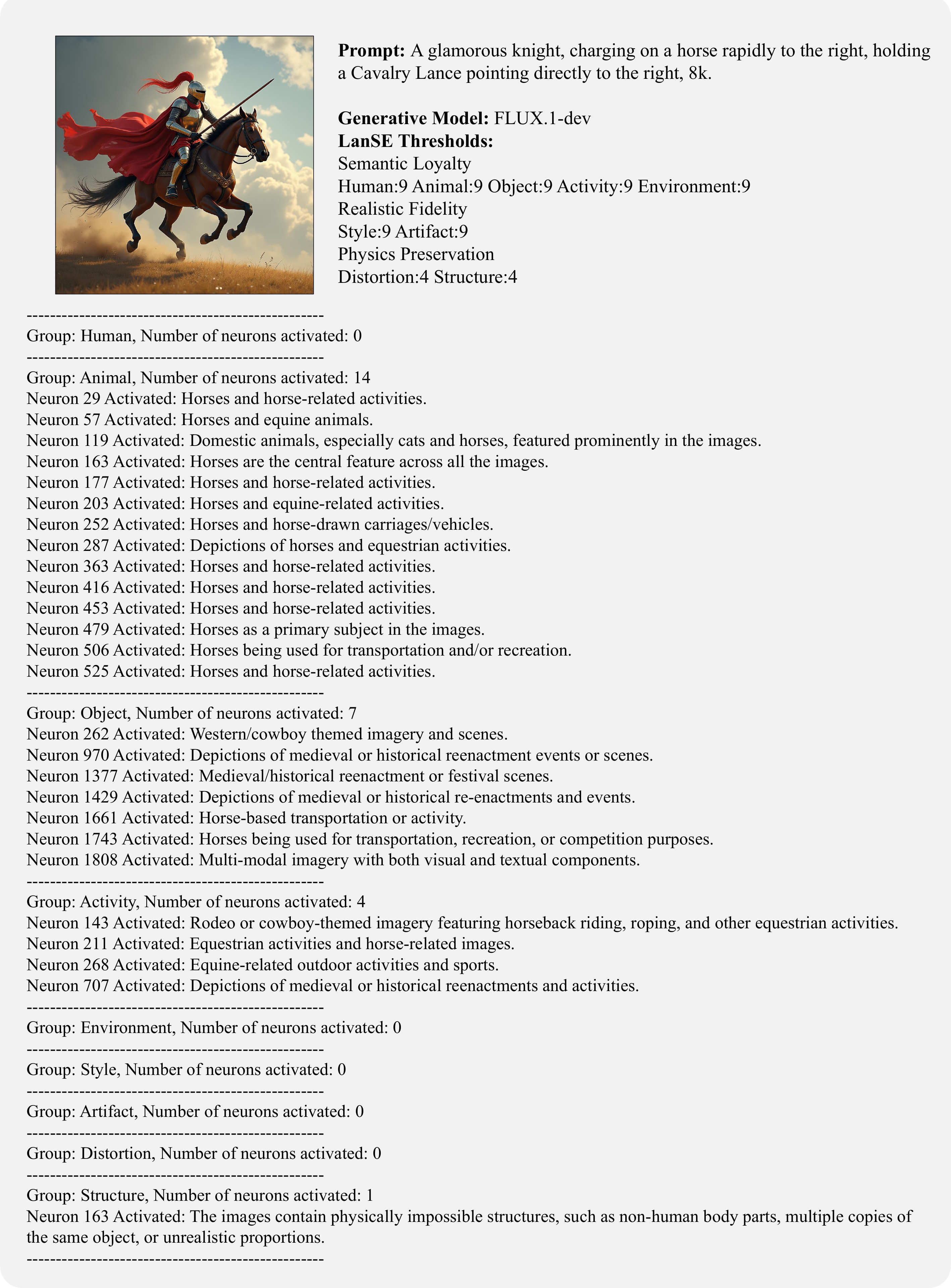}
    \caption{Image example and the LanSE neurons it highly activates.}
    \label{fig:B1}
\end{figure}
\begin{figure}[H]
    \centering
    \includegraphics[width=0.97\linewidth]{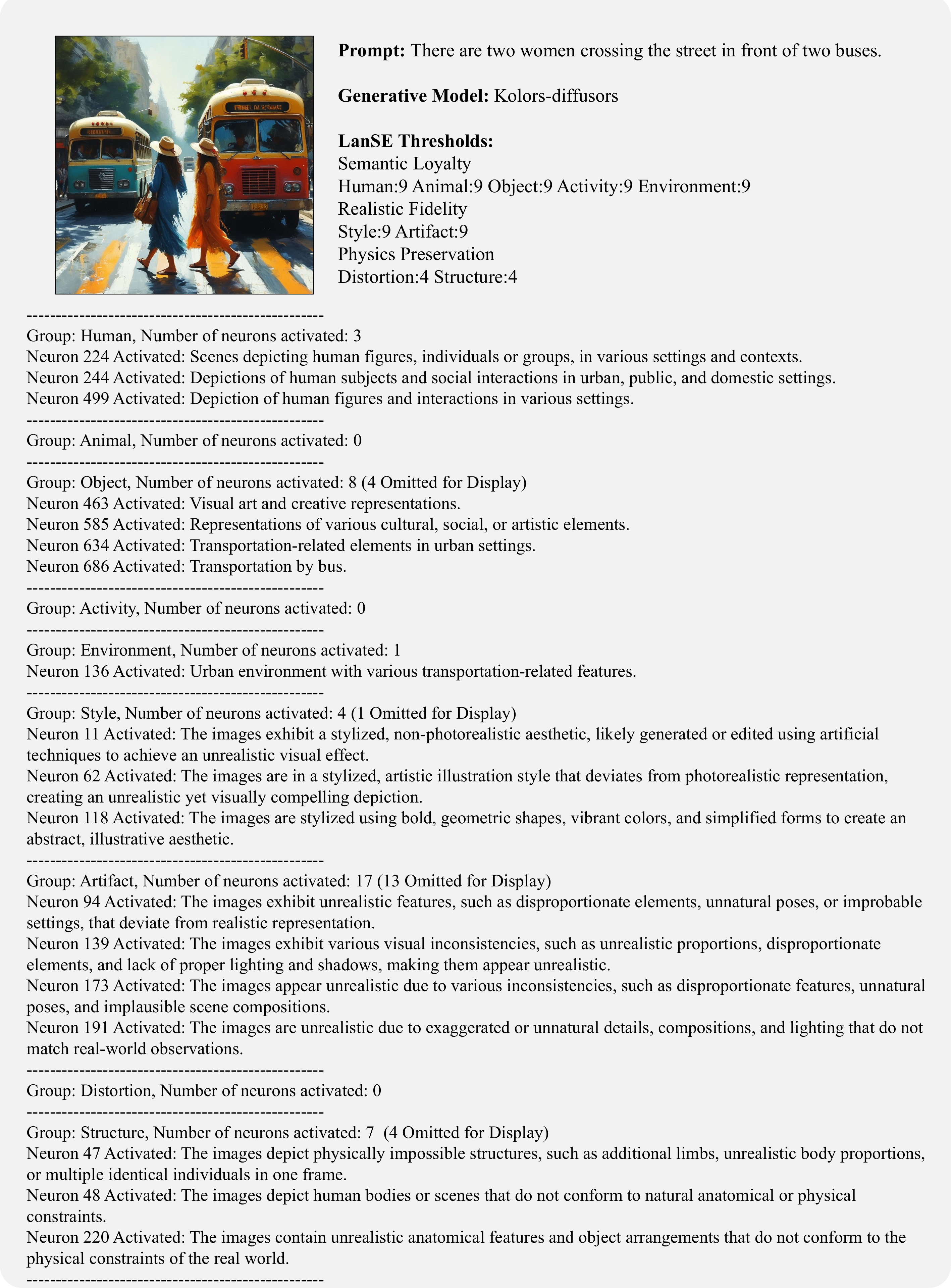}
    \caption{Image example and the LanSE neurons it highly activates.}
    \label{fig:B2}
\end{figure}

\begin{figure}[H]
    \centering
    \includegraphics[width=0.97\linewidth]{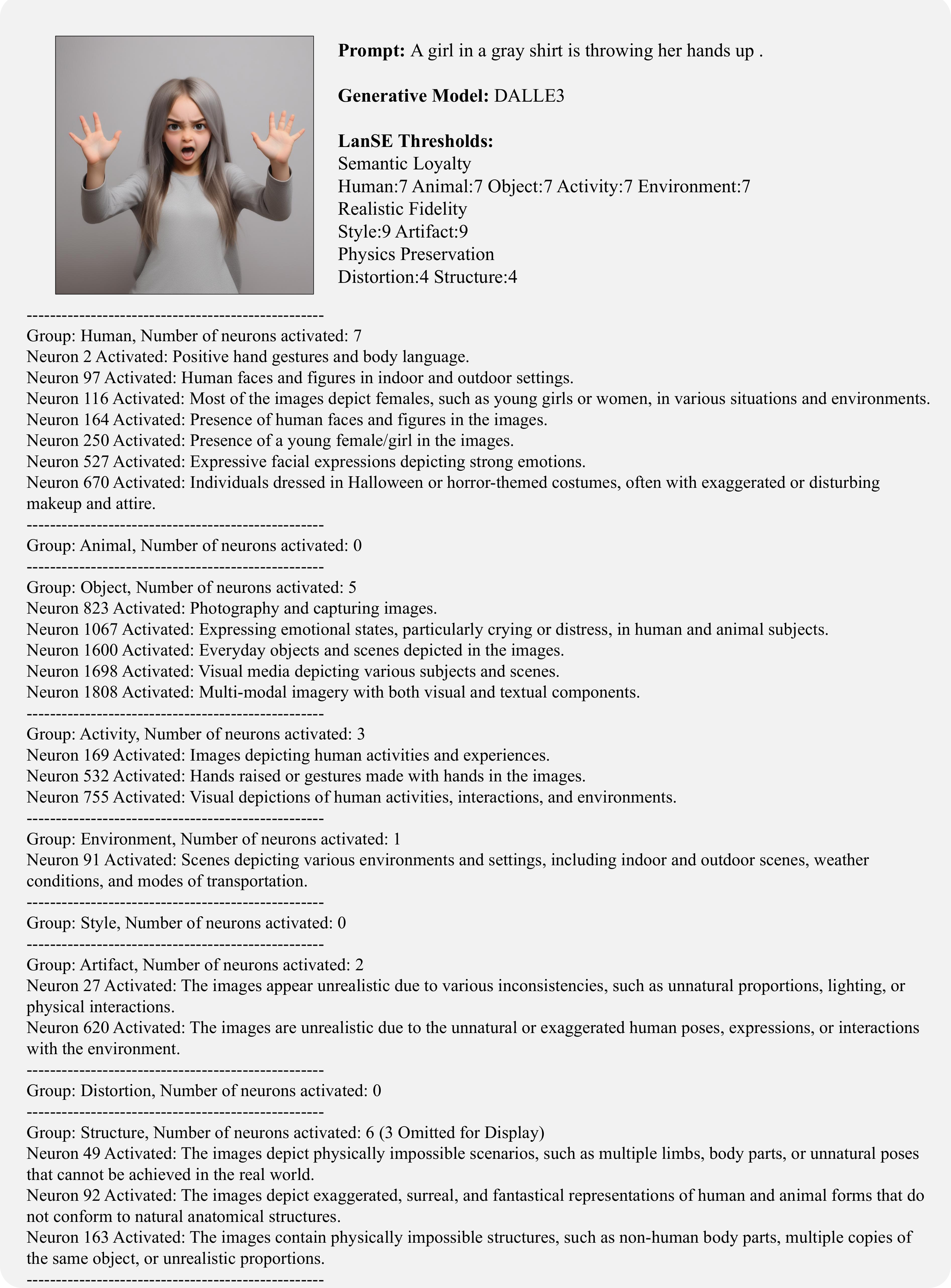}
    \caption{Image example and the LanSE neurons it highly activates.}
    \label{fig:B3}
\end{figure}
\begin{figure}[H]
    \centering
    \includegraphics[width=0.97\linewidth]{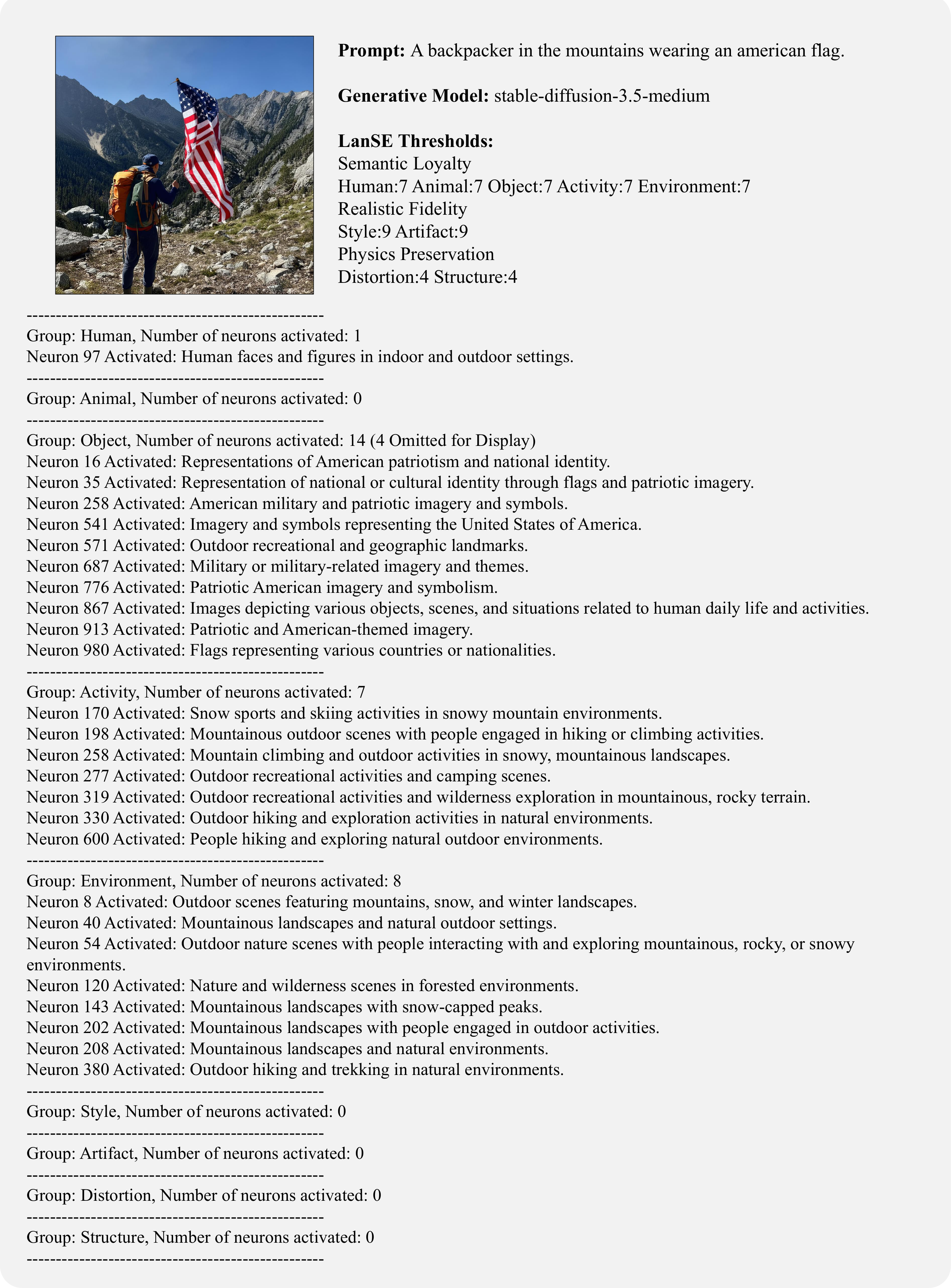}
    \caption{Image example and the LanSE neurons it highly activates.}
    \label{fig:B4}
\end{figure}

\newpage
\section*{Appendix C. LMM Analysis Prompts}
\addcontentsline{toc}{section}{Appendix C}
\refstepcounter{section}
\label{appendix:llm_analysis}

In this Appendix, we list the prompts we have used in our pipelines.

We use the following prompt to summarize the common patterns among one given neuron and its corresponding subpopulation that can highly activate it.
\begin{tcolorbox}[title=Pattern Summarization Prompt,fonttitle=\bfseries,colback=gray!5!white,colframe=gray!75!black]
You are an expert in visual pattern analysis.

\{(image)(caption)(image)(caption)...\}

Analyze the commonalities among these samples. Identify: If there exist one common visual pattern that is possessed by all the instances. Summarize and output exactly one visual pattern, for example: '[Commonality: Animal wildlife in natural habitats]' and '[Commonality: Strawberry-based dessert or dish]'. Only answer in general, do not analyze each image one by one, only generate one single concise phrase. Start answer with '[Commonality:'. End with ']'
\end{tcolorbox}

We use the following prompt to assign the semantic meaning neurons to their corresponding neuron groups, human, animal, object, activity or environment.
\begin{tcolorbox}[title=Semantics Pattern Categorization Prompt,fonttitle=\bfseries,colback=gray!5!white,colframe=gray!75!black]
You are a helpful assistant categorizing visual patterns into exactly one of the following groups: "[human, animal, object, activity, environment]."
 
Given the commonality description: {commonality}, which of the five categories does it best belong to? Please answer with a single word from this list: [human, animal, object, activity, environment]. Wrap your answer in [].
\end{tcolorbox}

We use the following prompt to assign the visual realism neurons to their corresponding neuron groups, style or artifact.
\begin{tcolorbox}[title=Realism Pattern Categorization Prompt,fonttitle=\bfseries,colback=gray!5!white,colframe=gray!75!black]
Here is a set of image samples along with their captions:

\{(image)(caption)(image)(caption)...\}

They are probably unrealistic for certain reason.

One type of reason is the style of the images is intended to be unrealistic, like cartoonish or stamp-like, or painting-like.
Another type of reason is visual artifact, like a photo of a real scene but is unrealistic because of the violation of conventions or details.

Please answer in the following format:
[Style, Explanation: \{explanation\}]
or
[Artifact, Explanation: \{explanation\}]
Only give me one answer that is common to all the images and can explain all of them, the explanation should be as short as possible, no more than 20 words.
\end{tcolorbox}

We use the following prompt to assign the physical plausibility neurons to their corresponding neuron groups, distortion or structure.
\begin{tcolorbox}[title=Physics Pattern Categorization Prompt,fonttitle=\bfseries,colback=gray!5!white,colframe=gray!75!black]
Here is a set of image samples along with their captions:

\{(image)(caption)(image)(caption)...\}

They all possess one common visual pattern that make them physically impossible.

One type of visual patterns is the distortion of the object surface, like distorted faces or blurred surfaces.
Another type of visual patterns is the structure of the object, like more than six fingers on a hand, one head on a single body or three legs on a single human.

What is the common visual pattern that make them physically impossible? Please answer in the following format:
[Distortion, Explanation: \{explanation\}]
or
[Structure, Explanation: \{explanation\}]
Only give me one answer that is common to all the images and can explain all of them, the explanation should be as short as possible, no more than 20 words.
\end{tcolorbox}

Additionally, for the hallucination detection part to compare LanSE's capability with LLMs we use the following prompt.
\begin{tcolorbox}[title=Hallucination Detection Prompt,fonttitle=\bfseries,colback=gray!5!white,colframe=gray!75!black]
Analyze this image for physical plausibility errors.

Look for:
- Wrong number of body parts (fingers, limbs, heads)
- Impossible anatomical structures
- Objects defying physics
- Impossible proportions or configurations

Does this image contain physical plausibility errors? 
Answer: YES or NO

Explanation:
\end{tcolorbox}

\newpage
\section*{Appendix D. Method Extention}
\begin{figure}[H]
    \centering
    \includegraphics[width=\linewidth]{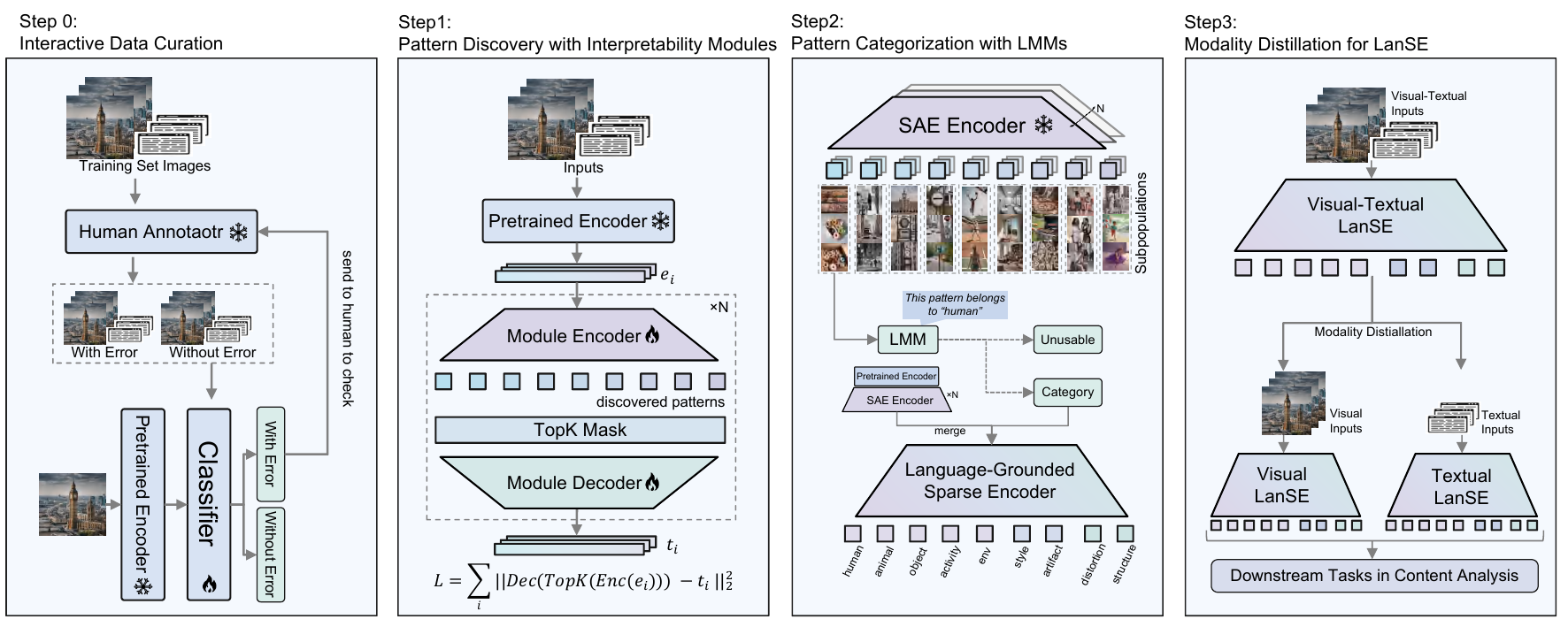}
    \caption{The general pipeline of our extended method used for discovering monosemantic neurons that selectively respond to targeted semantics (physics violations).}
    \label{fig:method_extension}
\end{figure}

\addcontentsline{toc}{section}{Appendix D}
\refstepcounter{section}
\label{appendix:method}
Certain failure modes in generative models, such as physics violations (e.g., distorted limbs, extra fingers), are both rare and critical, yet difficult to discover through random sampling. To address this, we develop an interactive human-in-the-loop framework that enables targeted visual pattern discovery at scale while reducing human annotation cost, as shown in Figure~\ref{fig:method_extension}.

\subsection*{Motivation}
While LanSE neurons are primarily derived from large-scale, sparsity-driven interpretability modules, the discovery of visual patterns depends on the availability of sufficiently frequent visual patterns in the training data. However, physics-related errors, such as anatomically impossible structures, are often of low frequency and hard to detect without guidance. Exhaustively humanly reviewing 250,000+ images is infeasible. Therefore, we build an iterative system that leverages weak supervision from lightweight classifiers to bootstrap the collection of relevant samples.

\subsection*{Interactive Data Curation Pipeline}
We begin by collecting an initial seed dataset of ~300 images containing known physics violations, along with an equal number of negative samples. Human annotators manually label each image as exhibiting physics-consistent or physics-violating content. These labels are then used to train a lightweight two-layer binary classifier on CLIP image embeddings. The classifier architecture is:
\begin{itemize}
    \item Input: CLIP image embedding.
    \item Hidden: Fully connected ReLU layer.
    \item Output: Sigmoid activation for binary label (violation / no violation).
\end{itemize}
Once trained, this classifier is used to scan the full image dataset and flag potentially erroneous images. These flagged samples is then sent to human annotators for relabeling. The updated labels are added to the training pool, and the classifier is retrained. This bootstrapping process continues iteratively, improving both coverage and precision of error discovery.

This approach enables the discovery of rare but critical visual patterns in a scalable manner, cutting down the annotation cost by an order of magnitude compared to exhaustive search.

\subsection*{Integration into LanSE}
After sufficient labeled data is accumulated through this pipeline, we train a sparse transcoder on the intermediate representations of the two-layer classifier. Specifically, we extract the hidden activations and train a sparse autoencoder over them using the same top-k sparsity constraints as in the main pipeline.

This generates a new set of interpretable neurons, which are curated via the same LMM-based explanation and categorization pipeline. These neurons form the distortion and structure groups within LanSE's physical plausibility dimension. By grounding these rare visual pattern detectors in both language and visual representations, LanSE gains the capacity to diagnose physics violations that would otherwise remain underrepresented.

One thing should be mentioned is that this interactive visual pattern discovery pipeline is based on the intuition that task-related patterns are more prevalent in task-related models. In particular, visual patterns that is related to physics violations are more likely to be discovered in a model that is used to classify images according to whether they have physics violations or not.

\newpage
\section*{Appendix E. Mathematical Definitions of the Four Metrics}
\addcontentsline{toc}{section}{Appendix E}
\refstepcounter{section}
\label{appendix:metrics}

Here are our mathematical definition and calculative derivations of the four diagnostic metrics, prompt match, visual realism, physical plausibility and content diversity.

\textbf{Prompt Match}:
This metric captures semantic mismatches between generated images and prompts, such as missing or extraneous content. Grouped into five semantic categories, the metric is computed as the average number of neurons that diverge across modalities. Higher values indicate mroe prompt mismatches. $\mathbf{prompt}$ can be altered with $\mathbf{human}$, $\mathbf{animal}$, $\mathbf{object}$, $\mathbf{activity}$, or $\mathbf{environment}$.

\begin{align}
    \mathcal{M}_{\mathbf{prompt}}(\mathcal{X}, \mathcal{Y}) 
    &= \mathbb{E}_{\mathbf{D}}||\mathrm{N}_{\mathbf{prompt}}^{\mathbf{v}}(\mathbf{x}) 
    \oplus \mathrm{N}_{\mathbf{prompt}}^{\mathbf{t}}(\mathbf{y})||_1 \notag \\
    &= \frac{1}{N} \sum_{\mathbf{D}} 
    \left\| \mathbf{I}\{\mathrm{S}_{\mathbf{prompt}}^{\mathbf{v}}(\mathbf{x_i}) > \tau\} - 
    \mathbf{I}\{\mathrm{S}_{\mathbf{prompt}}^{\mathbf{t}}(\mathbf{y_i}) > \tau\} \right\|_1.
\end{align}

\textbf{Visual Realism}:
This metric measures visual patterns that make the images unrealistic, such as special artistic styles or visual artifacts. Neurons are grouped into style and artifact subcategories. Higher value reflects more deviation from realistic appearances. We differentiate the style group and the artifact group because some artistic styles can be intentional. Higher scores indicate the images are more unrealistic. $\mathbf{real}$ can be altered with $\mathbf{style}$ or $\mathbf{artifact}$.

\begin{equation}
    \mathcal{M}_{\mathbf{real}}(\mathcal{X},\mathcal{Y}) = \mathbb{E}_{\mathbf{D}}||\mathrm{N}_{\mathbf{real}}(\mathbf{x},\mathbf{y})||_1 = \frac{1}{N} \sum_\mathbf{D}  \left\|\mathbf{I}\{\mathrm{S_{\mathbf{real}}}(\mathbf{x_i}, \mathbf{y_i}) > \tau\}\right\|_1.
\end{equation}

\textbf{Physical Plausibility}:
This metric detects intuitively erroneous visual patterns that can make an image physically implausible and unwanted, including distortions of the object surfaces or implausible structures like erroneous anatomies. Higher value suggests more errors generated. $\mathbf{phy}$ can be altered with $\mathbf{distortion}$ or $\mathbf{structure}$.
\begin{equation}
    \mathcal{M}_{\mathbf{phy}}(\mathcal{X},\mathcal{Y}) = \mathbb{E}_{\mathbf{D}}||\mathrm{N}_{\mathbf{phy}}^{\mathbf{v}}(\mathbf{x})||_1 = \frac{1}{N} \sum_\mathbf{D} \left\|\mathbf{I}\{\mathrm{S_{\mathbf{phy}}}(\mathbf{x_i}) > \tau\}\right\|_1.
\end{equation}

\textbf{Content Diversity}:
This metric measures how diverse a generative model's outputs are. Neurons in the five semantic meaning groups and the style group are considered as content neurons, and based on them we compute the average divergence of neurons given two image samples. Higher value suggests better model diversity. Expectation is computed on images ($(\mathbf{x_i}, \mathbf{y_i})$) in the distribution with $|\mathrm{N}_{\mathbf{con}}(\mathbf{x_i}, \mathbf{y_i})|\neq 0$. $\mathbf{con}$ can be altered with $\mathbf{human}$, $\mathbf{animal}$, $\mathbf{object}$, $\mathbf{activity}$, $\mathbf{environment}$ or $\mathbf{style}$.
\begin{align}
    \mathcal{M}_{\mathbf{con}}(\mathcal{X}, \mathcal{Y}) 
    &= \mathbb{E}_{\mathbf{D}} \frac{||\mathrm{N}_{\mathbf{con}}(\mathbf{x_i}, \mathbf{y_i}) 
    \oplus \mathrm{N}_{\mathbf{con}}(\mathbf{x_j}, \mathbf{y_j})||_1}
    {||\mathrm{N}_{\mathbf{con}}(\mathbf{x_i}, \mathbf{y_i})||_1 \cdot 
    ||\mathrm{N}_{\mathbf{con}}(\mathbf{x_j}, \mathbf{y_j})||_1} \notag \\
    &= \frac{1}{N} \sum_{\mathbf{D}} 
    \frac{1 - \left\| \mathbf{I}\{\mathrm{S}_{\mathbf{con}}(\mathbf{x_i}, \mathbf{y_i}) > \tau\} \cdot 
    \mathbf{I}\{\mathrm{S}_{\mathbf{con}}(\mathbf{x_j}, \mathbf{y_j}) > \tau\} \right\|_1}
    {\left\| \mathbf{I}\{\mathrm{S}_{\mathbf{con}}(\mathbf{x_i}, \mathbf{y_i}) > \tau\} \right\|_1 
    \cdot \left\| \mathbf{I}\{\mathrm{S}_{\mathbf{con}}(\mathbf{x_j}, \mathbf{y_j}) > \tau\} \right\|_1}.
\end{align}

\newpage
\section*{Appendix F. Human Evaluation Protocols}
\begin{figure}[H]
    \centering
    \includegraphics[width=0.9\linewidth]{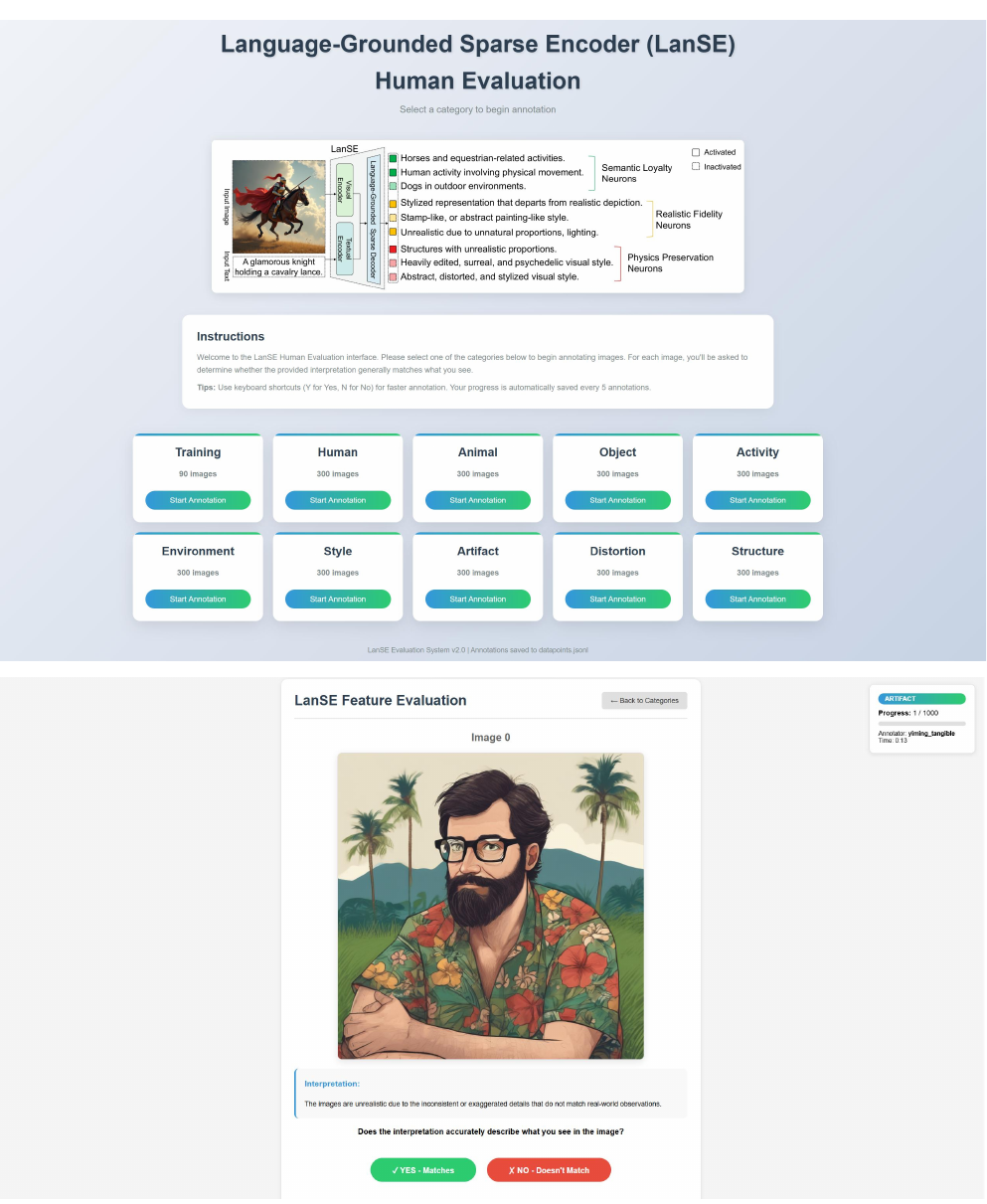}
    \caption{A demonstration of our human evaluation web interface. Human annotators are tasked to identify if the image generally matches with the LanSE interpretations.}
    \label{fig:human_evaluation_web}
\end{figure}

\addcontentsline{toc}{section}{Appendix F}
\refstepcounter{section}
\label{appendix:human_evaluation}
In this section, we detail the procedures we conduct for the human evaluations. Our human evaluation protocol for LansE's visual pattern detection accuracy is shown in Figure~\ref{fig:human_evaluation_web}. Human annotators are trained and tasked to identify if the image generally matches with the interpretation from LanSE. We release the human evaluation website for transparency: https://512d-223-109-239-18.ngrok-free.app/.

\newpage
\section*{Appendix G. The Choice of LanSE Neuron Groups}
\addcontentsline{toc}{section}{Appendix G}
\refstepcounter{section}
\label{appendix:neuron_groups}

The organization of LanSE neurons into nine specific groups reflects fundamental aspects of how practitioners evaluate generated images. When prompting a generative model, users typically assess outputs along three primary dimensions:

\textbf{1. Prompt Match}: Does the generated image match the prompt?
To evaluate this, we need to detect what semantic content appears in images versus what was requested in text. We organize semantic meaning neurons into five categories that comprehensively cover visual content:
\begin{itemize}
   \item \textbf{Human}: Detects people, faces, and human-related attributes
   \item \textbf{Animal}: Identifies animals and their characteristics
   \item \textbf{Object}: Recognizes man-made objects and items
   \item \textbf{Activity}: Captures actions, events, and dynamic scenes
   \item \textbf{Environment}: Identifies settings, locations, and backgrounds
\end{itemize}

\textbf{2. Visual Realism}: Does the image look realistic?
Unrealistic visual patterns fall into two distinct categories:
\begin{itemize}
   \item \textbf{Style}: Artistic styles or rendering techniques that deviate from photorealism (e.g., cartoon, watercolor)
   \item \textbf{Artifact}: Unintended visual artifacts that compromise image quality (e.g., noise, compression artifacts)
\end{itemize}

\textbf{3. Physical Plausibility}: Does the image contain visual patterns that make it unusable?
Critical errors that violate physical constraints include:
\begin{itemize}
   \item \textbf{Distortion}: Surface-level distortions and warping effects
   \item \textbf{Structure}: Anatomically impossible structures (e.g., extra fingers, impossible object configurations)
\end{itemize}

The category choices follow a hierarchical set of diagnostic questions. At the top level: (1) are there physical plausibility errors, and (2) does the image look realistic? These directly motivate the physical plausibility and visual realism groups. Within each, errors are further split by type — distortion vs. structure for physical plausibility, and style vs. artifact for visual realism. For semantic content, we ask: is the pattern about the environment or the main object? If the main object, is it alive or not? If alive, is it human or animal? Activity captures dynamic interactions that cut across these categories. This hierarchical process ensures the nine groups are mutually motivated and cover the evaluation dimensions practitioners care about most.

This hierarchical organization enables LanSE to provide comprehensive evaluation coverage while maintaining clear interpretability. Each group addresses specific failure modes that practitioners encounter, making the evaluation results actionable for model improvement.

We note that the chosen taxonomy is one reasonable approximation rather than a uniquely well-founded structure. Visual concepts are inherently compositional and can be ambiguous across categories, and alternative grouping schemes are equally plausible.

\newpage
\section*{Appendix H. Ablation Studies}
\addcontentsline{toc}{section}{Appendix H}
\refstepcounter{section}
\label{appendix:ablation}

To understand the influences of the hyperparameters to the LanSE method, we conduct two ablation studies on the activation threshold, $\tau$. We evaluate how the number of activated interpretable neurons changes with varying thresholds and how the accuracy of signals in each group change with varying thresholds.

\begin{figure}[H]
    \centering
    \includegraphics[width=\linewidth]{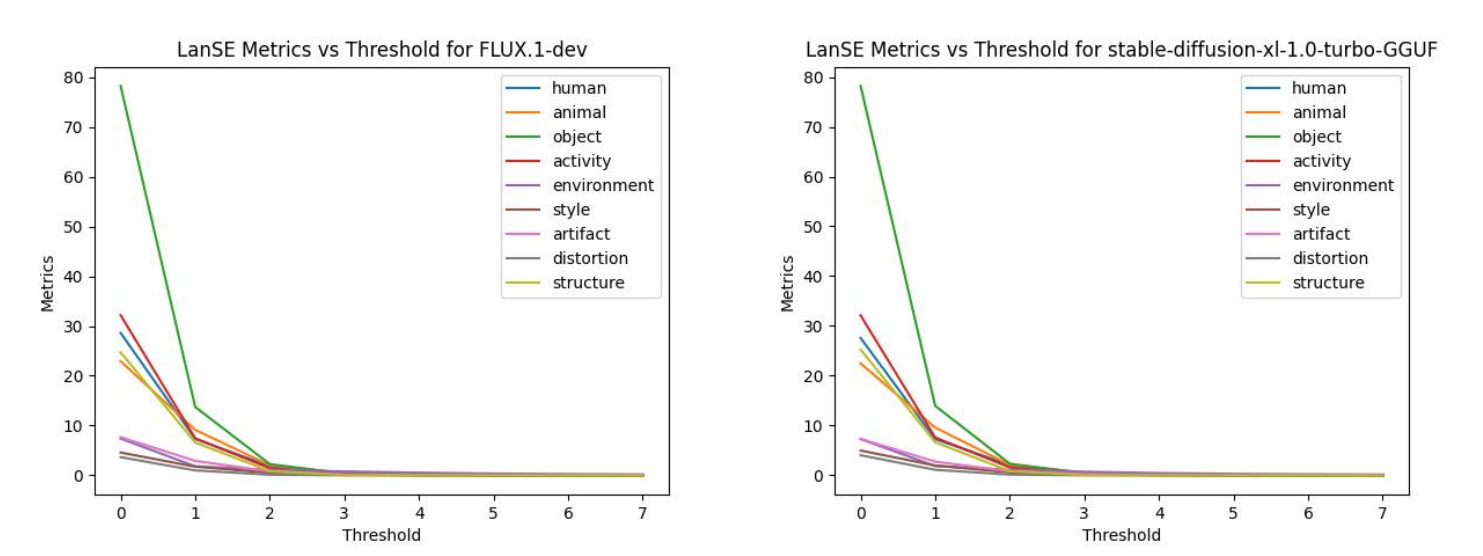}
    \caption{LanSE metrics varying according to the threshold choices for two models. Higher thresholds mean the neurons are less likely activated and leads to smaller metric values.}
    \label{fig:metric_threshold_model}
\end{figure}

\begin{figure}[H]
    \centering
    \includegraphics[width=\linewidth]{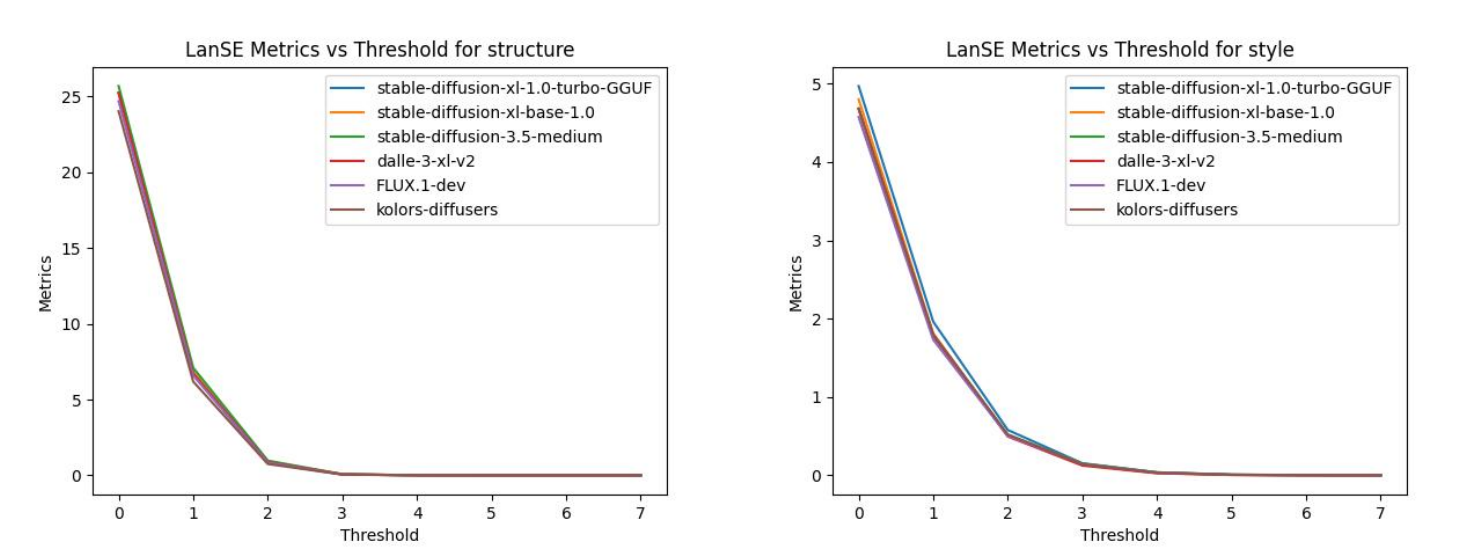}
    \caption{LanSE metrics varying according to the threshold choices for two groups. Higher thresholds mean the neurons are less likely activated and leads to smaller metric values. Lower metrics gives better contrast between models, making it easier to distinguish them.}
    \label{fig:metric_threshold_group}
\end{figure}

\textbf{Metrics v.s. the Activation Threshold $\tau$} We tune the activation threshold used for the LanSE neurons and collect all the diagnostic metrics. The results are listed in Figure~\ref{fig:metric_threshold_group} and Figure~\ref{fig:metric_threshold_model}.

\begin{figure}[H]
    \centering
    \includegraphics[width=\linewidth]{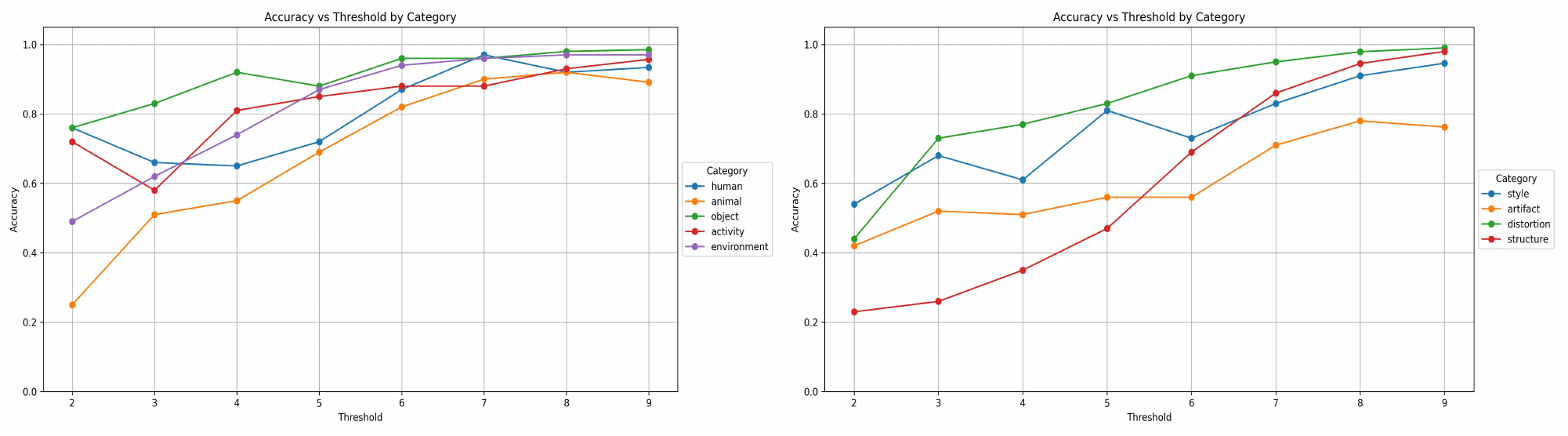}
    \caption{LanSE's signal accuracies varying according to the threshold choices. Higher thresholds means stricter constraints on the identification of visual pattern and therefore higher accuracies for the signals.}
    \label{fig:acc_threshold}
\end{figure}

\textbf{Accuracies v.s. the Activation Threshold $\tau$} We tune the activation threshold used for the LanSE neurons and utilize an LMM to evaluate the ratio of images that can activate the neurons generally match the natural language interpretations. The results are listed in Figure~\ref{fig:acc_threshold}.

We use the following prompt for the evaluation of neuron accuracies.

\begin{tcolorbox}[title=Accuracy Evaluation Prompt, fonttitle=\bfseries, colback=gray!5!white, colframe=gray!75!black]
\{(image)(explanation)\}.

Above is one image and one explanation of the image. Please check if the explanation generally matches the image. Answer only with "yes" or "no".
\end{tcolorbox}

\newpage
\section*{Appendix I. Limitations}
\addcontentsline{toc}{section}{Appendix I}
\refstepcounter{section}
\label{appendix:limitations}

\textbf{Language-Describable Visual Patterns Only.}
LanSE inherently evaluates only visual patterns that can be described in natural language, introducing an anthropocentric bias. Subtle failures or perceptually important visual patterns that lack linguistic representations may escape detection. While we argue this aligns with practical requirements, as humans primarily care about describable failures, this constraint limits our ability to capture the full spectrum of possible generation errors.

\textbf{Dependence on Base Encoder Quality.}
Our method's effectiveness is fundamentally limited by the representations learned by the underlying encoders (CLIP in our implementation). Known biases in CLIP, such as texture bias and limited cultural diversity, propagate through to LanSE evaluations. We recommend using visual encoders that encode comprehensive information and have certain semantical alignment. Future work should explore ensemble approaches using diverse encoders to mitigate these inherited biases.

\textbf{Computational and Scalability Constraints.}
Training 250 sparse autoencoders and curating 3.75 million candidate neurons requires substantial computational resources. While our filtering reduces this to 5,309 neurons, the initial discovery phase may be prohibitive for some applications. Additionally, the current architecture scales linearly with the number of evaluation aspects, potentially limiting real-time applications.

\textbf{Evaluation Scope and Coverage.}
Our method currently addresses four evaluation dimensions, leaving important aspects like aesthetic quality, cultural appropriateness, and implicit social biases unmeasured. Additionally, certain visual patterns, particularly those involving complex spatial relationships or subtle emotional expressions, may be underrepresented in our discovered visual patterns, creating evaluation blind spots.

\textbf{Context-Insensitive Evaluation.}
LanSE cannot distinguish between intentional stylistic choices and undesirable artifacts without additional context. A distorted perspective might be an artistic choice in one context but a failure in another. This limitation is particularly relevant for creative applications where "errors" may be features. Future work should explore incorporating user intent or task specifications into the evaluation pipeline.

\textbf{Static Visual Pattern Set.}
Our current implementation uses a fixed set of discovered patterns, which may become outdated as generative models evolve and develop new failure modes. While our LMM-based curation pipeline provides some flexibility, adapting to entirely new types of generation errors requires retraining the underlying sparse autoencoders. A recent development in the literature, Matryoshka Transcoders, can be utilized to enhance the visual pattern set with targeted pattern discovery \citep{tang2025doesmodelfailautomatic}.

\textbf{Ambiguity in Medical Pattern Boundaries.}
CXR-LanSE may exhibit disagreement in cases where clinically distinct conditions produce visually similar radiological appearances. A representative example is pulmonary edema and atelectasis, which can be difficult to distinguish even for expert clinicians. In such cases, a single neuron may respond to both conditions, while the LMM labels it with only one.

\end{document}